\documentclass[10pt,twocolumn,letterpaper]{article}

\usepackage{wacv}
\usepackage{times}
\usepackage{epsfig}
\usepackage{graphicx}
\usepackage{amsmath}
\usepackage{amssymb}
\usepackage{booktabs}

\makeatletter
\@namedef{ver@everyshi.sty}{}
\makeatother

\usepackage{url}
\usepackage{tikz}
\usetikzlibrary{arrows,shapes,arrows,calc}
\usepackage{pgfplots}

\usepackage{subfig}

\usepackage{multirow}
\usepackage{makecell}

\usepackage{cuted}

\usepackage[accsupp]{axessibility}

\usepackage{xcolor}
\definecolor{ForestGreen}{rgb}{0.13, 0.55, 0.13}
\usepackage{pifont}%

\usepackage{enumitem}
\setitemize{noitemsep,topsep=0pt,parsep=0pt,partopsep=0pt,leftmargin=*}
\setenumerate{noitemsep,topsep=0pt,parsep=0pt,partopsep=0pt,leftmargin=*}

\usepackage{amsmath,amsfonts,bm}

\def\eqref#1{equation~\ref{#1}}

\def\1{\bm{1}}

\def\vx{{\bm{x}}}

\DeclareMathAlphabet{\mathsfit}{\encodingdefault}{\sfdefault}{m}{sl}
\SetMathAlphabet{\mathsfit}{bold}{\encodingdefault}{\sfdefault}{bx}{n}

\definecolor{greenNew}{RGB}{100, 240, 100}
\definecolor{redOld}{RGB}{200, 100, 100}

\definecolor{orangeFlorian}{RGB}{252, 61, 9}

\newcommand{\pseudoparagraph}[1]{\textbf{#1.\quad} }

\newsavebox\CBox
\def\textBF#1{\sbox\CBox{#1}\resizebox{\wd\CBox}{\ht\CBox}{\textbf{#1}}}

\def\wacvPaperID{889} %

\wacvalgorithmstrack   %

\wacvfinalcopy %

\ifwacvfinal
\usepackage[breaklinks=true,bookmarks=false]{hyperref}
\else
\usepackage[pagebackref=true,breaklinks=true,colorlinks,bookmarks=false]{hyperref}
\fi

\usepackage[capitalize,noabbrev]{cleveref}
\begin{document}

\title{Neural Implicit Representations for \\
Physical Parameter Inference from a Single Video}

\author{Florian Hofherr$^{1,2}$
\qquad
Lukas Koestler$^{1,2}$
\qquad
Florian Bernard$^3$
\qquad
Daniel Cremers$^{1,2}$\\[2pt]
{\small$^1$Technical University of Munich \quad\ $^2$Munich Center for Machine Learning \quad\ $^3$University of Bonn}
\vspace{1em}
}

\maketitle

\begin{figure*}[t]%
    \centering%
    \newcommand\WidthIms{2.15cm}%
\newcommand\Raiseheight{0.03\textwidth}%
\setlength\tabcolsep{2 pt}%
\newcommand{\imlabel}[0]{{\scriptsize synth~$|$~real~~~~}}
\begin{tabular}{ccccccc}%
 \imlabel & \imlabel & \imlabel & \imlabel & \imlabel & \imlabel & {\scriptsize pendulum length}\vspace{-3mm}\\
\subfloat{\includegraphics[width=\WidthIms]{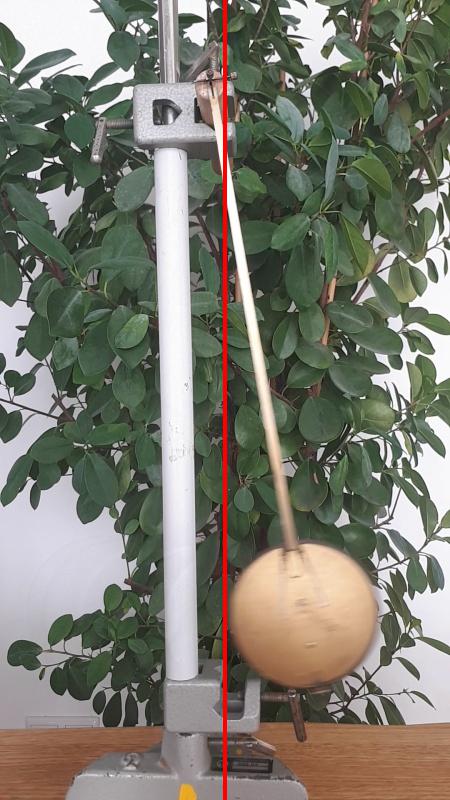}} &
\subfloat{\includegraphics[width=\WidthIms]{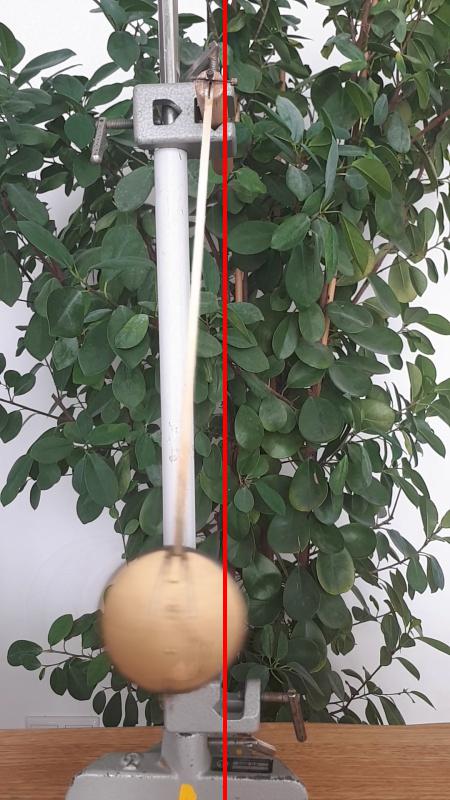}} &
\subfloat{\includegraphics[width=\WidthIms]{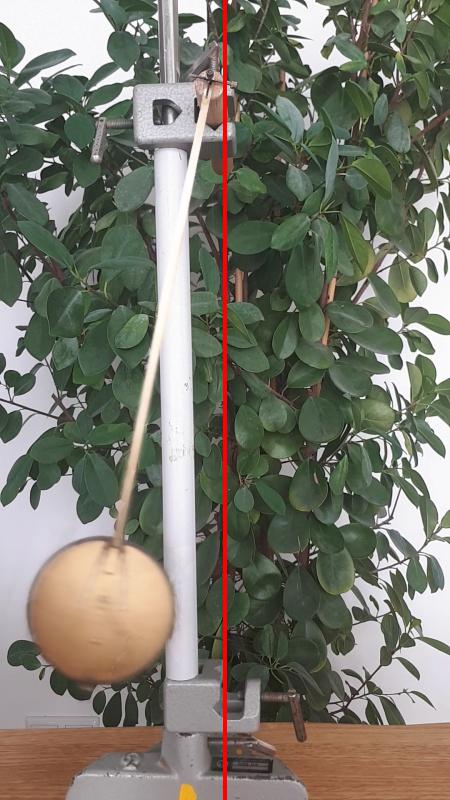}} &
\subfloat{\includegraphics[width=\WidthIms]{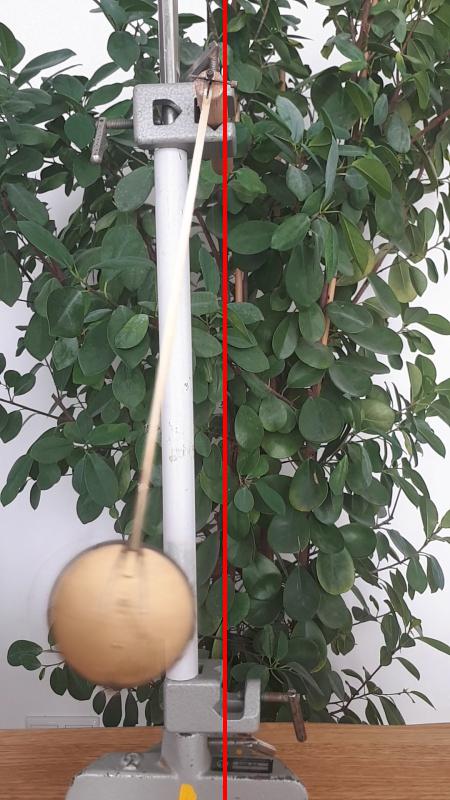}} &
\subfloat{\includegraphics[width=\WidthIms]{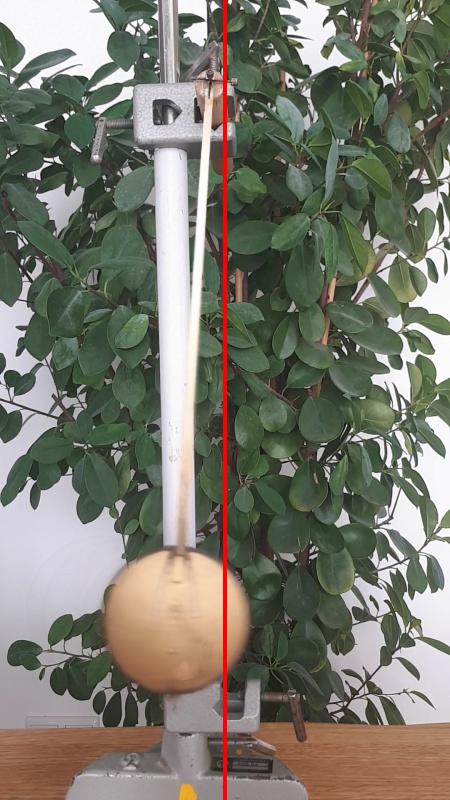}} &
\subfloat{\includegraphics[width=\WidthIms]{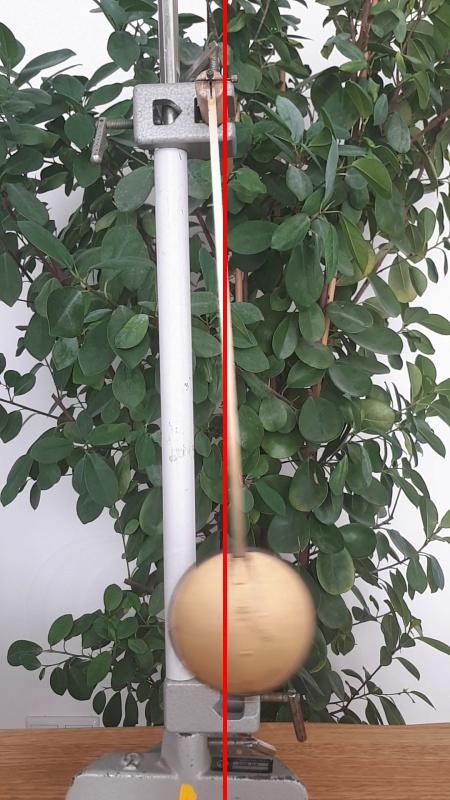}} &
~\subfloat{\includegraphics[width=\WidthIms]{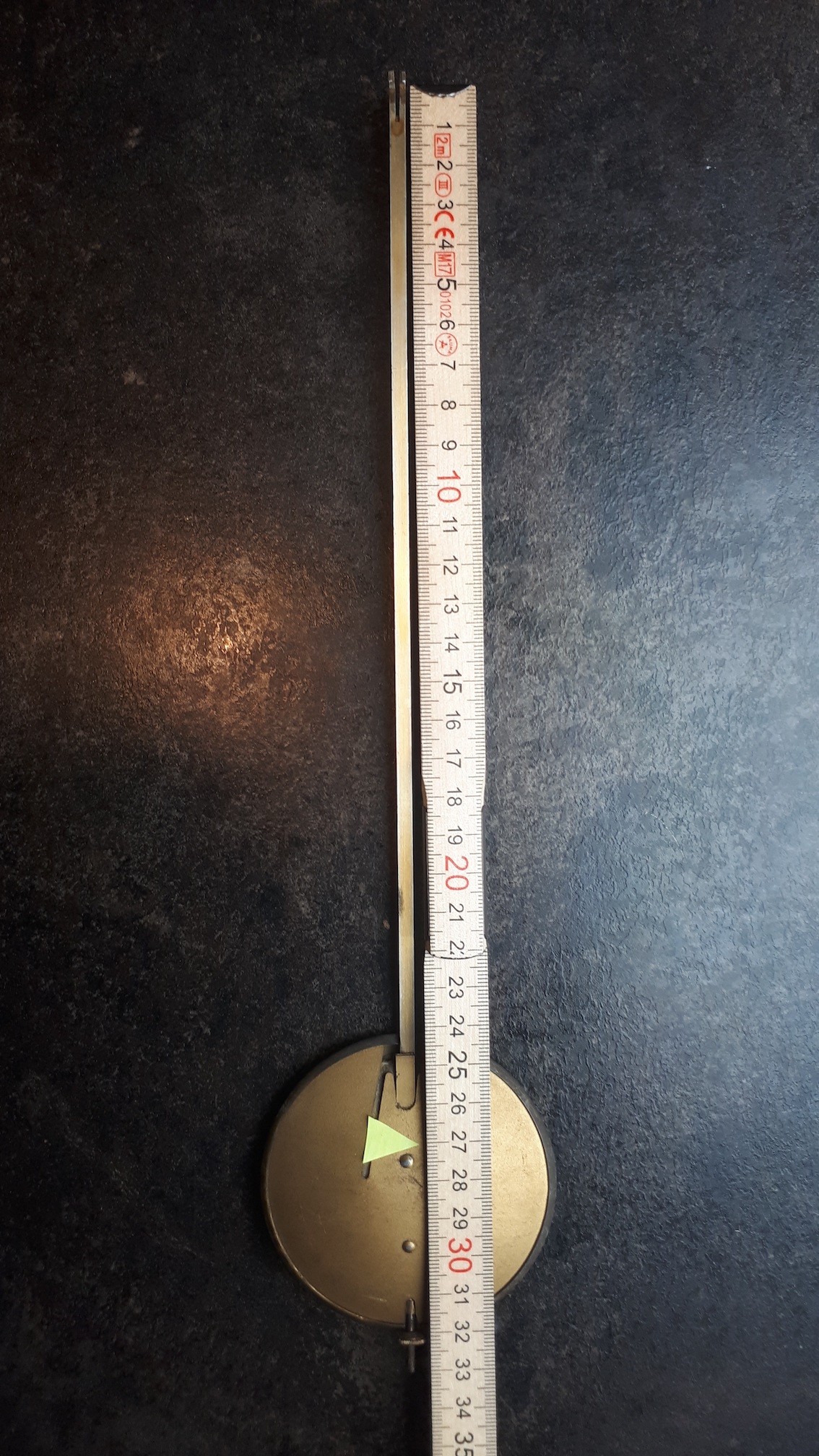}}
\end{tabular}%
    \caption{Our method infers physical parameters directly from real-world videos, like the shown pendulum motion. Separated by the red line, the right half of each image shows the input frame, and the left half shows our reconstruction based on physical parameters that we estimate from the input. We show 6 out of 10 frames that were used for training. The proposed model can precisely recover the metric length of the pendulum from the monocular video (relative error to true length is less than 4.1\%). Best viewed on screen with magnification. Please also consider the supplementary video.
    }
    \label{fig:realWorldPendulumMotion}
\end{figure*}

\begin{abstract}
   Neural networks have recently been used to analyze diverse physical systems and to identify the underlying dynamics. While existing methods achieve impressive results, they are limited by their strong demand for training data and their weak generalization abilities to out-of-distribution data. To overcome these limitations, we propose to combine neural implicit representations for appearance modeling with neural ordinary differential equations (ODEs) for modelling planar physical phenomena to obtain a dynamic scene representation that can be identified directly from visual observations. Our proposed model combines several unique advantages: (i) Contrary to existing approaches that require large training datasets, we are able to identify physical parameters from only a single video. (ii) The use of neural implicit representations enables the processing of high-resolution videos and the synthesis of photo-realistic images. (iii) The embedded neural ODE has a known parametric form that allows for the identification of interpretable physical parameters, and (iv) long-term prediction in state space. (v) Furthermore, the photo-realistic rendering of novel scenes with modified physical parameters becomes possible.
\end{abstract}

\section{Introduction}

For many physical phenomena, humans are able to infer (a rough estimation of) physical quantities from observing a scene, and are even capable to predict what is going to happen in the (near) future. In contrast,  physical understanding from videos is an open problem in machine learning. 
The physics of many real-world phenomena can be described concisely and accurately using differential equations. However, such equations are usually formulated in terms of abstracted quantities that are typically not directly observable using commodity sensors, such as cameras. For example, the dynamics of a pendulum are described by the deflection angle and the angular velocity as the time-varying state and the damping coefficient, and the pendulum's length as parameters. Automatically extracting those physical parameters directly from video data is challenging.
Due to the difficulties in direct observation of those quantities in images, and their complex relationship with the physical process, measuring such quantities in experiments often necessitates a trained expert operating customised equipment.

Recently, the combination of
deep learning and physics has become popular, particularly in the context of video prediction. While earlier works \cite{DBLP:conf/iclr/LutterRP19,DBLP:conf/nips/GreydanusDY19,raissi2019physics,DBLP:journals/corr/abs-2003-04630,DBLP:conf/iclr/ZhongDC20,PhysRevE.101.062207,iten2020discovering,raissi2020hidden} require coordinate data, i.e.~abstracted physical quantities that are not directly accessible from the video, more recent works directly use image data \cite{DBLP:conf/aaai/StewartE17,DBLP:conf/nips/Belbute-PeresSA18,DBLP:conf/iclr/JaquesBH20,DBLP:conf/dagm/KandukuriA0S20,DBLP:conf/iclr/TothRJRBH20,DBLP:conf/iclr/LevineCSLGB20,DBLP:conf/nips/ZhongL20,DBLP:conf/iclr/KossenSHVK20,DBLP:conf/nips/TakeishiK21}. A major downside of all these approaches is, that they rely on massive amounts of training data, and exhibit poor generalization abilities if the observation deviates from the training distribution, as we experimentally confirm. In contrast, our proposed combination of a parametric dynamics model with a neural scene representation allows for identification of the dynamics from only a single high resolution video. Also, due to our per-scene approach, our method does not suffer from generalization issues either.

Several of the previously mentioned works model physical systems using Lagrangian or Hamiltonian energy formulations \cite{DBLP:conf/iclr/LutterRP19,DBLP:conf/nips/GreydanusDY19,DBLP:journals/corr/abs-2003-04630,PhysRevE.101.062207,DBLP:conf/iclr/TothRJRBH20,DBLP:conf/iclr/ZhongDC20,DBLP:conf/iclr/LevineCSLGB20,DBLP:conf/nips/ZhongL20}, or other general physics models \cite{DBLP:conf/iclr/KossenSHVK20}. While those are a elegant approaches  that allow the model to adapt to different physical systems, they have two drawbacks. First, the general models are part of the reason why large amounts of data are required. Second, once the system dynamics have been identified, they are not easily interpretable. Questions like ``\emph{How would the scene look like if we double the damping}'' cannot be answered. In contrast, we estimate physically meaningful parameters of the underlying dynamics like the length of a pendulum or the friction coefficient of a sliding block. We find experimentally that using an ODE-based dynamics model gives more accurate long-term predictions. Moreover, due to the combination with the photo-realistic rendering capacities of our neural appearance representation, we are able to re-render the scene with adapted parameters.

\noindent
We summarize our main contributions as follows:
\begin{enumerate}
    \item We present the first method that uses neural implicit representations to identify physical parameters for planar dynamics from a single video.
    \item Our approach infers parameters of an underlying ODE-based physical model that directly allows for interpretability and long-term predictions.
    \item The unique combination of powerful neural implicit representations with rich physical models  allows to synthesize high-resolution and photo-realistic imagery. Moreover, it  enables physical editing by rendering  novel scenes with modified physical parameters.
    \item Contrary to existing learning-based approaches that require large corpora of training data, we propose a \emph{per-scene} model, so that only a single short video clip that depicts the physical phenomenon is necessary.
\end{enumerate}
See \href{https://florianhofherr.github.io/phys-param-inference/}{\textBF{\url{https://florianhofherr.github.io/phys-param-inference}}} and the appendix for architecture \& training details and the supplementary video.
This work is of fundamental character and thus has less immediate potential for negative societal impact. We discuss this in detail in the appendix.

\section{Related Work}
The combination of machine learning and physics has been addressed
across a broad range of topics. 
Machine learning was used to aid physics research \cite{bogojeski2020quantum,leclerc2020machine}, and physics was used within machine learning models, %
e.g.\
for automatic question answering from videos \cite{DBLP:conf/iclr/ChenM0WTG21,DBLP:journals/corr/abs-2106-08261}.
A great overview over physics-informed machine learning can be found in \cite{karniadakis2021physics}.
In this work we focus specifically on extracting physical models from 
videos, so that 
we discuss related works that we consider most relevant in this context.

\pseudoparagraph{Physical dynamics in the context of learning}
While neural networks have led to remarkable results across diverse domains, the inference and representation of physical principles like energy conservation, 
is still a challenge in the context of learning and requires additional constraints.
Generalized energy functions are one way to endow models with physics-based priors.
For example, \cite{DBLP:conf/nips/GreydanusDY19,PhysRevE.101.062207} and \cite{DBLP:conf/iclr/TothRJRBH20} use a neural network to parameterize the Hamiltonian of a system, which relates the total energy to the change of the state. This approach allows to infer the dynamics of systems with conserved energy, like an undamped pendulum. \cite{DBLP:journals/corr/abs-2201-10085} augment the Hamiltonian by a learned Rayleigh dissipation function to model systems that do not conserve energy, which are more common in the real world 
\cite{galley2013classical}.

One disadvantage of the Hamiltionian is that \emph{canonical coordinates} need to be used. To eliminate this constraint, other works use the Lagrangian to model the energy of the system. Since this formalism is more complex, \cite{DBLP:conf/iclr/LutterRP19} and \cite{DBLP:conf/nips/ZhongL20} restrict the Lagrangian to the case of rigid-body dynamics to model systems with multiple degrees of freedom, such as a pole on a cart, or a robotic arm. \cite{DBLP:journals/corr/abs-2003-04630} use a neural network to parameterize a general Lagrangian to infer the dynamics of a relativistic particle in a uniform potential.

Another problem of many previous approaches is that they do not allow for interpretation of individual learned system parameters. For example, \cite{DBLP:conf/cvpr/GuenT20a} learns dynamics in the form of a general PDE in a latent space, which, like the aforementioned works based on energy functions, prohibits interpretation of the learned physical model (e.g~in the form of interpretable parameters). In contrast, there are also approaches that explicitly incorporate the underlying dynamics into learning frameworks.
\cite{DBLP:conf/iclr/JaquesBH20} unroll the Euler integration of the ordinary differential equation of bouncing balls, as well as balls connected by a spring, to identify the physical parameters like the spring constant. \cite{DBLP:conf/dagm/KandukuriA0S20} and \cite{DBLP:conf/nips/Belbute-PeresSA18} propose to use a linear complementarity problem to differentiably simulate rigid multi-body dynamics that  can also handle object interaction and friction. \cite{raissi2018numerical} and \cite{raissi2018hidden} add uncertainty propagation by combining numeric stepping schemes with Gaussian processes. For our method, we also rely on the advantages of modelling the underlying physics explicitly to obtain interpretable parameter estimates.

\pseudoparagraph{Inferring physical properties from video}
While many approaches work with trajectories in state space, there are also several works that operate directly on videos. In this case, the information about physical quantities is substantially more abstract, so that uncovering 
dynamics from video data is a significantly more difficult problem. In their seminal work \cite{DBLP:conf/nips/WuYLFT15} consider objects sliding down a plane. By tracking the objects, they estimate velocity vectors that are used to supervise a rigid body simulation of the respective object.
Similarly, \cite{DBLP:conf/l4dc/JaquesABH22} track the trajectories of 
keypoints for more complex rigid body motions like a bouncing ball, and estimate the physical parameters 
and
the most likely model from a family of possible models by comparing the tracked trajectory to the projection of a simulated 3D trajectory. Both methods rely on correctly identifying the object tracks and do not use the rich information contained in the image directly. Also, video extrapolation is not easily possible. \cite{DBLP:conf/cvpr/Weiss0CWT20} and \cite{kair2022sft} consider deformable objects and solve a partial differential equation to simulate the deformations. Wile the first method uses depth values 
as supervision, the second one employs a photometric loss.
\cite{DBLP:journals/corr/abs-2104-02735} extract vibration modes from a video and identify the material parameters by comparing to the eigenmodes of the object mesh. While those methods show impressive results, all three require a 3D template mesh as additional information, which may limit their practical applicability.

More recently, several end-to-end learning approaches have been proposed. \cite{DBLP:conf/iclr/KossenSHVK20} combine the state prediction of an LSTM from an image with the prediction of a graph neural network from the previous state to propagate the state in time. Using the Sum-Product Attend-Infer-Repeat (SuPAIR) model (\cite{DBLP:conf/icml/StelznerPK19}) they render images from the state predictions and use the input image sequence as supervision.
\cite{DBLP:conf/nips/Belbute-PeresSA18,DBLP:conf/iclr/JaquesBH20} and \cite{DBLP:conf/dagm/KandukuriA0S20} use an encoder to extract the initial state of several objects from the combination of images and object masks. After propagating the physical state over time, they use carefully crafted decoders to transform the state back into images to allow for end-to-end training.
\cite{DBLP:conf/nips/ZhongL20} and \cite{DBLP:conf/iclr/TothRJRBH20} use a variational autoencoder (VAE) to predict posterior information about the initial state and combine this with an energy based representation of the dynamics and a final decoding stage. \cite{DBLP:conf/nips/TakeishiK21} improve the VAE based approach by using known explicit physical models as prior knowledge.
\cite{DBLP:journals/corr/abs-1911-11893} combine Mask R-CNN \cite{DBLP:conf/iccv/HeGDG17} with a VAE to predict symbolic equations.
All of these approaches require large amounts of data to train the complex encoder and decoder modules. In contrast, our approach does not rely on trainable encoder or decoder structures. Instead it combines neural implicit representations to model the scene appearance with the estimation of the parameters of a known, parameteric ODE, and is able to infer physical models from a single video.

\pseudoparagraph{Implicit representations}
Recently, neural implicit representations have gained popularity due to their theoretical elegance and performance in novel view synthesis. The idea is to use a neural network to parametrize a function that maps a spatial location to a spatial feature. For example occupancy values \cite{DBLP:conf/cvpr/MeschederONNG19,DBLP:conf/cvpr/ChenZ19,DBLP:conf/eccv/PengNMP020}, or signed distance functions \cite{DBLP:conf/cvpr/ParkFSNL19,DBLP:conf/icml/GroppYHAL20,DBLP:conf/cvpr/AtzmonL20} can be used to represent geometric shapes. In the area of multiview 3D surface reconstruction as well as novel view synthesis, a representation for density or signed distance, is combined with neural color fields to represent shape and appearance \cite{DBLP:conf/nips/SitzmannZW19,DBLP:conf/eccv/MildenhallSTBRN20,DBLP:conf/nips/YarivKMGABL20,DBLP:conf/cvpr/NiemeyerMOG20,DBLP:journals/corr/abs-2104-04532}. To model dynamic scenes, there have been several approaches that parametrize a displacement field and model the scene in a reference configuration \cite{DBLP:conf/iccv/NiemeyerMOG19,DBLP:journals/corr/abs-2011-12948,DBLP:conf/cvpr/PumarolaCPM21}.
On the other hand, several approaches \cite{xian2021space,li2021neural,du2021neural} include the time as an input to the neural representation and regularize the network using constraints based on appearance, geometry, and pre-trained depth or flow networks -- however, none of these methods uses physics-based constraints, e.g. by enforcing Newtonian motion.
An exeption is the work by Song et al.\ that use the solution of an ODE as regularization of a motion network to crate dynamic NeRFs \cite{DBLP:conf/mm/SongLLLDXY21}. In contrast to our work, this approach does not enforce the physics to be exact.
While the majority of works on implicit representations focuses on shape, \cite{DBLP:conf/nips/SitzmannMBLW20} show the generality of implicit representations by representing images and audio. We combine such representations with explicit physical models.

\section{Estimating Physical Models with Neural Implicit Representations}
\begin{figure*}
    \centering
    \input{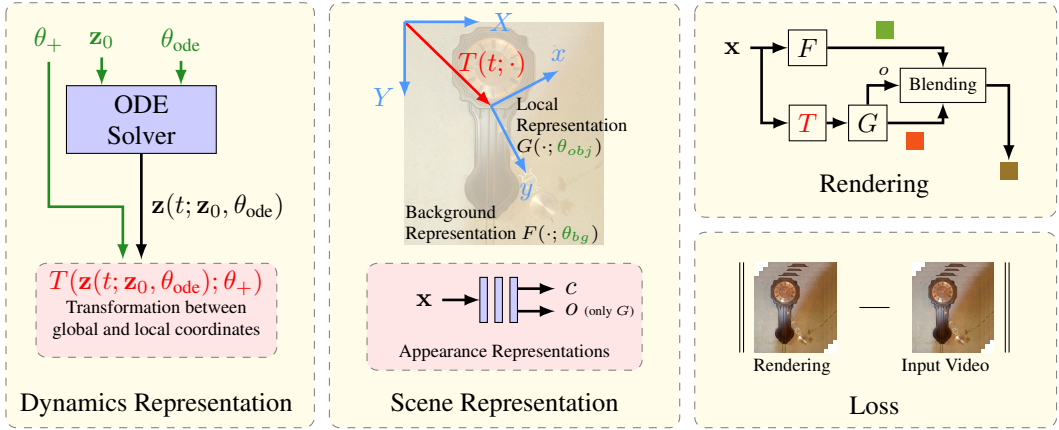}
    \vspace{-0.1cm}
    \caption{Overview of our approach. \textbf{Dynamics Representation:} The dynamics in the video are modelled by an ordinary differential equation (ODE), which is solved depending on unknown initial conditions $\mathbf{z}_0$ and unknown physical parameters $\theta_{\text{ode}}$. The solution curve $\mathbf{z}\left(t; \mathbf{z}_0, \theta_{\text{ode}}\right)$ is used to parametrize a time-dependent transformation $T(\mathbf{z}\left(t; \mathbf{z}_0, \theta_{\text{ode}}\right),\theta_{+})$ from the global coordinates $XY$ of the background to the local coordinates $xy$ of the moving object. The (unknown) parameters $\theta_{+}$ encode additional degrees of freedom of the transformation, for example the pivot point of a pendulum.
    \textbf{Scene Representation:}
    The functions $F(\cdot;\theta_{\text{bg}})$ and $G(\cdot;\theta_{\text{obj}})$ are neural networks that model the appearance of the background and of the object,
    using color $c$ and opacity $o$ (only for the foreground objects).
    \textbf{Rendering: }Rendering is done by blending the foreground and the background color based on the opacity of the foreground objects.
    \textbf{Loss:}
    We can estimate the unknown physical parameters for a given video based on a rendering loss which penalizes the discrepancy between the input video frames and the rendered video. All estimated parameters and network weights are shown in green text in the figure.
    }
    \vspace{-0.1cm}
    \label{fig:overview}
\end{figure*}

Our main goal is the estimation of physical parameters from a single video. We focus on the setting of a static camera, a static background, and rigid objects that are moving according to some physical phenomenon and exhibit a motion that can be restricted to a plane.
We model the dynamics of the objects using an ordinary differential equation (ODE) and use implicit neural representations to model the appearance, where the static background and the planar dynamics allow us to model the appearance in 2D. Our objective is to estimate the unknown physical parameters, and the initial conditions, of the ODE, as well as the parameters of the appearance representations. For estimating these quantities directly from an input video, we utilise a photometric loss that imposes similarity between the generated frames and the input. Due to the parametric dynamics model and the photorealistic appearance representation, we can use the result also as a generative model to render videos with varying physical parameters. We would like to note that neural radiance fields have shown convincing performance in 3D and hence the proposed method is a promising step towards physical parameter estimation in three dimensions.

\subsection{Modeling the Dynamics}
For most of the dynamics that can be observed in nature, the temporal evolution of the state can be described by an ODE. For example, for a pendulum the state variables are the deflection angle and the angular velocity, and a two dimensional ODE can be used to describe the dynamics. %

In general, we write $\dot{\mathbf{z}}=f\left(\mathbf{z}, t; \theta_{ode}\right)$ to describe the ODE\footnote{W.l.o.g.~we consider first-order ODEs here, since it is possible to reduce the order to one by introducing additional state variables.}, where $\mathbf{z}\in\mathbb{R}^n$ denotes the state variable, $t\in\mathbb{R}$ denotes time and $\theta_{ode}\in\mathbb{R}^m$ are the unknown physical parameters. 
Using the initial conditions $\mathbf{z}_0\in\mathbb{R}^n$ at the initial time $t_0$, we can write the solution of the ODE as
\begin{equation}\label{eq:generalSolutionODE}
    \mathbf{z}\left(t; \mathbf{z}_0, \theta_{ode}\right) = \mathbf{z}_0 + \int_{t_0}^{t}f\left(\mathbf{z}(\tau), \tau; \theta_{ode}\right)\text{d}\tau.
\end{equation}
Note that the solution curve $\mathbf{z}\left(t; \mathbf{z}_0, \theta_{ode}\right)\subset \mathbb{R}^n$ depends both on the unknown initial conditions $\mathbf{z}_0$, as well as on the unknown physical parameters $\theta_{ode}$.

In practice, the solution to \Cref{eq:generalSolutionODE} is typically approximated by numeric integration. In our context of physical parameter estimation from videos, we build upon \cite{DBLP:conf/nips/ChenRBD18}, who proposed an approach to compute gradients of the solution curve of an ODE with respect to its parameters. With that, it becomes possible to differentiate through the solution in \Cref{eq:generalSolutionODE} and therefore we can use gradient-based methods to estimate $\mathbf{z}_0$ and $\theta_{ode}$.

\subsection{Differentiable Rendering of the Video Frames}
To render the video frames, we draw inspiration from the recent advances in neural implicit representations.
To this end, we combine one representation for the static background with a representation for appearance and shape of dynamic foreground objects. By composing the learned background with the dynamic foreground objects, whose poses are determined by the solution of the ODE encoding the physical phenomenon, we obtain a dynamic representation of the overall scene. Doing so allows us to query the color values on a pixel grid, so that we are able to render video frames in a differentiable manner (cf.~\Cref{fig:overview}).

\pseudoparagraph{Representation of the background}
The appearance of the static background is modeled by a function $F(\cdot;\theta_{\text{bg}})$ that maps a 2D location $\mathbf{x}$ to a color value $\mathbf{c}\in\mathbb{R}^3$. We use a neural network with learnable parameters $\theta_{\text{bg}}$ to represent $F(\cdot;\theta_{\text{bg}})$.
To improve the ability of the neural network to learn high frequency variations in appearance, we use Fourier features~\cite{DBLP:conf/nips/TancikSMFRSRBN20} that map the input $\mathbf{x}\in\mathbb{R}^2$ to a vector $\gamma\left(\mathbf{x}\right)\in\mathbb{R}^{N_\text{Fourier}}$, where $N_\text{Fourier}$ is the numbers Fourier features used. The full representation of the background then reads $c_{\text{bg}}\left(\mathbf{x}\right) = F(\gamma\left(\mathbf{x}\right);\theta_{\text{bg}})$. For a more detailed discussion of the architecture, we refer to the appendix.

\pseudoparagraph{Representation of dynamic objects}
To compose the static background and the dynamically moving objects into the full scene, we draw inspiration from 
both \cite{DBLP:conf/cvpr/OstMTKH21} and \cite{DBLP:conf/cvpr/YuanLSL21}, who use implicit representations to decompose a dynamic 3D scene into a background representation and dynamically moving local representations. A drawback of their works is that they do not use physical dynamics models to constrain the dynamics, and therefore require strong supervisory signals like the trajectories and the dimensions of the bounding boxes in the first case or data from a multi-camera rig in the second case. In contrast, we use the transformation $T_t = T\left(\mathbf{z}\left(t;\mathbf{z}_0, \theta_{\text{ode}}\right),\theta_{+}\right)$ that is parametrized by the simulation of a physical phenomenon to position the dynamically moving local representation in the overall scene. Besides the unknown initial condition $\mathbf{z}_0$ and the physical parameters $\theta_{\text{ode}}$ of the ODE, we can use additional parameters $\theta_{+}$ for the parametrization. 
In case of the pendulum $\mathbf{z}_0$ are initial angle and anglular velocity, $\theta_\text{ode}$ contains the length and the damping and $\theta_{+}$ is the pivot point of a pendulum. See the appendix for more details.
$T_t$ is a time dependent, affine 2D transformation that maps from global to local coordinates and therefore can model a movements of the object in a plane that is parallel to the (static) camera.

Similarly to the background, the appearance of each individual dynamic object is modeled in terms of an implicit neural representation (in the local coordinate system).
In contrast to the background, we augment the color output $c\in\mathbb{R}^C$ of the dynamic object representation with an additional opacity value $o\in\left[0, 1\right]$, which allows us to model  objects with arbitrary shape.
We write the representation of a dynamic object in the global coordinate system as $\left(c_{\text{obj}}\left(\mathbf{x}\right), o\left(\vx\right)\right) = G(\gamma\left(\vx'\right);\theta_{\text{obj}})$, where $G(\cdot;\theta_{\text{obj}})$ is represented as a neural network with weights $\theta_{\text{obj}}$, $\gamma$ denotes the mapping to Fourier features, and $\vx' = T_t(\vx)$ is the local coordinate representation of the (global) 2D location $\vx$.

\pseudoparagraph{Homography to correct for non-parallel planes}
Since $T_t$ is an affine transformation, it can only model movements that are parallel to the (static) camera plane. However, in particular for the real world examples, the plane of the movements does not need to be parallel to the image plane, but could be tilted. The resulting nonlinear effects can be modeled by adding a learnable homography to the transformation from global to local coordinates. For clarity, we will not explicitly write the homography down, but rather consider it as a part of $T_t$. Note that no additional supervision is necessary to identify the homography.

\pseudoparagraph{Differentiable rendering}
For rendering we evaluate the composed scene appearance at a regular pixel grid, where
we use the opacity value of the local object representation to blend the color of the background and the dynamic objects. To obtain the color for the pixel $\vx$, we evaluate
\begin{equation}
\begin{aligned}
c(\vx, t) = (1 - o(\vx)) \, c_{{\text{bg}}}(\vx) + o(\vx) c_{{\text{obj}}}(\vx).
\end{aligned}
\end{equation}

Note that $c(\vx, t)$ is time dependent due to the time dependence of the transformation $T_t$. This allows us to render the frames of the sequence over time.

\subsection{Loss Function}
We jointly optimize for the parameters of the neural implicit representations $\theta_{\text{bg}}$ and $\theta_{\text{obj}}$ and estimate the physical parameters $\theta_{\text{ode}}$ and $\mathbf{z}_0$ and the transformation parameters $\theta_{+}$ and the homography matrix. To this end, we use a simple mean squared error loss between the predicted pixel values and the given images. During training we form batches of $N_{batch}$ pixels. To make the training more stable and help the model to identify the motion of the objects, we adopt the online training approach from \cite{DBLP:conf/cvpr/YuanLSL21} and progressively increase the number of frames used during the optimization. More details on the training can be found in the appendix.

\section{Experiments}
We use four challenging physical models to 
evaluate our proposed approach. To analyze our method and to compare to previous work, we first consider 
synthetic
data. Afterwards, we show that our method achieves strong results also on real-world data. For additional implementation details and 
results we refer the reader to the appendix.

Although several learning-based approaches that infer physical models from image data have been proposed \cite{DBLP:conf/nips/Belbute-PeresSA18,DBLP:conf/iclr/JaquesBH20,DBLP:conf/dagm/KandukuriA0S20,DBLP:conf/nips/ZhongL20,DBLP:conf/iclr/TothRJRBH20}, existing approaches are particularly tailored towards settings with large training corpora.
However, these methods typically suffer from decreasing estimation accuracy when training data are scarce or when out-of-distribution generalization is required, as we show in the appendix.
In contrast, our proposed approach is able to predict physical parameters from a single short video clip. Due to the lack of existing baselines tailored towards estimation from a single video, we adapt the recent work of \cite{DBLP:conf/iclr/JaquesBH20} and \cite{DBLP:conf/nips/ZhongL20} to act as baseline methods.

\subsection{Synthetic Data}
We compare the proposed method to two published methods \cite{DBLP:conf/iclr/JaquesBH20,DBLP:conf/nips/ZhongL20} and two baselines on synthetic data.

\begin{figure}%
    \newcommand\WidthIms{0.9cm}
\newcommand\Raiseheight{0.03\textwidth}
\newcommand\fontsizeCaption{\scriptsize}
\definecolor{plotRed}{rgb}{0.85000,0.32500,0.09800}
\setlength\tabcolsep{1 pt}
\resizebox{\columnwidth}{!}{%
\begin{tabular}{ccccccccc}%
\scriptsize%
\rotatebox{90}{\fontsizeCaption{}\hspace{-0.5mm}B: Overfit~}~\includegraphics[width=\WidthIms]{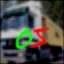} &
\includegraphics[width=\WidthIms]{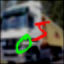}&%
\includegraphics[width=\WidthIms]{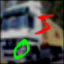} &
\includegraphics[width=\WidthIms]{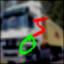} &
\includegraphics[width=\WidthIms]{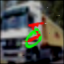} &
\includegraphics[width=\WidthIms]{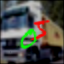} &
\includegraphics[width=\WidthIms]{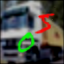} &
\includegraphics[width=\WidthIms]{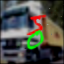} \\[-2.5ex]
\rotatebox{90}{~\fontsizeCaption B: Full}~\subfloat{\includegraphics[width=\WidthIms]{exampleSpring/Sequence6_paig_Overfit/00_recon.jpg}} &
\subfloat{\includegraphics[width=\WidthIms]{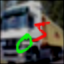}}&%
\subfloat{\includegraphics[width=\WidthIms]{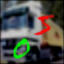}} &
\subfloat{\includegraphics[width=\WidthIms]{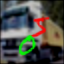}} &
\subfloat{\includegraphics[width=\WidthIms]{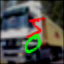}} &
\subfloat{\includegraphics[width=\WidthIms]{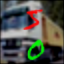}} &
\subfloat{\includegraphics[width=\WidthIms]{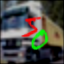}} &
\subfloat{\includegraphics[width=\WidthIms]{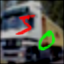}} \\[-2.5ex]
\rotatebox{90}{~~\fontsizeCaption Ours}~\subfloat{\includegraphics[width=\WidthIms]{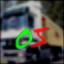}} &
\subfloat{\includegraphics[width=\WidthIms]{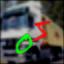}} &
\subfloat{\includegraphics[width=\WidthIms]{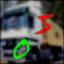}} &
\subfloat{\includegraphics[width=\WidthIms]{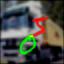}} &
\subfloat{\includegraphics[width=\WidthIms]{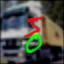}} &
\subfloat{\includegraphics[width=\WidthIms]{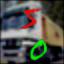}} &
\subfloat{\includegraphics[width=\WidthIms]{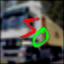}} &
\subfloat{\includegraphics[width=\WidthIms]{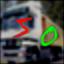}}  \\[-2.5ex]
\rotatebox{90}{~~~\fontsizeCaption GT}~\subfloat{\includegraphics[width=\WidthIms]{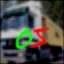}} &
\subfloat{\includegraphics[width=\WidthIms]{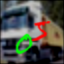}}&%
\subfloat{\includegraphics[width=\WidthIms]{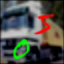}} &
\subfloat{\includegraphics[width=\WidthIms]{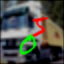}} &
\subfloat{\includegraphics[width=\WidthIms]{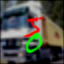}} &
\subfloat{\includegraphics[width=\WidthIms]{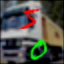}} &
\subfloat{\includegraphics[width=\WidthIms]{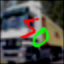}} &
\subfloat{\includegraphics[width=\WidthIms]{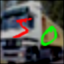}} \vspace{-1.5mm}\\%
\vspace{-1.5mm}
\scriptsize{1} &
\scriptsize{4} &
\scriptsize{7} &
\scriptsize{10} &
\scriptsize{13} &
\scriptsize{16} &
\scriptsize{22} &
\scriptsize{25} \\
 &
 &
 &
 &
\raisebox{0.15cm}{\begin{tikzpicture} \draw [-stealth, plotRed](0,0) -- (0.7,0);\end{tikzpicture}}&
 &
 &
 &
\end{tabular}%
}
\vspace{-0.45cm}
    \caption{Two masses spring system in which MNIST digits are connected by an (invisible) spring  (\cite{DBLP:conf/iclr/JaquesBH20} sequence 6). %
    The arrow indicates the start of the prediction of unseen frames. 
    We compare our results to  \cite{DBLP:conf/iclr/JaquesBH20}, both trained on the full dataset (B: Full) and overfitted to sequence 6 (B: Overfit).
    For the spring constant and equilibrium distance ($k$, $l$) the different methods achieve the  relative errors 
    $(2.7 \%,~81.0 \%)$ (B: Overfit); $(3.7 \%,~1.8 \%)$ B: Full; and $(\mathbf{0.7} \%,~\mathbf{0.7} \%)$ (Ours).  (Best viewed magnified on screen)
    }
    \label{fig:springSeq6}
\end{figure}

\pseudoparagraph{Two masses connected by a spring}
First, we consider the two moving MNIST digits connected by an (invisible) spring on a CIFAR background, from \cite{DBLP:conf/iclr/JaquesBH20}, see \Cref{fig:springSeq6}.
The 
dynamics are modeled as a two dimensional two-body system.
We use two separate local representations and enable the model to identify the object layering by using the maximum of both occupancy values. Besides the initial positions and velocities of the digits, the spring constant $k$, the equilibrium distance $l$ are the parameters that need to be identified. To guide the model in learning which local representation represents which digit, we use an additional segmentation loss with very rough object masks as supervision on the \emph{first} frame of the sequence. This loss is gradually reduced to enable the learning of the fine shape of the objects. For more details see the appendix.

The approach of \cite{DBLP:conf/iclr/JaquesBH20} uses a learnable encoder and velocity estimator to obtain initial positions and velocities of a known number of objects from the video. 
After integrating the known parametric model, they use a learnable coordinate-consistent decoder in combination with learned object masks and colors to render frames from the integrated trajectories. Using a photometric loss they require 
5000 
sequences of the same two masses spring system to train the model and identify the parameters. We report the results of their model trained on the full dataset (`B: Full'). 
In addition, to compare 
to our work in the setting of parameter estimation from a single video, we train their model on individual sequences of the test dataset (`B: Overfit').

\begin{table}[t]
    \centering
    \footnotesize
    \begin{tabular}{c|c|c|c}
 & PSNR $\uparrow$ & Param (Mean) $\downarrow$ & Param (Median) $\downarrow$ \\
 \hline\hline
\cite{DBLP:conf/iclr/JaquesBH20}: Overfit & 17.66 & 64.77 & 69.55 \\
\cite{DBLP:conf/iclr/JaquesBH20}: Full & 21.40 & \phantom{0}2.55 & \phantom{0}2.55 \\
Ours & \textbf{30.30} & \textbf{\phantom{0}2.47} & \textbf{\phantom{0}0.76}
\end{tabular}
\vspace{-0.2cm}
    \caption{PSNR and relative parameter errors (``Param'') in percent for our method and the overfitted and full baseline averaged over 10 test seqs.\ of the MNIST digits by \cite{DBLP:conf/iclr/JaquesBH20}. 
    }
    \label{tab:springNumbers}
\end{table}
\Cref{fig:springSeq6} shows a qualitative comparison of our results to the method of \cite{DBLP:conf/iclr/JaquesBH20} trained in the two settings explained above. We observe that for this sequence all approaches yield reasonable results for the reconstruction of the training frames. However, for extrapolation to unseen points in time, the overfitted model of \cite{DBLP:conf/iclr/JaquesBH20} performs significantly worse, indicating that the physical model is poorly identified from a single video. While both, the baseline trained on the full dataset and our method are able to identify the parameters with high accuracy, our methods achieves an even lower error, which leads to a more precise prediction of the future frames. The fact that we achieve comparable results while using significantly less data highlights the advantage of combining the explicit dynamics model with the implicit representation for the objects. Note that we chose sequence 6 in particular, since it yielded the best visual results for the baseline. \Cref{tab:springNumbers} shows a quantitative analysis of all 10 test sequences, highlighting again the advantages of our method in the setting of a single sequence as well as the competitiveness against the usage of considerably more data. More results can be found in the appendix.

\pseudoparagraph{Nonlinear damped pendulum}
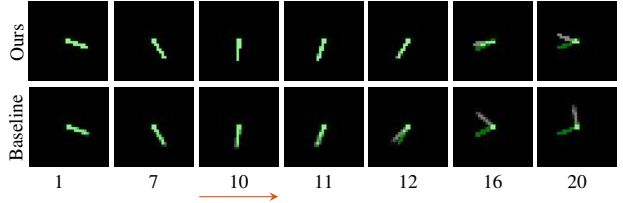
\begin{figure}
    \centering
    \newcommand\WidthIms{1.0cm} %
\newcommand\fontsizeCaption{\scriptsize}
\newcommand\Raiseheight{0.03\textwidth}
\definecolor{plotRed}{rgb}{0.85000,0.32500,0.09800}
\setlength\tabcolsep{1 pt}
\resizebox{\columnwidth}{!}{
\begin{tabular}{cccccccc}%
\scriptsize%
\rotatebox{90}{~~~\fontsizeCaption ~Ours}~\includegraphics[width=\WidthIms]{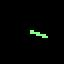} &
\includegraphics[width=\WidthIms]{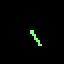} &
\includegraphics[width=\WidthIms]{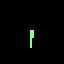} &
\includegraphics[width=\WidthIms]{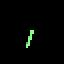} &
\includegraphics[width=\WidthIms]{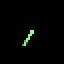} &
\includegraphics[width=\WidthIms]{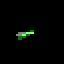} &
\includegraphics[width=\WidthIms]{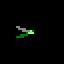} \\[-2.5ex]
\rotatebox{90}{\fontsizeCaption ~Baseline}~\subfloat{\includegraphics[width=\WidthIms]{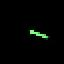}} &
\subfloat{\includegraphics[width=\WidthIms]{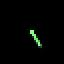}} &
\subfloat{\includegraphics[width=\WidthIms]{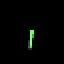}} &
\subfloat{\includegraphics[width=\WidthIms]{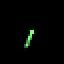}} &
\subfloat{\includegraphics[width=\WidthIms]{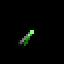}} &
\subfloat{\includegraphics[width=\WidthIms]{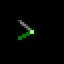}} &
\subfloat{\includegraphics[width=\WidthIms]{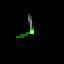}} \vspace{-1.5mm}\\%
\vspace{-1.5mm}
\scriptsize{1} &
\scriptsize{7} &
\scriptsize{10} &
\scriptsize{11} &
\scriptsize{12} &
\scriptsize{16} &
\scriptsize{20} \\
 &
 &
\raisebox{0.15cm}{\begin{tikzpicture} \draw [-stealth, plotRed](0,0) -- (1,0);\end{tikzpicture}}&
 &
 &
 &
\end{tabular}
}
\vspace{-0.45cm}

    \caption{Prediction when training on the first 10 frames of sequence 0 of the pendulum test data by \cite{DBLP:conf/nips/ZhongL20}.  Each image shows the prediction of the respective method in white, and the ground truth as green overlay.  For both methods, the prediction of images seen during training (frames 1,7,10) works well. For unseen  data (frames 11,12,16,20), our method leads to more reliable predictions, meaning that our physical parameter estimation is more accurate.
}
    \label{fig:qualitativeComparisonLagrangianVAE}
\end{figure}
We now consider the renderings of a nonlinear pendulum from \cite{DBLP:conf/nips/ZhongL20} (cf.~\cref{fig:qualitativeComparisonLagrangianVAE}). The sequences are created by OpenAI Gym~\cite{DBLP:journals/corr/BrockmanCPSSTZ16} and contain no RGB data, but only object masks. \cite{DBLP:conf/nips/ZhongL20} uses a coordinate aware variational encoder to obtain the initial state from object masks. After the state trajectory is integrated using a learnable Lagrangian function parametrizing the dynamics of the system, a coordinate aware decoder is used to render frames from the trajectories. To compare to our method in the setting of a single video, we train the model again using only the first $N$ frames of sequences from the test set.

In contrast to the baseline, we do not assume a known pivot point and use a more general pendulum model with damping. For a nonlinear damped pendulum the unknown parameters are the initial angle and velocity, the pivot point $A$, the pendulum length $l$ and the damping coefficient $c$. For more details see the appendix. Since this dataset does not include image data, we employ a binary cross entropy loss wrt.\ the object mask using the same frames as the baseline.

Qualitative results for a single sequence are presented in \Cref{fig:qualitativeComparisonLagrangianVAE}. We observe similar behavior as before: Both methods fit the given training data very well, however, in case of the baseline the pendulum motion significantly slows down for unseen time steps and the predictions for unseen data are not very accurate. We emphasize that this happens because due to the general dynamics model used, the baseline requires significantly larger training datasets, and it performs poorly in the single-video setting considered in this paper.
In contrast, our method shows a significantly better performance, which highlights the strength of directly modelling  physical phenomena to constrain the learnable dynamics in an analysis-by-synthesis manner. 

For a quantitative evaluation of the prediction quality, we report the intersection over union (IoU) averaged over all frames of the test sequences. Averaged over the first 20 sequences of the test set, the overfitted baseline achieves an IoU of $0.54$ while our method achieves a score of $0.76$. If we predict the test sequences using the baseline trained on the full dataset, we obtain an IoU of $0.73$. We point out again, that our method achieves results that are en par with the baseline trained on the full dataset, while requiring only a single sequence. Moreover, as we show in the appendix, that the baseline exhibits poor generalization abilities for observations that deviate from the training distribution, while our method does not encounter such problems.

\pseudoparagraph{Nonlinear damped pendulum - high resolution}
\begin{figure}
    \centering
    \newcommand\WidthIms{0.25\columnwidth}
\newcommand\Raiseheight{0.07\columnwidth}
\newcommand\distanceRows{-0.04cm}
\newcommand\distanceColumns{-0.5cm}
\newcommand\offsetLeftImage{-0.35cm}
\tikzstyle{arrow}=[->, line width=0.75pt, -to]
\begin{tabular}{c|cc}%
\rotatebox{90}{~~~~~~~Ours}~
\includegraphics[width=\WidthIms]{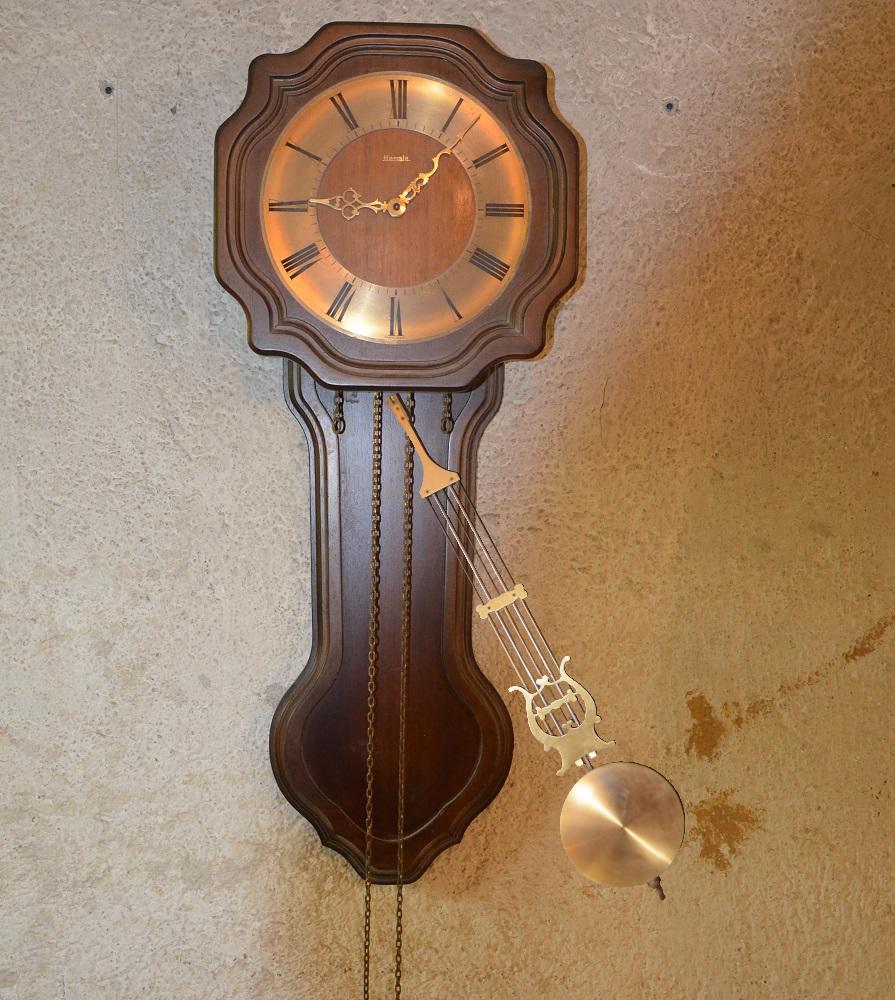} &
\hspace{\offsetLeftImage}
\begin{tikzpicture}
\node[anchor=south west, inner sep=0] at (0, 0) {\includegraphics[width=\WidthIms]{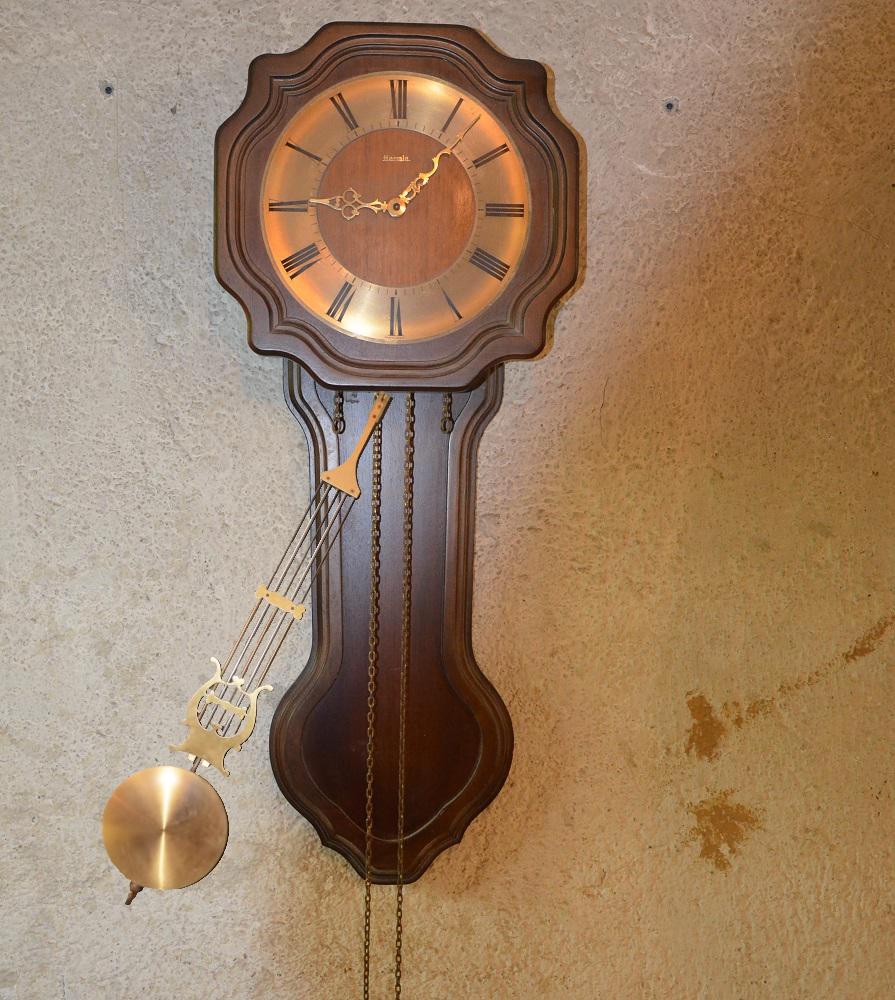}};
\draw [arrow, green] (0.2, 1.1) -- (0.55, 0.9) node[] {};
\end{tikzpicture}
\hspace{\distanceColumns} &
\includegraphics[width=\WidthIms]{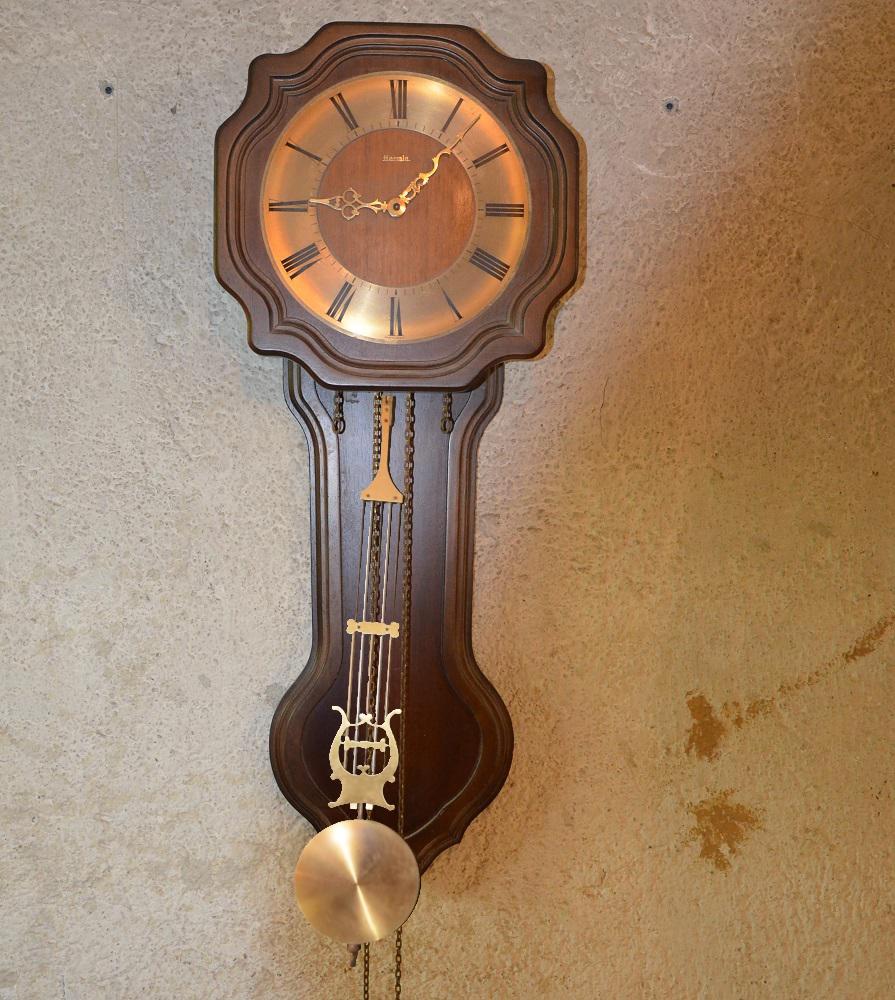} \vspace{\distanceRows}
\\
\rotatebox{90}{~~~~Baseline-t }~
\includegraphics[width=\WidthIms]{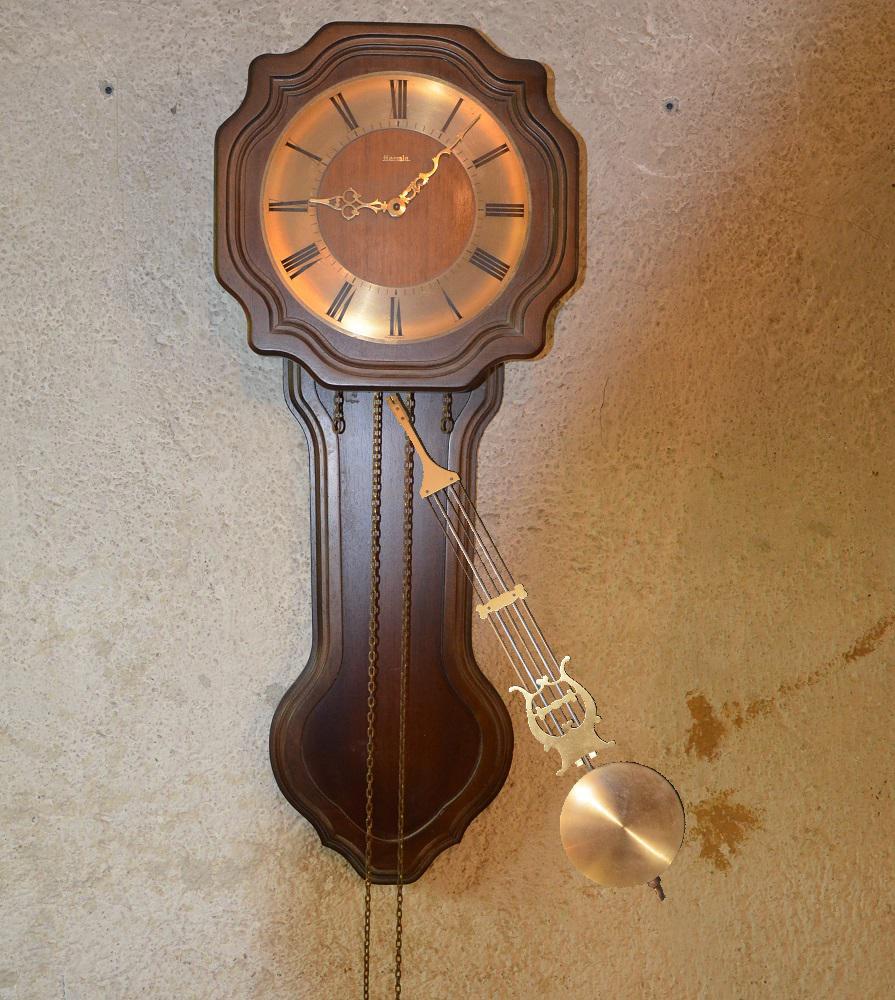} &
\hspace{\offsetLeftImage}
\begin{tikzpicture}
\node[anchor=south west, inner sep=0] at (0, 0) {\includegraphics[width=\WidthIms]{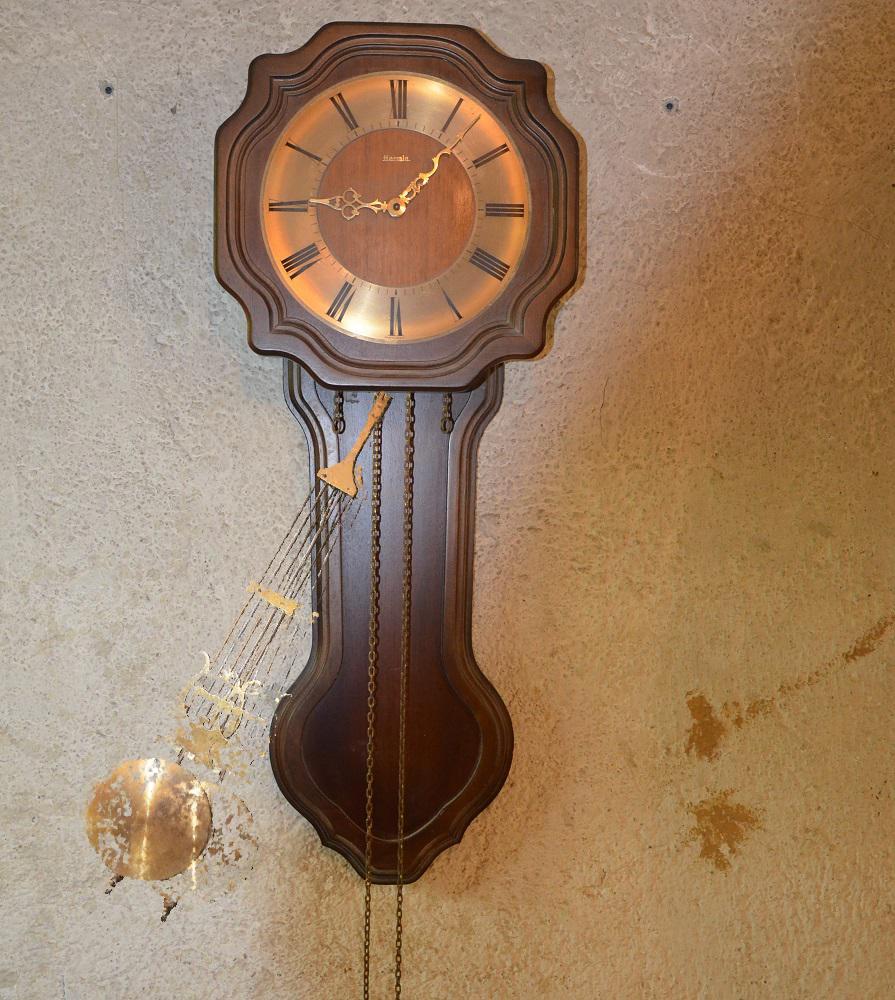}};
\draw [arrow, red] (0.15, 0.02) -- (0.24, 0.22) node[] {};
\draw [arrow, red] (0.55, 0.02) -- (0.42, 0.2) node[] {};
\end{tikzpicture}
\hspace{\distanceColumns} &
\includegraphics[width=\WidthIms]{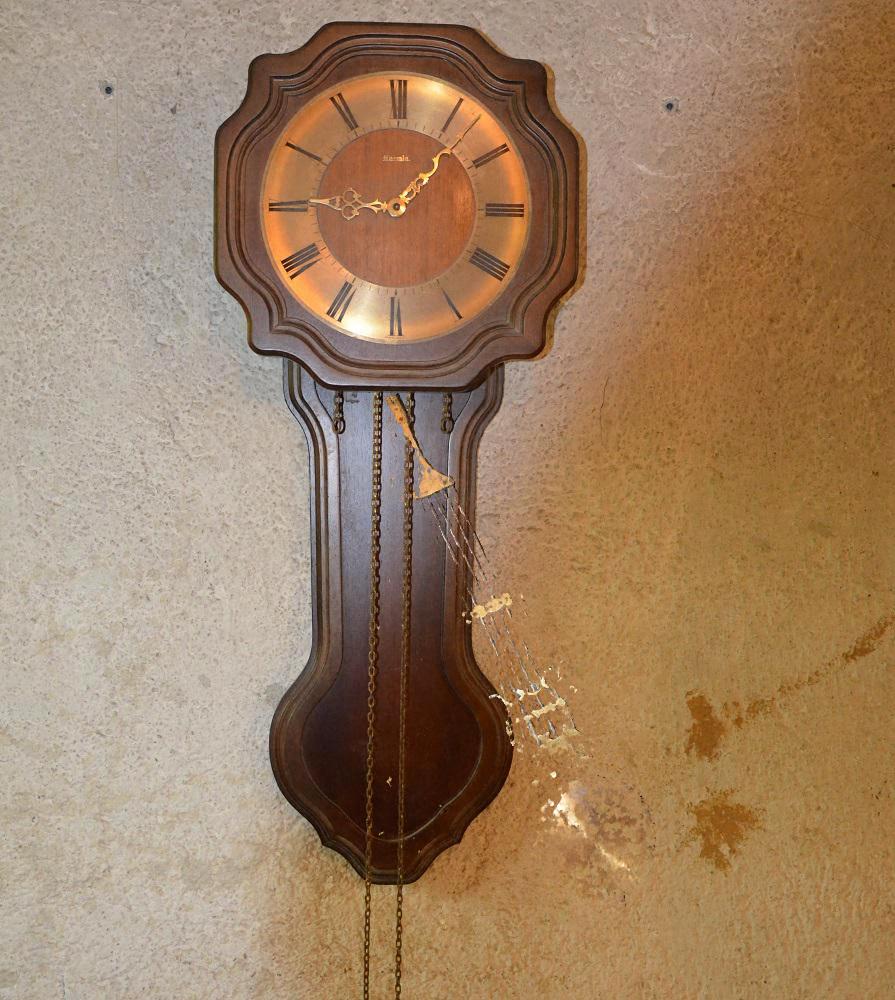} \vspace{\distanceRows}
\\
\rotatebox{90}{~~~~~~~~~~GT}~
\includegraphics[width=\WidthIms]{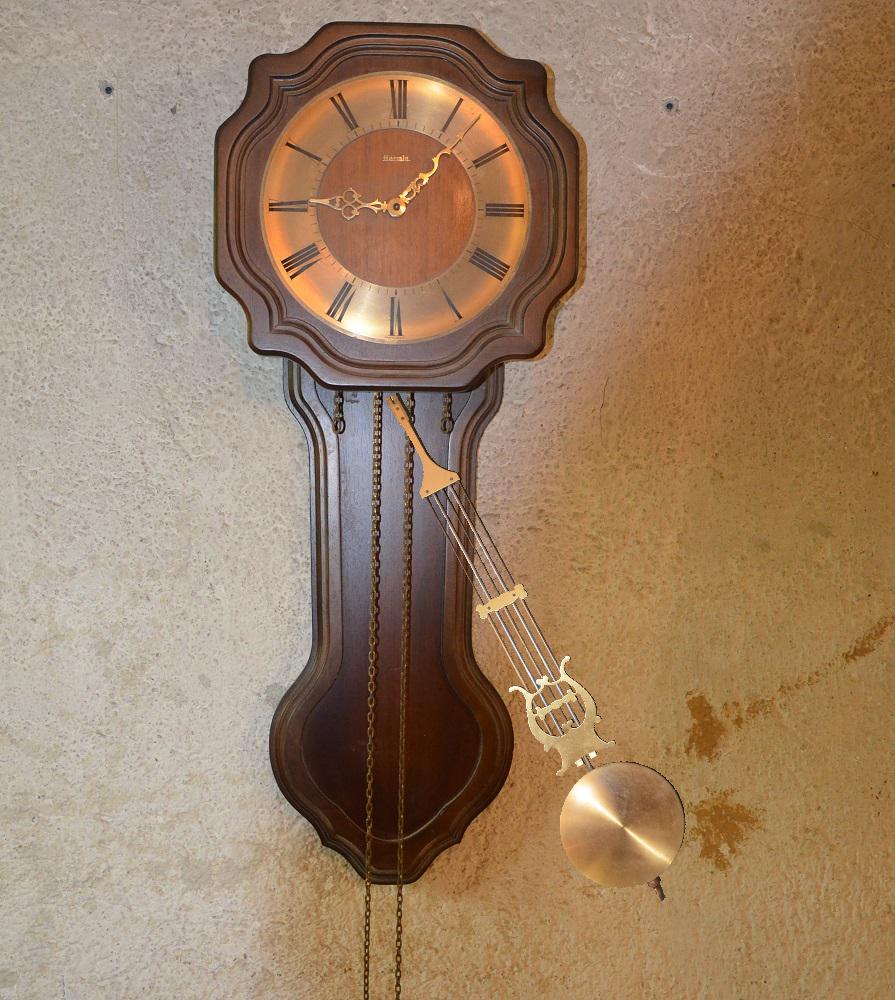} &
\hspace{\offsetLeftImage}
\begin{tikzpicture}
\node[anchor=south west, inner sep=0] at (0, 0) {\includegraphics[width=\WidthIms]{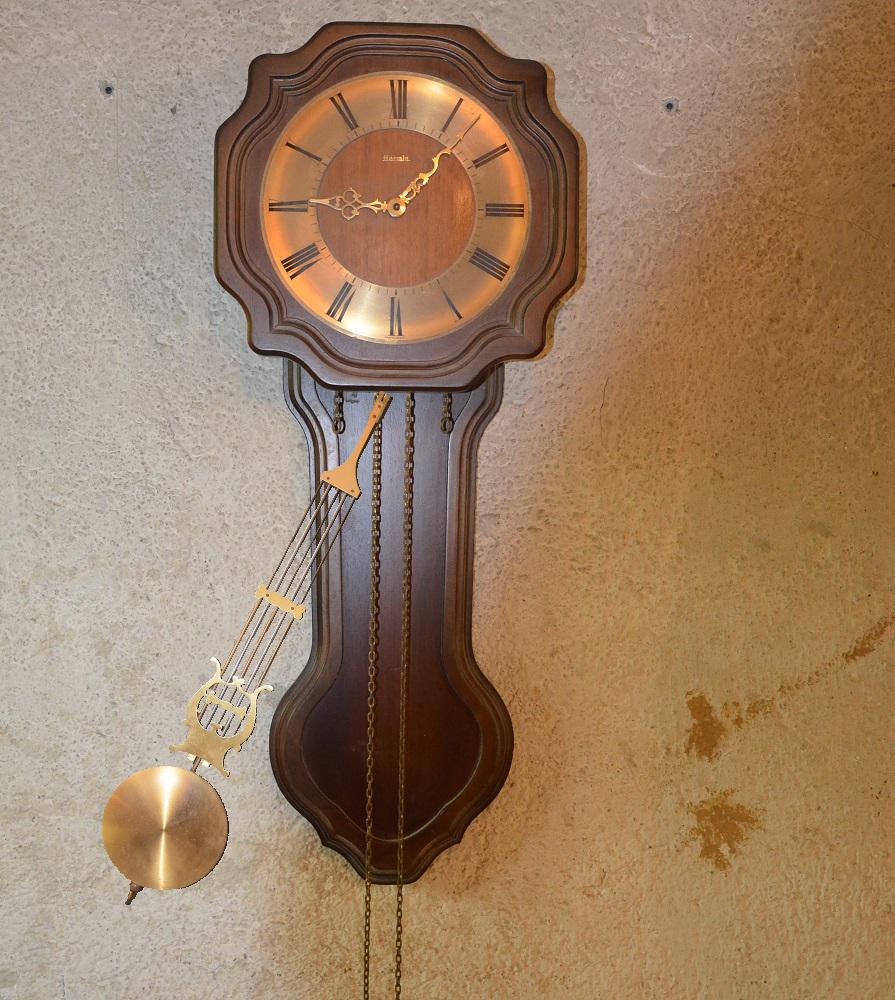}};
\end{tikzpicture}
\hspace{\distanceColumns} &
\includegraphics[width=\WidthIms]{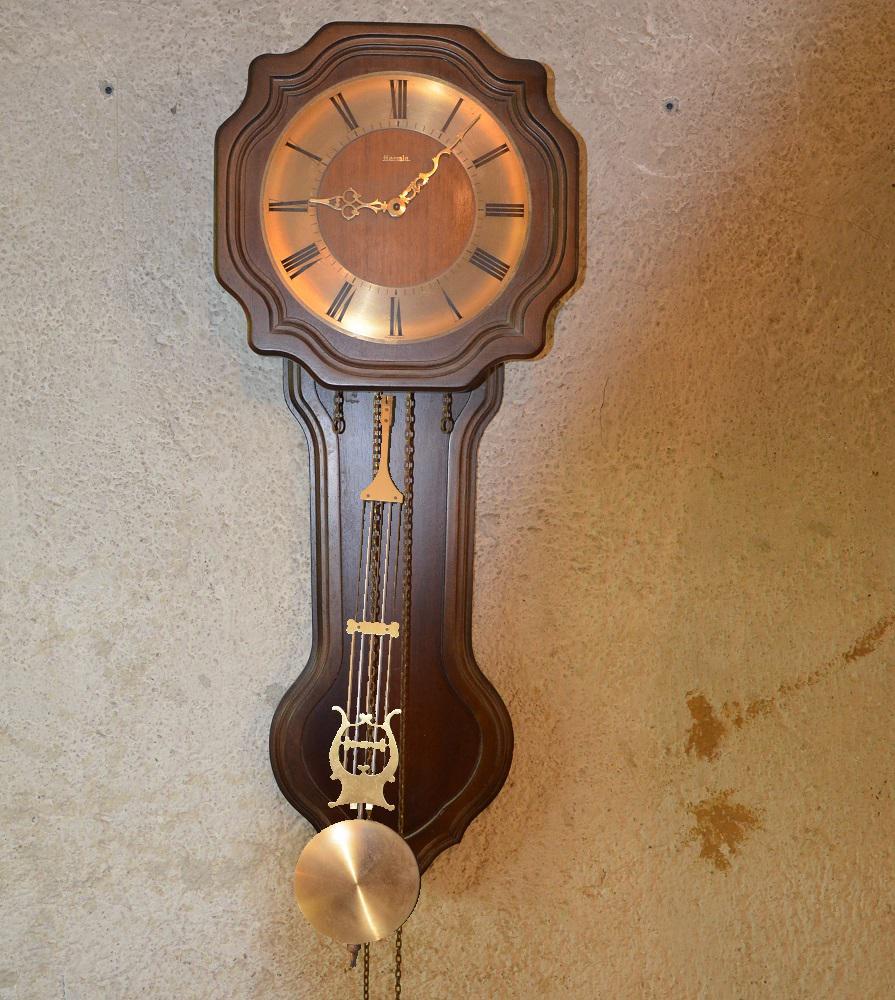}  \vspace{\distanceRows}
\\
{\footnotesize \quad training frame} & {\footnotesize \quad unseen frame 1} & {\footnotesize unseen frame 2} \\
\end{tabular}

\vspace{-0.2cm}
    \caption{Rendered frames for sequence 1 of the wallclock. The left image is part of the training set, ``unseen frame 1'' is between two training frames, ``unseen frame 2'' is a future frame after the interval seen during training. While our methods makes photorealistic predictions for both unseen frames, the time-dependent background (``Baseline-t'') fails in both cases. Note the visible blending between the neighboring frames in the baseline (red arrows) and the fine detail on the pendulum recovered by our method (green arrow).
    }
    \label{fig:results_synthetic_pendulum}
\end{figure}
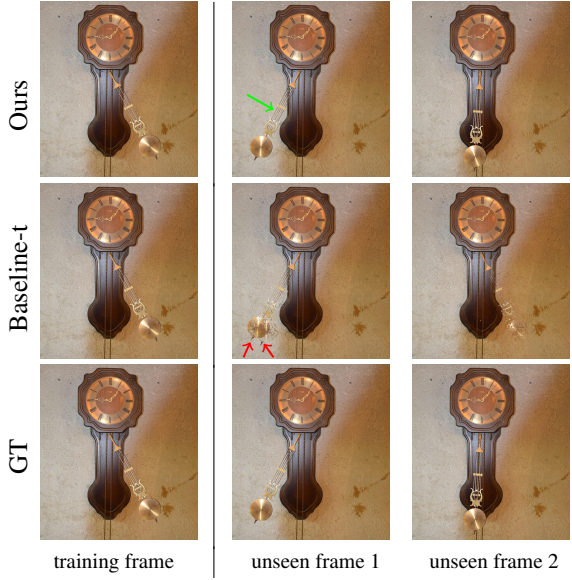
\begin{table*}[]
    \centering
    \footnotesize
    \begin{tabular}{c|ccc | ccc | ccc}
& \multicolumn{3}{c}{Woodwall} & \multicolumn{3}{c}{Stonewall} & \multicolumn{3}{c}{Wallclock} \\
& \scriptsize PSNR $\uparrow$ & \scriptsize IoU $\uparrow$ & \scriptsize Param $\downarrow$ & \scriptsize PSNR $\uparrow$ & \scriptsize IoU $\uparrow$ & \scriptsize Param $\downarrow$ & \scriptsize PSNR $\uparrow$ & \scriptsize IoU $\uparrow$ & \scriptsize Param $\downarrow$ \\
\hline\hline
Baseline-t &
29.29 & - & -     %
&
25.00 & - & -     %
& 
26.36 & - & - \\  %
Baseline-p &
33.48 & 0.86 & -     %
&
31.73 & 0.88 & -     %
& 
32.82 & 0.86 & - \\  %
Ours &
\textbf{42.07} & \textbf{0.98} & 0.01     %
&
\textbf{36.40} & \textbf{0.99} & 0.02     %
& 
\textbf{40.98} & \textbf{0.99} & 0.05 \\  %
\end{tabular}

\vspace{-0.1cm}
    \caption{Reconstruction quality on the test frames for the synthetic examples. We report IoU of the predicted vs.\ groundtruth masks and the relative error of all estimated physical parameters in percent (``Param'') averaged over the 9 sequences of each dataset. Our method achieves excellent reconstruction quality, mask consistency and parameter estimation, while the baselines perform worse or do not identify those quantities, which is indicated by ``-``.}
    \label{tab:numbers_synthetic_pendulum}
\end{table*}
In contrast to the approaches of \cite{DBLP:conf/iclr/JaquesBH20} and \cite{DBLP:conf/nips/ZhongL20}, we also tackle high-resolution videos with complex background and textured objects with our approach, see \Cref{fig:results_synthetic_pendulum}.
To analyze our method, we created several synthetic videos by simulating a pendulum motion with known parameters and then rendered the images of 3 different pendulums on top of each of 3 different images, creating 9 sequences per background image. The simulated sequences allow us to compare against groundtruth parameters and object masks. We select 15 frames for training and use 26 frames for evaluation. The latter frames are selected both between training frames as well as after the training interval.

To show the advantage of explicitly modelling the physical dynamics, we compare against two baselines. First, we augment the background representation by an additional input for positional-encoded time (``Baseline-t''). This  gives a simple representation for a dynamic scene without any local representations. Second, we follow the idea from \cite{DBLP:conf/cvpr/YuanLSL21} and use a blending of background and foreground representation, where we position the foreground by learnable $\text{SE}(2)$ transformations for each training frame (``Baseline-p''). To obtain time continuous transformations, we interpolate linearly between the poses estimated for the frames. 

Qualitative results for a single scene can be seen in \Cref{fig:results_synthetic_pendulum}, \Cref{tab:numbers_synthetic_pendulum} shows a quantitative evaluation over all sequences. For more results we refer to the appendix. We see that our model produces photorealistic renderings of the scene, even for the predicted frames. While both baselines yield similar results on the training frames, the quality of the prediction on the test frames reduces for both methods. As can be seen in \cref{fig:results_synthetic_pendulum}, the time dependent background effectively blends between the training images, which means that for unseen time instances, the two pendulum positions from the neighboring training frames can be seen in the blending process. While the posed baseline does not suffer from such effects, the linear interpolation of the poses does not reflect the physical process well, and therefore the prediction quality reduces, as can be seen in \cref{tab:numbers_synthetic_pendulum}. While the time dependent baseline shows undefined behavior for the prediction in the future, it is not even clear how to extrapolate the posed baseline beyond the training interval (and therefore we did not include such frames in the evaluation for this method). In contrast, our method shows physically correct prediction between the training frames and, due to the parametric physical model, is also able to make accurate predictions for future observations. We also would like to point out, that the results show, that our methods allows accurate object segmentation for the given physical systems.

\subsection{Real World Data}
\begin{table}[]
    \centering
    \footnotesize
    \begin{tabular}{c|cc|cc|cc}
& \multicolumn{2}{c}{Pendulum} & \multicolumn{2}{c}{Sliding Block} & \multicolumn{2}{c}{Ball} \\
& \scriptsize PSNR$\uparrow$ & \scriptsize $\Delta H$ & \scriptsize PSNR$\uparrow$ & \scriptsize $\Delta H$ & \scriptsize PSNR$\uparrow$ & \scriptsize $\Delta H$\\
\hline\hline

{ w/o hom.} & %
32.74 & -  %
&
35.34 & - %
&
29.47 & -  %
\\
Full &
\textbf{32.91} & 0.07  %
&
\textbf{36.57} & 0.18 %
&
\textbf{31.74} & 0.29   %
\\
\end{tabular}

\vspace{-0.1cm}
    \caption{Reconstruction quality for the real world examples. The PSNR is averaged over all unseen test frames. We also show an ablation of the homography and report the Frobenius norm of the difference between the estimated homography matrix and a unit matrix ($\Delta H$). The results show that the homography does improve the reconstruction.
    }
    \label{tab:RealWorldNumbers}
\end{table}
\begin{figure}
    \centering
    \newcommand\WidthIms{2.5cm}%
\newcommand\Raiseheight{0.03\textwidth}%
\newcommand\distanceRows{-0.4cm}
\setlength\tabcolsep{2pt}%
\begin{tabular}{ccc}%
\rotatebox{90}{~~~Ours}~~\subfloat{\includegraphics[width=\WidthIms]{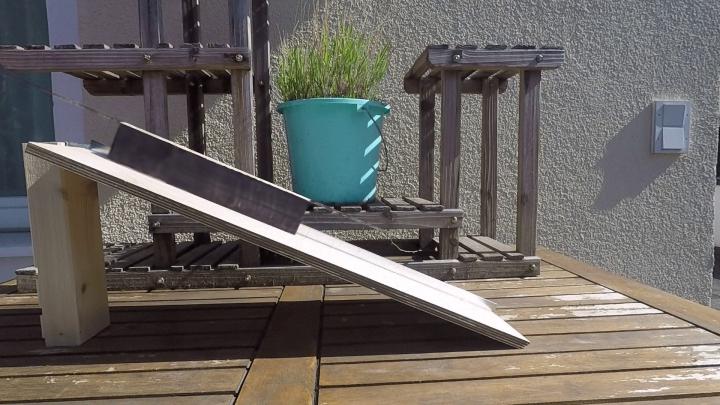}} &
\subfloat{\includegraphics[width=\WidthIms]{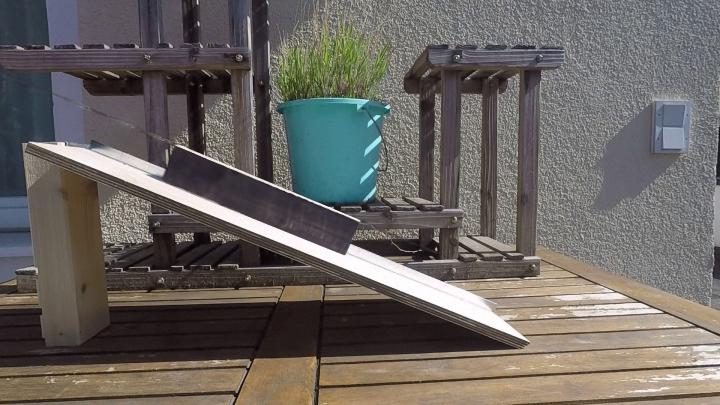}} &
\subfloat{\includegraphics[width=\WidthIms]{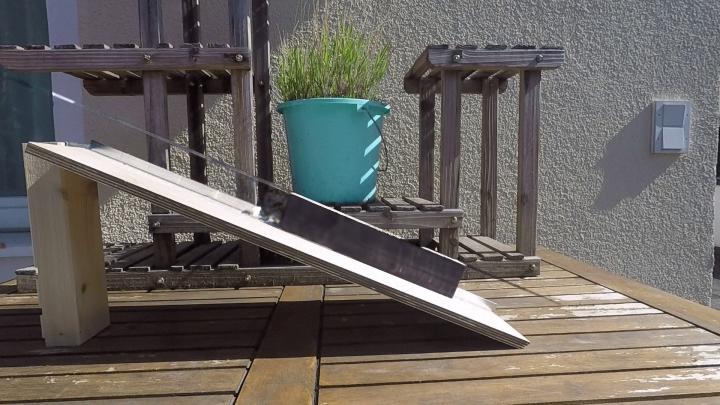}} \vspace{\distanceRows} \\
\rotatebox{90}{~~~~~GT}~~\subfloat{\includegraphics[width=\WidthIms]{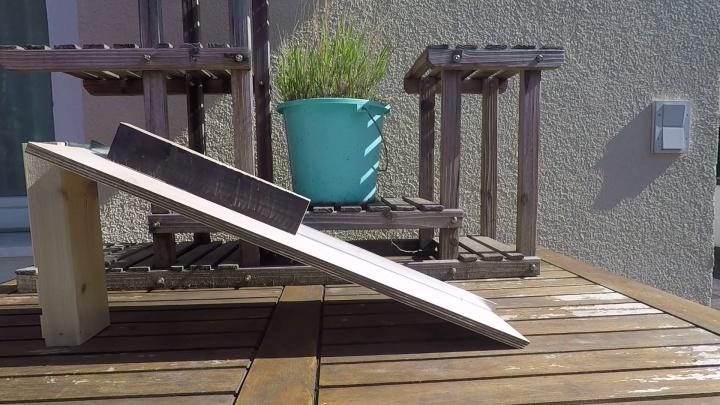}} &
\subfloat{\includegraphics[width=\WidthIms]{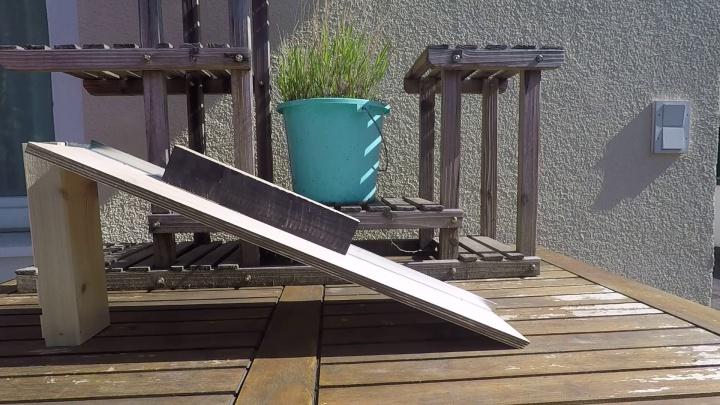}} &
\subfloat{\includegraphics[width=\WidthIms]{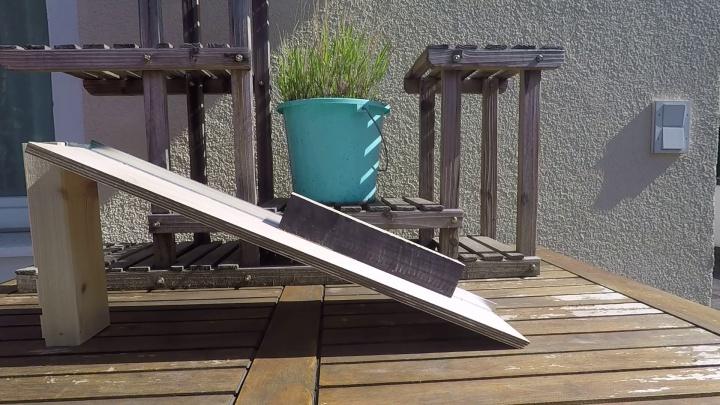}} \vspace{5pt}\\ [-1ex]
\hline \\[-3.9ex]
\rotatebox{90}{~~~Ours}~~\subfloat{\includegraphics[width=\WidthIms]{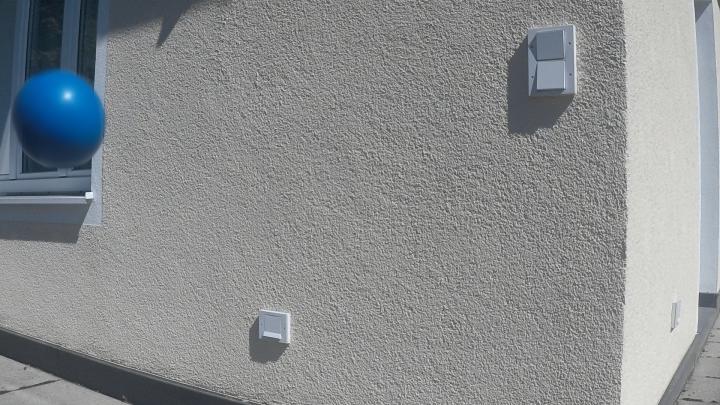}} &
\subfloat{\includegraphics[width=\WidthIms]{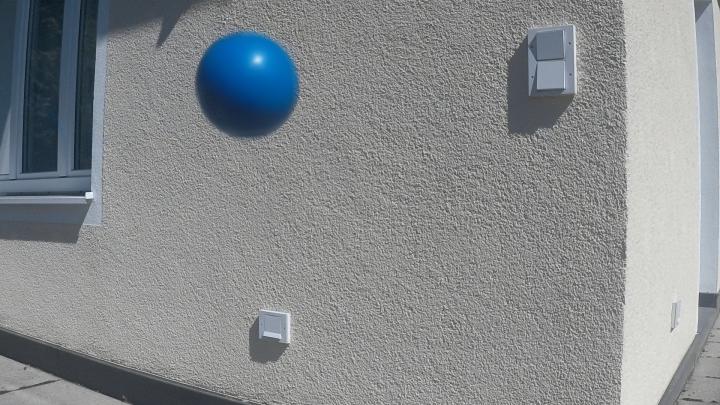}} &
\subfloat{\includegraphics[width=\WidthIms]{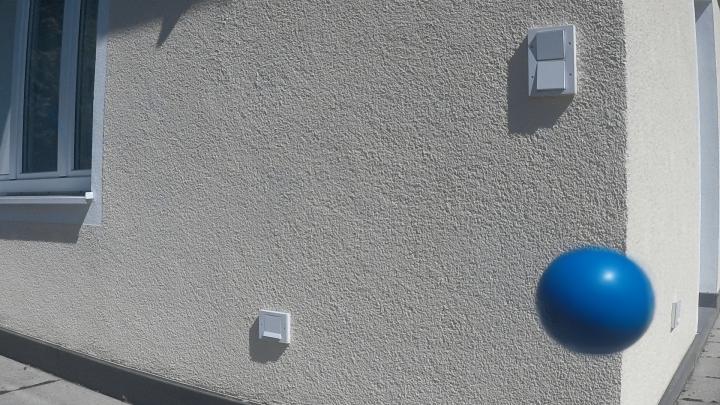}} \vspace{\distanceRows}\\
\rotatebox{90}{~~~~~GT}~~\subfloat{\includegraphics[width=\WidthIms]{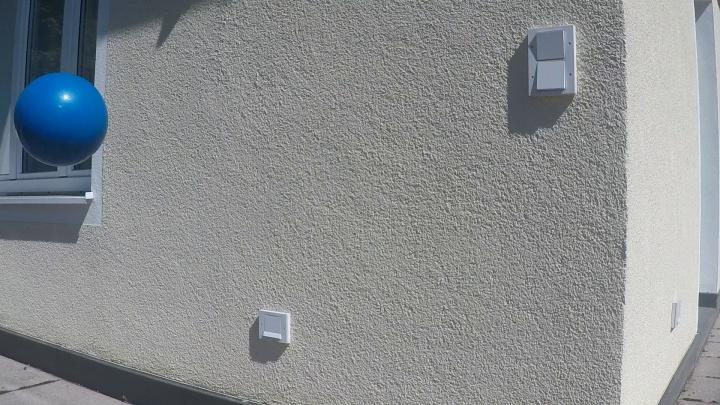}} &
\subfloat{\includegraphics[width=\WidthIms]{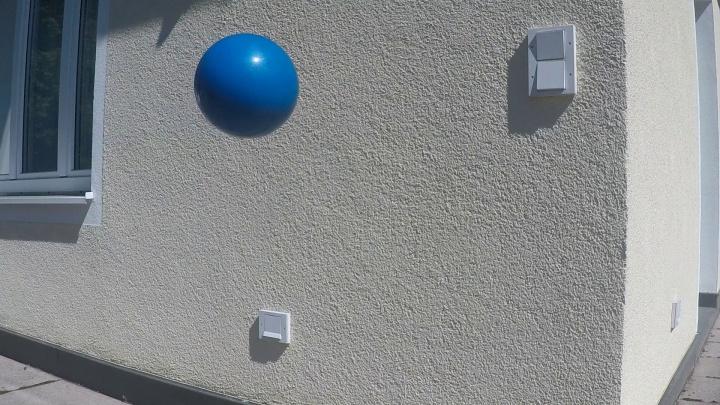}} &
\subfloat{\includegraphics[width=\WidthIms]{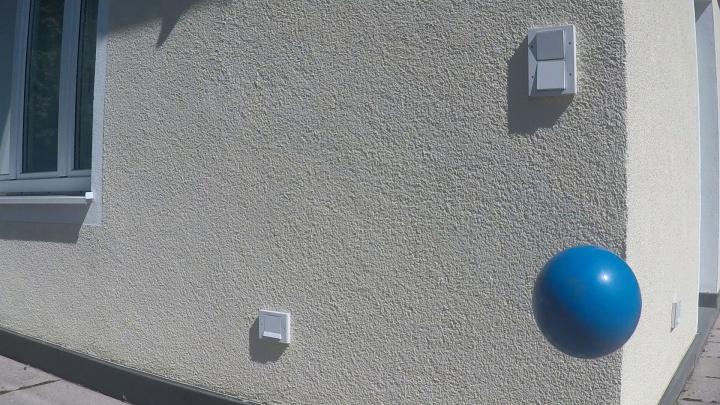}} \\
\end{tabular}

\vspace{-0.2cm}
    \caption{Reconstruction results for the sliding block and the thrown ball on three test frames. Our method produces realistic predictions for previously unseen frames, confirming that the physical parameters have been identified well.%
    \vspace{-0.05cm}
    }
    \label{fig:renderingsRealWorld}
\end{figure}
To show the capabilities of our approach on real world data, we captured videos of three physical systems: A block sliding on an inclined plane, a thrown ball, see \cref{fig:renderingsRealWorld}, and a pendulum, see \cref{fig:realWorldPendulumMotion}. For the block, the initial position and velocity, the angle of the plane and the coefficient of friction are the unknown parameters. For the ball, the initial position and velocity, need to be identified. We use the model for the damped pendulum introduced earlier. See the appendix for the full dynamics models.

The real world data is more challenging than the synthetic data, due to image noise and motion blur. We employ the homography to account for a plane of movement that is not parallel to the image plane. For training, we extract a subset of the frames and evaluate on the remaining frames.%

\Cref{tab:RealWorldNumbers} shows, that we achieve very good reconstruction on previously unseen frames, which also confirms, that the physical parameters have been well identified. While groundtruth for most of the parameters is hard to acquire, the length of the pendulum, and the angle of the inclined plane are quantities that can be obtained using a ruler. 
The estimated quantities deviate from our measured values by $4.1\%$ and $3.6\%$, respectively (relative errors). We would like to emphasize, that this shows, that for certain physical phenomena, we are able to estimate real world scale in a monocular video. To show the effectiveness of using the homography, we ablate it and report the results in \cref{tab:RealWorldNumbers}.

\section{Conclusion}
In this work we presented a solution for identifying the parameters of a physical model from a video while also creating a photorealistic representation of the appearance of the scene objects. To this end, we proposed to combine neural implicit representations and neural ODEs in an analysis-by-synthesis fashion.
Unlike existing learning-based approaches that require large training corpora, a single video clip is sufficient for our approach. In contrast to prior works that use encoder-decoder architectures specifically tailored to 2D images, we build upon neural implicit representations that have been shown to give impressive results for 3D scene reconstruction. Therefore, the extension of the proposed method to 3D is a promising direction for future work.

We present diverse experiments in which the ODE parametrizes a rigid-body transformation of the foreground objects. We emphasize that conceptually our model is not limited to rigid-body motions, and that it can directly be extended to other cases, for example to nonlinear transformations for modelling soft-body dynamics.
The focus of this work is on learning a physical model of a phenomenon from a short video. Yet, the high fidelity of our model’s renderings, together with the easy modifiability of the physical parameters, enables various computer graphics applications such as the artistic re-rendering of scenes, which we demonstrate in our video.
Overall, our per-scene model combines a unique set of favorable properties, including the interpretability of physical parameters, the ability to perform long-term predictions, and the synthesis of high-resolution images. We believe that our work may serve as inspiration for follow-up works on physics-based machine learning using neural implicit representations.

\vspace{-0.2cm}
\paragraph{Acknowledgements}
{\small This work was supported by the ERC Advanced Grant SIMULACRON, by the DFG Forschergruppe 2987 ``Learning and Simulation in Visual Computing'', 
by the CRC ``Discretization in Geometry and Dynamics''
and by the Deutsche Forschungsgemeinschaft (DFG - German Research Foundation) - Project-ID 195170736 - TRR109.}

{\small
\bibliographystyle{ieee_fullname}
\bibliography{references}
}

\appendix
\onecolumn

\section*{Appendix}
In this section we give further details on the architecture and the dynamics models (\cref{sec:ModelDetails}) as well as on the training procedure (\cref{sec:trainingDetails}) to ensure reproducibility of the work. Moreover, we indicate chosen parameter values for all experiments in \cref{sec:additional_results_appendix}, where we also show additional results and figures for all experiments. This section also includes details on the additional loss terms for the spring example (\cref{sec:resultsMassSpring}) as well as the analysis on the generalization ability of the Lagrangian variational autoencoder (\cref{sec:appendix_lagrangian_vae}). Finally, we discuss potential negative societal impact in \cref{sec:negativeSocietalImpact}.

\section{Model Details}
\label{sec:ModelDetails}
\subsection{Architecture Background and Object Representation}
\begin{figure}[]
    \centering
    \includegraphics[width=\textwidth]{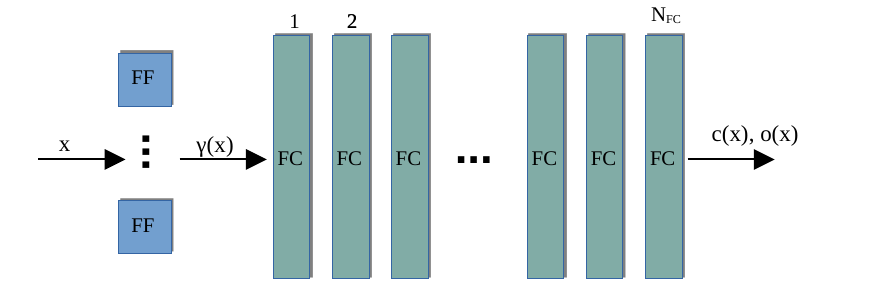}
    \caption{Overview of our architecture for the implicit shape and appearance representations. The input vector $\mathbf{x}$ is passed through a layer of $N_\text{Fourier}$ Fourier features (FF) to obtain the encoding $\gamma(\mathbf{x})$. The following neural network is constructed from $N_\text{FC}$ fully connected layers (FC) of width $W_\text{FC}$ with ReLU activations between the layers. We feed the output of the last layer through a sigmoid function, to achieve values for the color $c$ and the opacity $o$ (only for the local representation) in the range $[0,1]$. We indicate the values chosen for each experiment in the respective sections.}
    \label{fig:architecture}
\end{figure}
We adopt the architecture used in \cite{DBLP:conf/eccv/MildenhallSTBRN20} for both the representation of the background as well as the representations of the objects. See \Cref{fig:architecture} for the basic structure. Since the skip connection did not seem to give a noticeable benefit in our case we did not include it. We follow \cite{DBLP:conf/nips/TancikSMFRSRBN20} to obtain the Fourier mapping for $\mathbf{x}\in\mathbf{R}^d$ as
\begin{equation}
    \gamma(\mathbf{x}) = [\cos(2\pi \mathbf{B}\mathbf{x}), \sin(2\pi \mathbf{B}\mathbf{x})]^\top,
\end{equation}
where $\mathbf{B}\in\mathbb{R}^{N_{Fourier}\times d}\sim \mathcal{N}(0, \sigma^2)$ is sampled from a Gaussian distribution and $\sigma\in\mathbf{R}$ is a hyperparameter that is chosen for each scene. We state the values chosen for each experiment in the respective sections.

\subsection{Modeling the dynamics}
\paragraph{Two Masses Spring system}\label{sec:2_mass_spring_dynamics_model}
The system is modeled as two-body system where the dynamic of each object is described by Newton's second law of motion, i.e. $F=m\Ddot{x}$, where $F$ is the force. Since only the ratio between force and mass can be identified without additional measurement, we fix $m=1$, analogously to the work of \cite{DBLP:conf/iclr/JaquesBH20}. Using Hooke's law, we write the force applied to object $i$ by object $j$ as
\begin{equation}
    F_{i,j} =  -k\left((p_i - p_j) - 2l\frac{p_i-p_j}{\left\|p_i-p_j\right\|}\right),
\end{equation}
where $k>0$ is the spring constant and $l>0$ is the equilibrium distance. Using the position $p_i(t; k, l)\in\mathbb{R}^2$ of the objects to parametrize the trajectory of two local coordinate systems, we can write the time-dependent 2D spatial transformation to the local coordinate system $i$ as $T_t^{(i)}(x)=x-p_i(t; k, l)$. Besides the initial positions and velocities, $l$ and $k$ are learnable parameters.

\paragraph{Nonlinear damped pendulum}\label{sec:pendulum_dynamics_model}
The dynamics of a damped pendulum can be modelled as
\begin{equation}
    \dot{\begin{bmatrix}\varphi \\ \omega \end{bmatrix}} = \begin{bmatrix} \omega \\ -\frac{g}{l}\sin\left(\varphi\right) - c\omega\end{bmatrix},
    \label{eq:odePendulum}
\end{equation}
where $\varphi\in\mathbb{R}$ is the deflection angle, $\omega\in\mathbb{R}$ is the angular velocity, $g$ is the (known) gravitational acceleration, $l>0$ is the (physical) length of the pendulum, and $c>0$ is the damping constant. For the sake of simplicity we assume that the gravitational acceleration $g$ always points downwards in the global image coordinate system. We use the solution curve $\varphi\left(t;l, c\right)$ to parameterize the time-dependent 2D spatial transformation as $T_t\left(x\right) = R\left(\varphi\left(t;l, c\right)\right)x + A$, where $R\in \text{SO}\left(2\right)$ is a rotation matrix and $A\in\mathbb{R}^2$ is the pivot point of the pendulum. For the full model $A, l, c$ as well as the initial angle and angular velocity are learnable parameters.

\paragraph{Sliding block}
We model the sliding block using Newton's second law and gravity that is pointing downward in the global image coordinate system. We model the dynamics as a 1D movement along the inclined plane. Using a friction term with the friction coefficient $\mu>0$, the ODE for a block on a plane inclined by $\alpha$ reads
\begin{equation}
    \dot{\begin{bmatrix}x \\ v \end{bmatrix}} = \begin{bmatrix} v \\ g(\sin(\alpha) - \mu\cos(\alpha)\end{bmatrix},
\end{equation}
where $x\in\mathbb{R}$ is the position along the inclined plane, $v\in\mathbb{R}$ is the velocity in this direction and $g$ is again the gravitational acceleration.

\paragraph{Thrown ball}
We model a thrown object using again Newton's law where only gravity is acting on the object. We assume again, that gravity is pointing downwards in the global image coordinate system. The ODE describing the resulting 2D motion reads
\begin{equation}
    \dot{\begin{bmatrix}x \\ y \\ v_x \\ v_y \end{bmatrix}} = \begin{bmatrix} v_x \\ v_y \\ 0 \\ g\end{bmatrix},
\end{equation}
where $x$ and $y$ are the positions in the image coordinate system, $v_x$ and $v_y$ are the velocities in the respective directions and $g$ is the gravitational acceleration.

\section{Additional training details}
\label{sec:trainingDetails}
\subsection{Optimization}
We train our model using the Adam optimizer \cite{DBLP:journals/corr/KingmaB14} with exponential learning rate decay, which reads
\begin{equation}
    r(e) = r_0 \cdot \beta^{e/n_\text{decay}}
\end{equation}
where $r(e)$ is the learning rate depending on the epoch $e$, $r_0$ is the initial learning rate, $\beta$ is the decay rate and $n_\text{decay}$ is the decay step size. 

One important aspect of the training is to use different learning rates for the parameters  $\theta_{\text{bg}}$ and $\theta_{\text{obj}}$ of the implicit representations on the one hand and the physical parameters $\theta_{\text{ode}}, \mathbf{z}_0$ and $\theta_{+}$ on the other hand.

Due to the solution of the ODE, our objective function is generally non-convex and non-linear. Therefore, we rely on a good initialization for the ODE parameters and the parameters of the transformation to achieve good convergence in the optimization. In an earlier version of this work, we used object masks in addition to the images of the sequence to supervise the occupancy values by an additional loss term. While we were able to remove the masks for the supervision, we kept the previous approach to estimate initial values for position and velocities based on the masks.

For the initialization of the pendulum we estimate the pivot point $A$ by averaging all masks and use the the pixel with the highest value. Note that this approach will fail when the pivot point is not contained in the image. To obtain an estimate for the initial angle, we perform a principal component analysis (PCA) on the pixel locations covered by the mask and use the angle between the first component and the vertical direction. The angular velocity is estimated as the angular difference between the first principal components of the first and the second frame divided by the time difference. For the synthetic experiments, averaged over all 27 sequences, this leads to an initialization with a relative error of $14\%$ for the pivot point $A$ and of $40\%$ for the initial values (initial angle and angular velocity).

For the remaining systems, we initialize the initial positions at the center of the masks and the initial velocity as positional difference between the first two frames divided by the time difference. We report the initialization of the remaining parameters in the respective subsection of \cref{sec:additional_results_appendix}.

\subsection{Loss term}
We use a mean squared error photometric loss defined over all the pixel values, which reads
\begin{equation}\label{eq:photo}
    \mathcal{L}_{photometric} = \frac{1}{\left|\mathcal{I}\right|\left|\mathcal{T}\right|}\sum_{t\in\mathcal{T}}\sum_{x\in\mathcal{I}}
    \left\|I\left(\vx, t\right) - c\left(\vx, t\right)\right\|^2,
\end{equation}
where $\mathcal{T}$ is the set of all given time steps, $\mathcal{I}$ is the set of all pixel coordinates and $I\left(\vx,t\right)$ are the given images. We found, that for some backgrounds with little distinguishable details, a regularization of the mask is helpful. We use the term
\begin{equation}
    \mathcal{L}_{maskReg} = \frac{1}{\left|\mathcal{I}\right|\left|\mathcal{T}\right|}\sum_{t\in\mathcal{T}}\sum_{x\in\mathcal{I}}
    o(\vx)(1-o(\vx)),
\end{equation}
that encourages the occupancy $o$ predicted by the local representation to be either close to $1$ or close to zero $0$. To avoid ``burning'' artefacts into the masks, we activate this term after $N_{reg}$ epochs. To balance the term with the photometric loss we use a weight of $\lambda_{reg}$. This additional loss is only used for the real world examples and the high resolution synthetic data.

For the training, we randomly sample batches of up to $N_{batch}=2^{16}=65536$ pixels for each optimization iteration. We found that this large batch size has a stabilizing effect on the optimization.

\subsection{Online Training}
We adopt the approach from \cite{DBLP:conf/cvpr/YuanLSL21} and increase the number of frames used for the loss term during the optimization. Starting from $n_{fr,0}$ we increase the number of frames by $1$ every $n_{incrT}$ steps. We found that this strategy improves the convergence behavior of the approach and seems to make it more robust to the initialization of the parameters.

\section{Further experimental details and results}
\label{sec:additional_results_appendix}
In the following we consider specific details for the different experiments.
\subsection{Two Masses Spring System}
\label{sec:resultsMassSpring}
\begin{figure}
    \centering
    \includegraphics{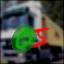}\ \ \includegraphics{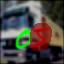}
    \caption{Coarse occupancy masks used as supervision in the first frame for the spring sequences. We use the masks shown as overlay in the image to supervise the occupancy of the respective local representation using a binary cross entropy loss. This supervision makes the assignment of the two representations to the digits unique. The weighting of the loss is reduced during training to enable the learning of the fine object structures. Note that the rough object masks are only required in the \emph{first} frame of the sequence. }
    \label{fig:spring_coarse_segmentation_masks}
\end{figure}
\paragraph{Experimental details} As described in \Cref{sec:2_mass_spring_dynamics_model}, we employ two independent local representations to model the two digits. By using the maximum of both occupancy values, we enable the model to identify the layering of the objects. Since the two local representations are not explicitly assigned to the digits, we found that we need to guide the model with an \emph{additional} loss term. We use a binary cross entropy loss on very coarse object masks in the \emph{first} frame of the sequence, see \cref{fig:spring_coarse_segmentation_masks}. The loss is initially weighted by a factor of $0.01$ compared to the photometric loss. We reduce this loss term every $100$ epochs by a factor of $0.2$ to enable the model to learn the fine structures.

The physical system appears to have a scale freedom in terms of equilibrium length and the points where the spring is attached to the digits.\footnote{Intuitively, if the motion is only in one direction (linear), we can vary the equilibrium length and adjust the spring attachments without changing the motion. Similar effects are present in particular 2D motions.} We observe similar effects when overfitting the model of \cite{DBLP:conf/iclr/JaquesBH20} to a single sequence. When training on the full dataset, the effect seems to be averaged out, and is not observed. We add an additional MSE loss to keep the spring attachment close to the origin of the local coordinate system. This loss is weighted by $0.05$. 

Finally, we use another MSE loss term to keep the opacity value close to zero outside of (but close to) the visible area. We found this to be necessary, since otherwise artefacts might appear in the prediction, when previously unseen parts of the mask appear in the visible area. This term is weighted by a factor of $1.0$.

\paragraph{Model parameters}
For the background we use an MLP with $N_\text{FC}=6$ fully connected layers of width $W_\text{FC}=64$ and a Fourier mapping with $N_\text{Fourier}=64$ Fourier features and variance $\sigma=5.0$. To represent the local objects we use $N_\text{FC}=6$ fully connected layers of width $W_\text{FC}=64$ and a Fourier mapping with $N_\text{Fourier}=64$ Fourier features and variance $\sigma=2.2$.

We use an initial learning rate of $r_\text{MLP, 0}=0.001$ for the parameters of the implicit representations and $r_\text{param, 0}=0.005$ for the physical parameters. We set $\beta_\text{MLP}=0.99954, n_\text{decay,MLP}=50$. We do not decay the learning rate for the physical parameters.

For the online training scheme, we start with $n_{fr,0}=2$ frames and increase the number of frames by one every $n_{incrT}=30$ steps. We train for $1200$ epochs, where one epoch is completed, when all the pixels have been considered.

The initial spring constant is set to $k=1.5$ and the equilibrium distance is initialized as the distance between the estimates of the initial positions.

\begin{figure}
    \centering
    \newcommand\WidthIms{1.35cm}
\newcommand\Raiseheight{0.03\textwidth}
\definecolor{plotRed}{rgb}{0.85000,0.32500,0.09800}
\setlength\tabcolsep{1 pt}
\begin{tabular}{cccccccccc}%
\footnotesize%
\rotatebox{90}{~B: Overfit}~\subfloat{\includegraphics[width=\WidthIms]{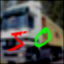}} &
\subfloat{\includegraphics[width=\WidthIms]{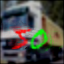}}&%
\subfloat{\includegraphics[width=\WidthIms]{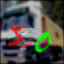}} &
\subfloat{\includegraphics[width=\WidthIms]{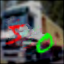}} &
\subfloat{\includegraphics[width=\WidthIms]{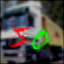}} &
\subfloat{\includegraphics[width=\WidthIms]{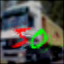}} &
\subfloat{\includegraphics[width=\WidthIms]{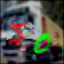}} &
\subfloat{\includegraphics[width=\WidthIms]{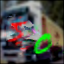}} &
\subfloat{\includegraphics[width=\WidthIms]{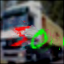}} \\[-2.5ex]
\rotatebox{90}{~B: Full}~\subfloat{\includegraphics[width=\WidthIms]{exampleSpring/Sequence0_paig_Overfit/00_recon.jpg}} &
\subfloat{\includegraphics[width=\WidthIms]{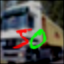}}&%
\subfloat{\includegraphics[width=\WidthIms]{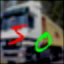}} &
\subfloat{\includegraphics[width=\WidthIms]{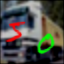}} &
\subfloat{\includegraphics[width=\WidthIms]{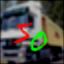}} &
\subfloat{\includegraphics[width=\WidthIms]{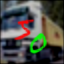}} &
\subfloat{\includegraphics[width=\WidthIms]{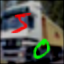}} &
\subfloat{\includegraphics[width=\WidthIms]{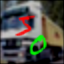}} &
\subfloat{\includegraphics[width=\WidthIms]{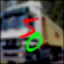}} \\[-2.5ex]
\rotatebox{90}{~~~~Ours}~\subfloat{\includegraphics[width=\WidthIms]{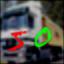}} &
\subfloat{\includegraphics[width=\WidthIms]{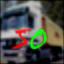}} &
\subfloat{\includegraphics[width=\WidthIms]{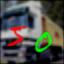}} &
\subfloat{\includegraphics[width=\WidthIms]{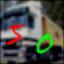}} &
\subfloat{\includegraphics[width=\WidthIms]{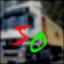}} &
\subfloat{\includegraphics[width=\WidthIms]{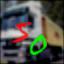}} &
\subfloat{\includegraphics[width=\WidthIms]{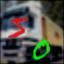}} &
\subfloat{\includegraphics[width=\WidthIms]{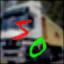}} &
\subfloat{\includegraphics[width=\WidthIms]{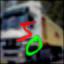}}  \\[-2.5ex]
\rotatebox{90}{~~~~~~GT}~\subfloat{\includegraphics[width=\WidthIms]{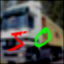}} &
\subfloat{\includegraphics[width=\WidthIms]{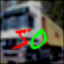}}&%
\subfloat{\includegraphics[width=\WidthIms]{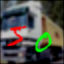}} &
\subfloat{\includegraphics[width=\WidthIms]{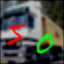}} &
\subfloat{\includegraphics[width=\WidthIms]{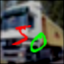}} &
\subfloat{\includegraphics[width=\WidthIms]{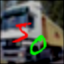}} &
\subfloat{\includegraphics[width=\WidthIms]{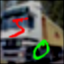}} &
\subfloat{\includegraphics[width=\WidthIms]{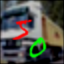}} &
\subfloat{\includegraphics[width=\WidthIms]{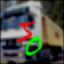}} \\
{1} &
{4} &
{7} &
{10} &
{13} &
{16} &
{19} &
{22} &
{25} \\
 &
 &
 &
 &
\begin{tikzpicture} \draw [-stealth, plotRed](0,0) -- (1,0);\end{tikzpicture}&
 &
 &
 &
\\
\end{tabular}

    \caption{Two masses spring system, where MNIST digits are connected by an (invisible) spring. Reconstruction and prediction for test sequence 0. The arrow indicates where the prediciton starts. For the spring constant and equilibrium distance ($k$, $l$) the different methods achieve the following relative errors respectively: $(19.7 \%,~ 57.6\%)$ (B: Overfit), $(3.7 \%,~1.8 \%)$ (B: Full), and $(\mathbf{0.3}\%,~ \mathbf{0.7}\%)$ (Ours).}
    \label{fig:mass_spring_seq_0}
\end{figure}

\begin{figure}
    \centering
    \newcommand\WidthIms{1.35cm}
\newcommand\Raiseheight{0.03\textwidth}
\definecolor{plotRed}{rgb}{0.85000,0.32500,0.09800}
\setlength\tabcolsep{1 pt}
\begin{tabular}{cccccccccc}%
\footnotesize%
\rotatebox{90}{~B: Overfit}~\subfloat{\includegraphics[width=\WidthIms]{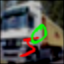}} &
\subfloat{\includegraphics[width=\WidthIms]{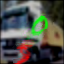}}&%
\subfloat{\includegraphics[width=\WidthIms]{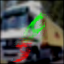}} &
\subfloat{\includegraphics[width=\WidthIms]{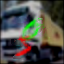}} &
\subfloat{\includegraphics[width=\WidthIms]{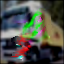}} &
\subfloat{\includegraphics[width=\WidthIms]{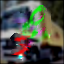}} &
\subfloat{\includegraphics[width=\WidthIms]{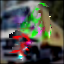}} &
\subfloat{\includegraphics[width=\WidthIms]{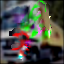}} &
\subfloat{\includegraphics[width=\WidthIms]{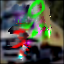}} \\[-2.5ex]
\rotatebox{90}{~B: Full}~\subfloat{\includegraphics[width=\WidthIms]{exampleSpring/Sequence1_paig_Overfit/00_recon.jpg}} &
\subfloat{\includegraphics[width=\WidthIms]{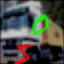}}&%
\subfloat{\includegraphics[width=\WidthIms]{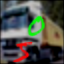}} &
\subfloat{\includegraphics[width=\WidthIms]{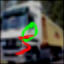}} &
\subfloat{\includegraphics[width=\WidthIms]{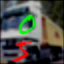}} &
\subfloat{\includegraphics[width=\WidthIms]{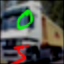}} &
\subfloat{\includegraphics[width=\WidthIms]{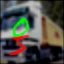}} &
\subfloat{\includegraphics[width=\WidthIms]{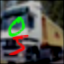}} &
\subfloat{\includegraphics[width=\WidthIms]{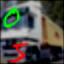}} \\[-2.5ex]
\rotatebox{90}{~~~~Ours}~\subfloat{\includegraphics[width=\WidthIms]{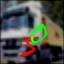}} &
\subfloat{\includegraphics[width=\WidthIms]{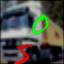}} &
\subfloat{\includegraphics[width=\WidthIms]{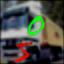}} &
\subfloat{\includegraphics[width=\WidthIms]{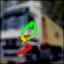}} &
\subfloat{\includegraphics[width=\WidthIms]{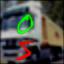}} &
\subfloat{\includegraphics[width=\WidthIms]{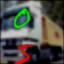}} &
\subfloat{\includegraphics[width=\WidthIms]{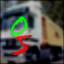}} &
\subfloat{\includegraphics[width=\WidthIms]{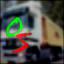}} &
\subfloat{\includegraphics[width=\WidthIms]{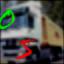}}  \\[-2.5ex]
\rotatebox{90}{~~~~~~GT}~\subfloat{\includegraphics[width=\WidthIms]{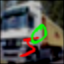}} &
\subfloat{\includegraphics[width=\WidthIms]{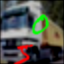}}&%
\subfloat{\includegraphics[width=\WidthIms]{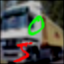}} &
\subfloat{\includegraphics[width=\WidthIms]{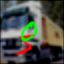}} &
\subfloat{\includegraphics[width=\WidthIms]{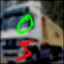}} &
\subfloat{\includegraphics[width=\WidthIms]{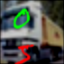}} &
\subfloat{\includegraphics[width=\WidthIms]{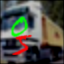}} &
\subfloat{\includegraphics[width=\WidthIms]{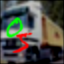}} &
\subfloat{\includegraphics[width=\WidthIms]{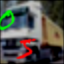}} \\
{1} &
{4} &
{7} &
{10} &
{13} &
{16} &
{19} &
{22} &
{25} \\
 &
 &
 &
 &
\begin{tikzpicture} \draw [-stealth, plotRed](0,0) -- (1,0);\end{tikzpicture}&
 &
 &
 &
\\
\end{tabular}

    \caption{Two masses spring system, where MNIST digits are connected by an (invisible) spring. Reconstruction and prediction for test sequence 1. The arrow indicates where the prediciton starts. For the spring constant and equilibrium distance ($k$, $l$) the different methods achieve the following relative errors respectively: $(13.6\%,~ 90.9\%)$ (B: Overfit), $(3.7 \%,~\mathbf{1.8} \%)$ (B: Full), and $(\mathbf{0.1}\%, ~ 4.9\%)$ (Ours}
    \label{fig:mass_spring_seq_1}
\end{figure}

\paragraph{Additional results} In \Cref{fig:mass_spring_seq_0} and \Cref{fig:mass_spring_seq_1} we present additional results for sequence 0 and sequence 1 of the test dataset. We see, that for both sequences, overfitting the baseline is not able to produce a reasonable extrapolation of the data and even produces severe artifacts for the reconstruction part of the sequence. One reason for this is that the model is unable to identify the physical parameters correctly as can be seen by the large relative errors. Our model, on the other hand, is able to estimate the parameters with high accuracy that is even slightly better than the baseline trained on the full training dataset, which again shows the strength of our approach, considering, that we use a single video as input.

\subsection{Comparison with the Lagrangian Variational Autoencoder}
\label{sec:appendix_lagrangian_vae}
\paragraph{Experimental details} Since the data used in this experiment does not include image data, we use a binary cross entropy loss to penalize the discrepancy between the given object masks and our rendered occupancy values. Since the predicted masks are obtained only from the local representation, we do not use an implicit representation for the background in this example.

\paragraph{Model parameters}
For the local representation we use an MLP with $N_\text{FC}=6$ fully connected layers of width $W_\text{FC}=64$ and a Fourier mapping with $N_\text{Fourier}=64$ Fourier features and variance $\sigma=0.1$.

We use an initial learning rate of $r_\text{MLP, 0}=0.005$ for the parameters of the implicit representations and $r_\text{param, 0}=0.01$ for the physical parameters. We do not use any learning rate decay in this example.

For the online training scheme, we start with $n_{fr,0}=5$ and increase the number of frames every $n_{incrT}=20$ steps. We train for $2000$ epochs, where one epoch is completed, when all the pixels have been considered.

We initialize the damping as $c=0.25$ and the pendulum length as $1.5$.

\paragraph{Generalization of the Lagrangian Varational Autoencoder}
\label{sec:generalization_lagrangian_vae}
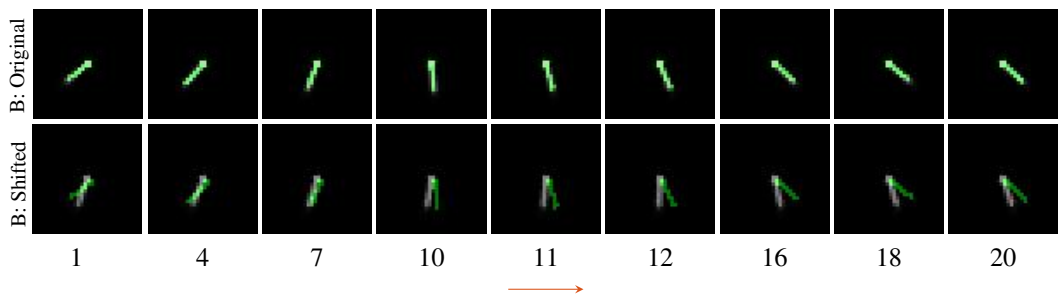
\begin{figure}[t!]
    \centering
    \newcommand\WidthIms{1.45cm}
\newcommand\Raiseheight{0.03\textwidth}
\newcommand\fontsizeCaption{\footnotesize}
\definecolor{plotRed}{rgb}{0.85000,0.32500,0.09800}
\setlength\tabcolsep{1 pt}
\begin{tabular}{cccccccccc}%
\footnotesize%
\rotatebox{90}{~\fontsizeCaption B: Original}~\subfloat{\includegraphics[width=\WidthIms]{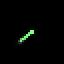}} &
\subfloat{\includegraphics[width=\WidthIms]{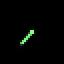}}&%
\subfloat{\includegraphics[width=\WidthIms]{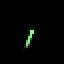}} &
\subfloat{\includegraphics[width=\WidthIms]{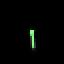}} &
\subfloat{\includegraphics[width=\WidthIms]{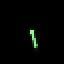}} &
\subfloat{\includegraphics[width=\WidthIms]{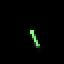}} &
\subfloat{\includegraphics[width=\WidthIms]{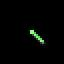}} &
\subfloat{\includegraphics[width=\WidthIms]{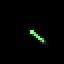}} &
\subfloat{\includegraphics[width=\WidthIms]{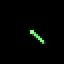}} \\[-2.5ex]
\rotatebox{90}{~\fontsizeCaption B: Shifted}~\subfloat{\includegraphics[width=\WidthIms]{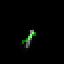}} &
\subfloat{\includegraphics[width=\WidthIms]{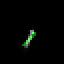}} &
\subfloat{\includegraphics[width=\WidthIms]{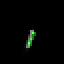}} &
\subfloat{\includegraphics[width=\WidthIms]{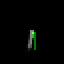}} &
\subfloat{\includegraphics[width=\WidthIms]{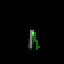}} &
\subfloat{\includegraphics[width=\WidthIms]{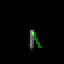}} &
\subfloat{\includegraphics[width=\WidthIms]{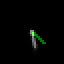}} &
\subfloat{\includegraphics[width=\WidthIms]{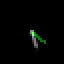}} &
\subfloat{\includegraphics[width=\WidthIms]{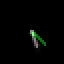}} \\
{1} &
{4} &
{7} &
{10} &
{11} &
{12} &
{16} &
{18} &
{20} \\
 &
 &
 &
 &
\raisebox{0.1cm}{\begin{tikzpicture} \draw [-stealth, plotRed](0,0) -- (1,0);\end{tikzpicture}}&
 &
 &
 &
\\
\end{tabular}

    \caption{Prediction of the fully trained model of \cite{DBLP:conf/nips/ZhongL20} for sequence $3$ of the test dataset. While the prediction for the original data is perfect, the prediction for the frames shifted by one pixel in each direction is significantly worse. This shows, that the model does not generalize well to input frames where the pivot point of the pendulum is not in the center of the frame.}
    \label{fig:LagrangianVAE_shift}
\end{figure}
One drawback of learning-based approaches for visual estimation of physical models is the poor generalization to data that deviates from the training data distribution. We confirm this for the model of \cite{DBLP:conf/nips/ZhongL20} trained on the full dataset. While the IoU averaged over the first 20 sequences of the test set is $0.73$, the value drops to $0.21$ when we shift the frames of the test dataset by as much as 1 pixel in each direction. This shift corresponds to the case of input videos, where the pivot point of the pendulum is not in the center of the image, which is different from the training data. This effect is visualized in \Cref{fig:LagrangianVAE_shift}, which shows the output of the model for sequence $3$ of the test data set with zero control input, both in the original version and in the shifted version. We observe that the small shift of only one pixel in each direction leads to results that are significantly off, and not even the first frame is predicted correctly. While \cite{DBLP:conf/nips/ZhongL20} propose to use a coordinate-aware encoder based on spatial transformers, this introduces additional complexity to the model. In contrast, our approach does not suffer from such issues, as we estimate the parameters per scene.

\subsection{Synthetic Experiments - High Resolution}

\paragraph{Model parameters}
For the background we use an MLP with $N_\text{FC}=8$ fully connected layers of width $W_\text{FC}=512$ and a Fourier mapping with $N_\text{Fourier}=256$ Fourier features and variance $\sigma=30.0$. For the representation of the local objects we use $N_\text{FC}=8$ fully connected layers of width $W_\text{FC}=128$ and a Fourier mapping with $N_\text{Fourier}=256$ Fourier features and variance $\sigma=10.0$.

We use an initial learning rate of $r_\text{MLP, 0}=9e-4$ for the parameters of the implicit representations and $r_\text{param, 0}=1e-3$ for the physical parameters. We set $\beta_\text{MLP}=0.9, n_\text{decay,MLP}=25$. We do not decay the learning rate for the physical parameters. We activate the mask regularization after $N_{reg}=400$ epochs and use $\lambda_{reg}=5e-4$ to balance the regularization with the photometric loss.

For the online training scheme, we start with $n_{fr,0}=5$ frames and increase the number of frames by one every $n_{incrT}=10$ steps. We train for $1200$ epochs, where one epoch is completed, when all the pixels have been considered.

We initialize the damping as $c=0.6$ and the pendulum length as $1.9$.

\begin{figure}
    \centering
    \newcommand\WidthIms{0.2\columnwidth}
\newcommand\Raiseheight{0.07\columnwidth}
\newcommand\spacing{\qquad\qquad\quad}
\begin{tabular}{c|cc}%
\rotatebox{90}{\spacing Ours}~
\subfloat{\includegraphics[width=\WidthIms]{syntheticPendulum/wallclock/renderingsOurs/0_train.jpg}} &
\subfloat{\includegraphics[width=\WidthIms]{syntheticPendulum/wallclock/renderingsOurs/5_test.jpg}} &
\subfloat{\includegraphics[width=\WidthIms]{syntheticPendulum/wallclock/renderingsOurs/22_test.jpg}} \\[-1.0ex]
\rotatebox{90}{\spacing GT}~
\subfloat{\includegraphics[width=\WidthIms]{syntheticPendulum/wallclock/renderingsOurs/0_gt_train.jpg}} &
\subfloat{\includegraphics[width=\WidthIms]{syntheticPendulum/wallclock/renderingsOurs/5_gt_test.jpg}} &
\subfloat{\includegraphics[width=\WidthIms]{syntheticPendulum/wallclock/renderingsOurs/22_gt_test.jpg}} \\
\hline \\[-3.9ex]
\rotatebox{90}{\spacing Ours}~
\subfloat{\includegraphics[width=\WidthIms]{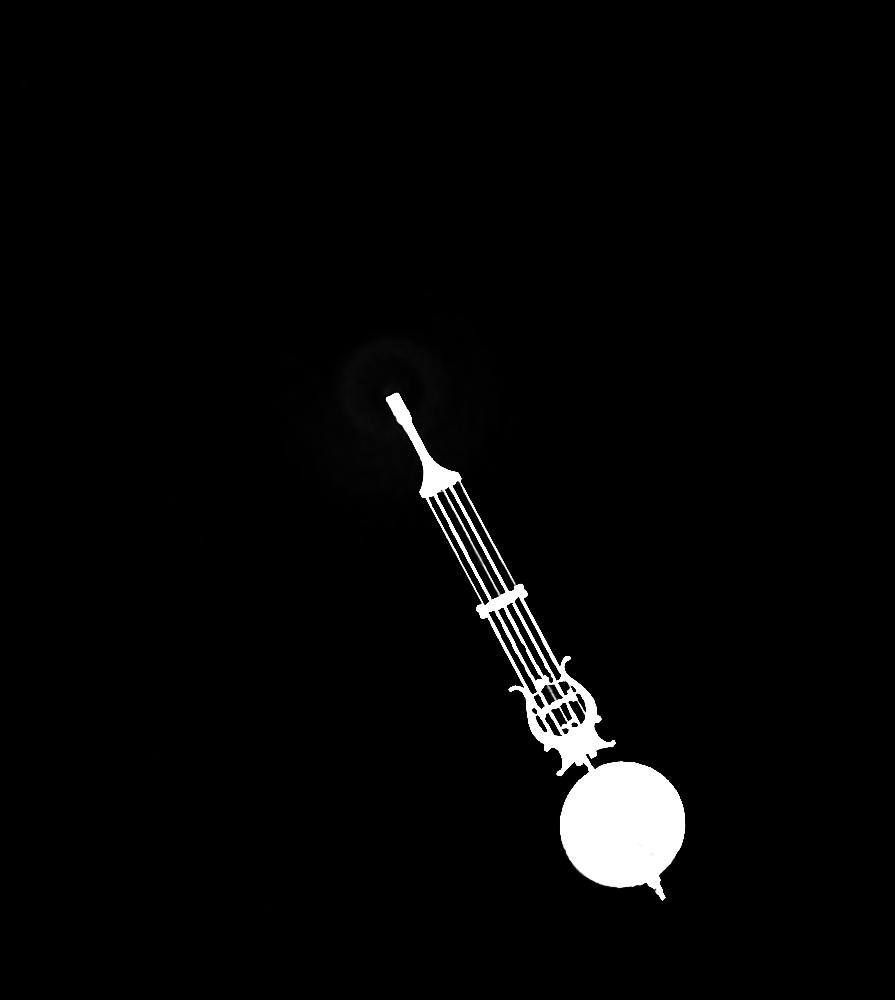}} &
\subfloat{\includegraphics[width=\WidthIms]{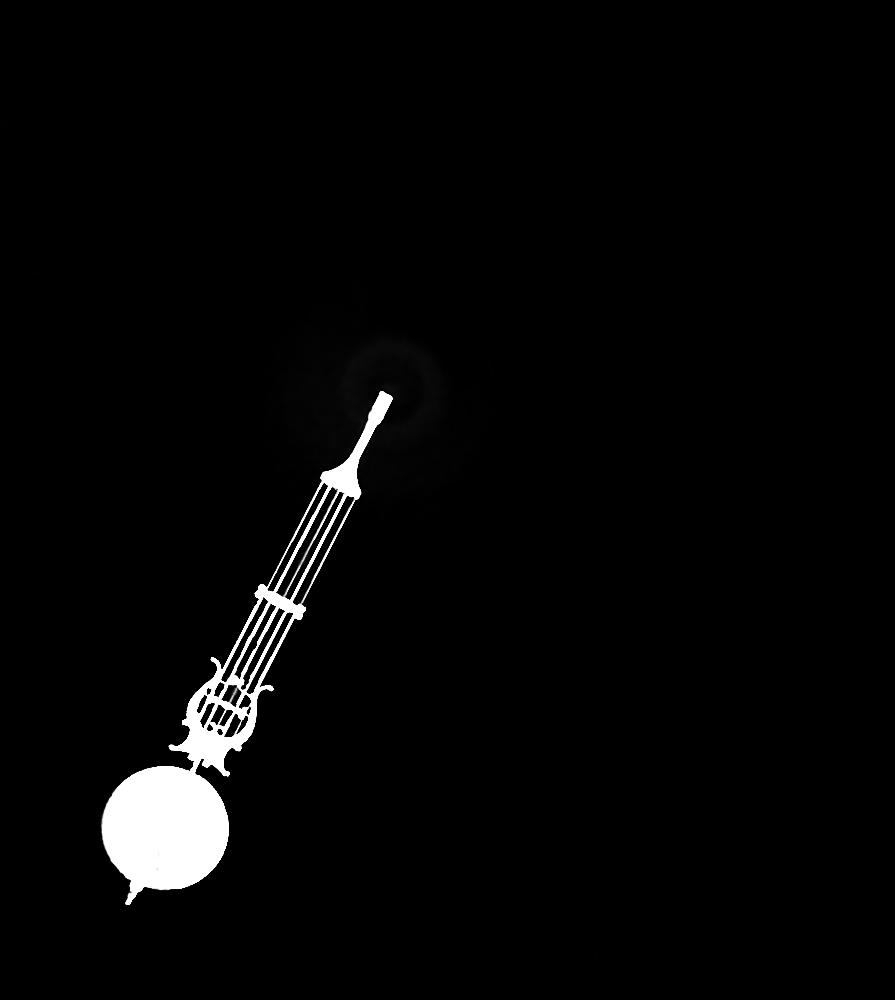}} &
\subfloat{\includegraphics[width=\WidthIms]{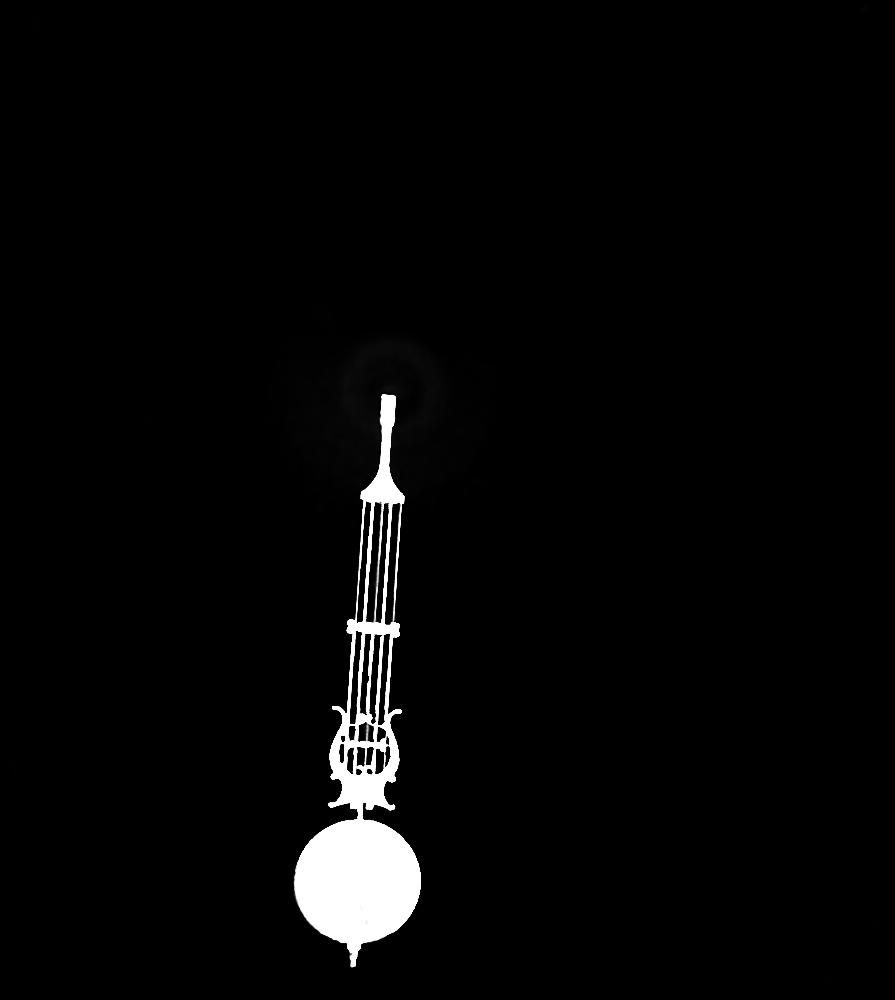}} \\[-1.0ex]
\rotatebox{90}{\spacing GT}~
\subfloat{\includegraphics[width=\WidthIms]{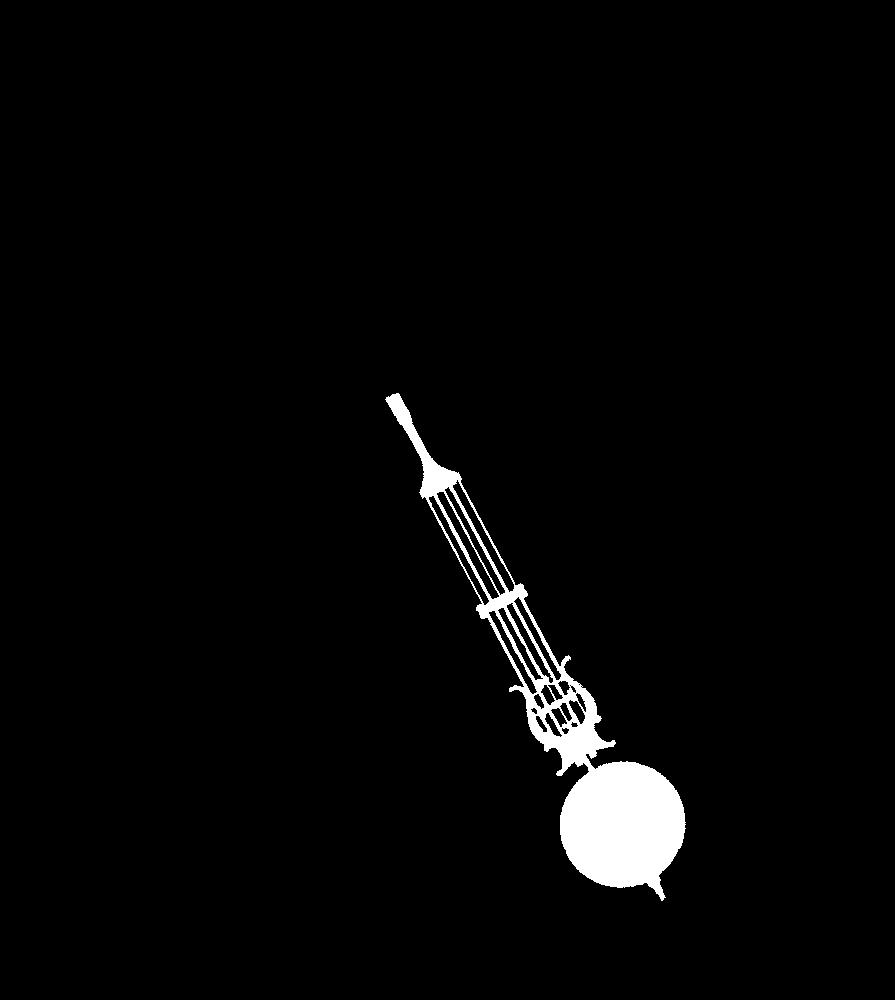}} &
\subfloat{\includegraphics[width=\WidthIms]{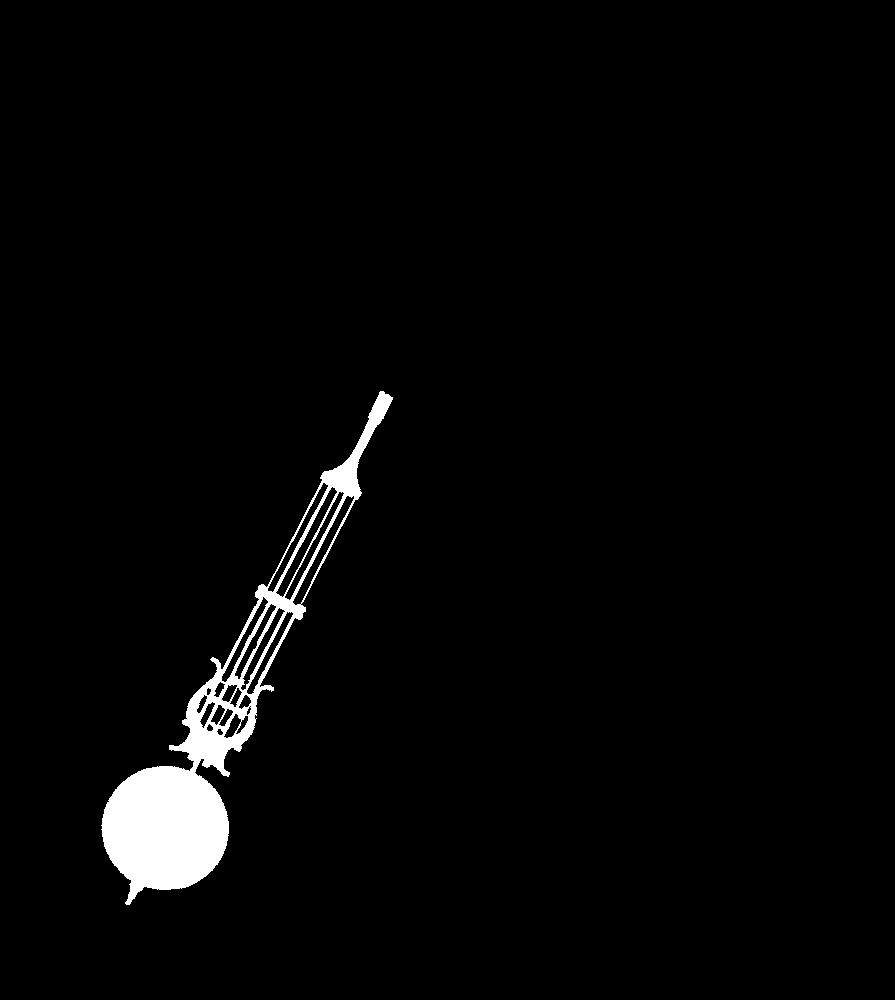}} &
\subfloat{\includegraphics[width=\WidthIms]{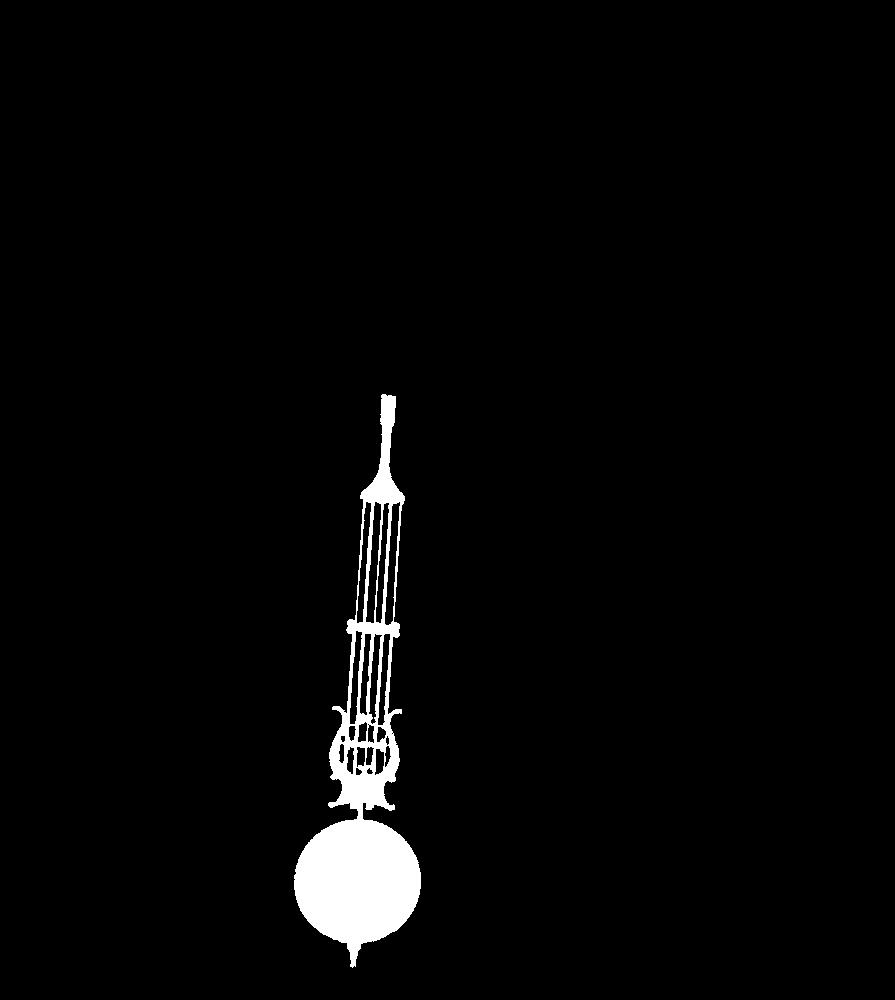}} \\
 {~training frame} & {unseen frame 1} & {unseen frame 2} \\
\end{tabular}
    \caption{Rendered frames for sequence 1 of the wallclock background (same sequence as in the main text, rendered images are shown again for convenience). The left image is part of the training set, ``unseen frame 1'' is between two training frames, ``unseen frame 2'' is a future frame after the interval seen during training. Our method produces photorealistic predictions for the unseen time instances. Also, it predicts accurate segmentation masks for the object.}
    \label{fig:resultsWallclock}
\end{figure}

\begin{figure}
    \centering
    \newcommand\WidthIms{0.27\columnwidth}
\newcommand\Raiseheight{0.07\columnwidth}
\newcommand\spacing{\qquad\qquad}
\begin{tabular}{c|cc}%
\rotatebox{90}{\spacing Ours}~
\subfloat{\includegraphics[width=\WidthIms]{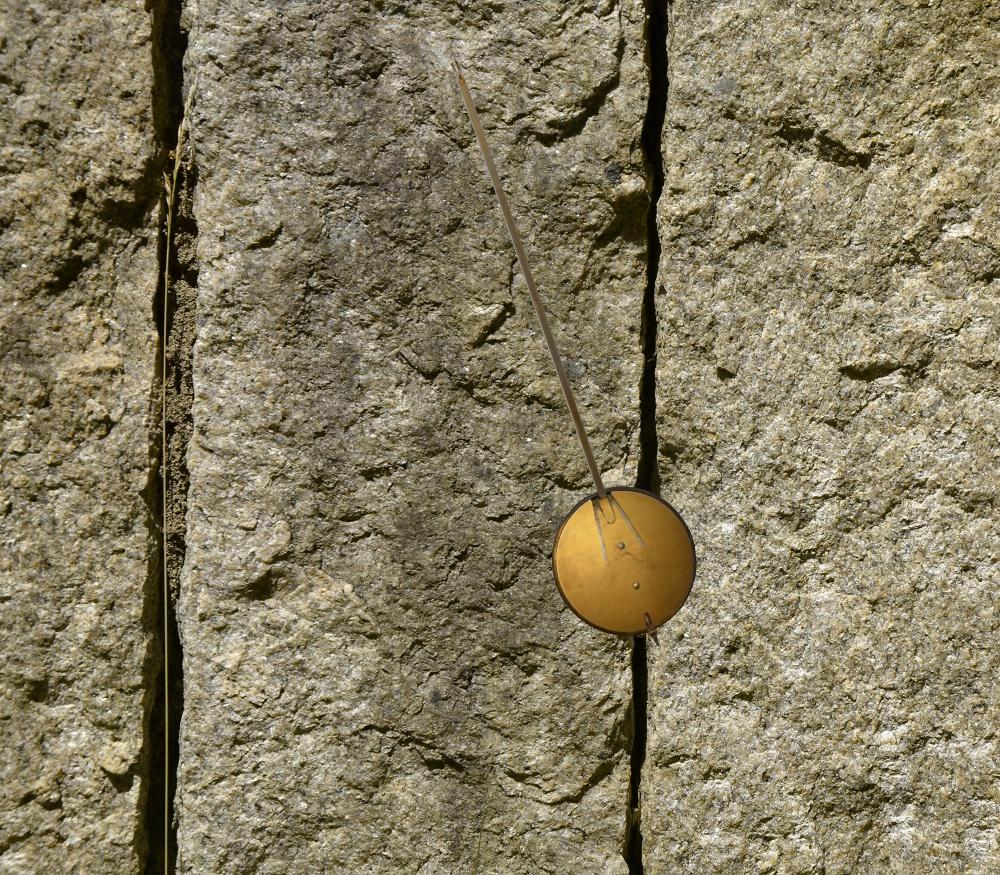}} &
\subfloat{\includegraphics[width=\WidthIms]{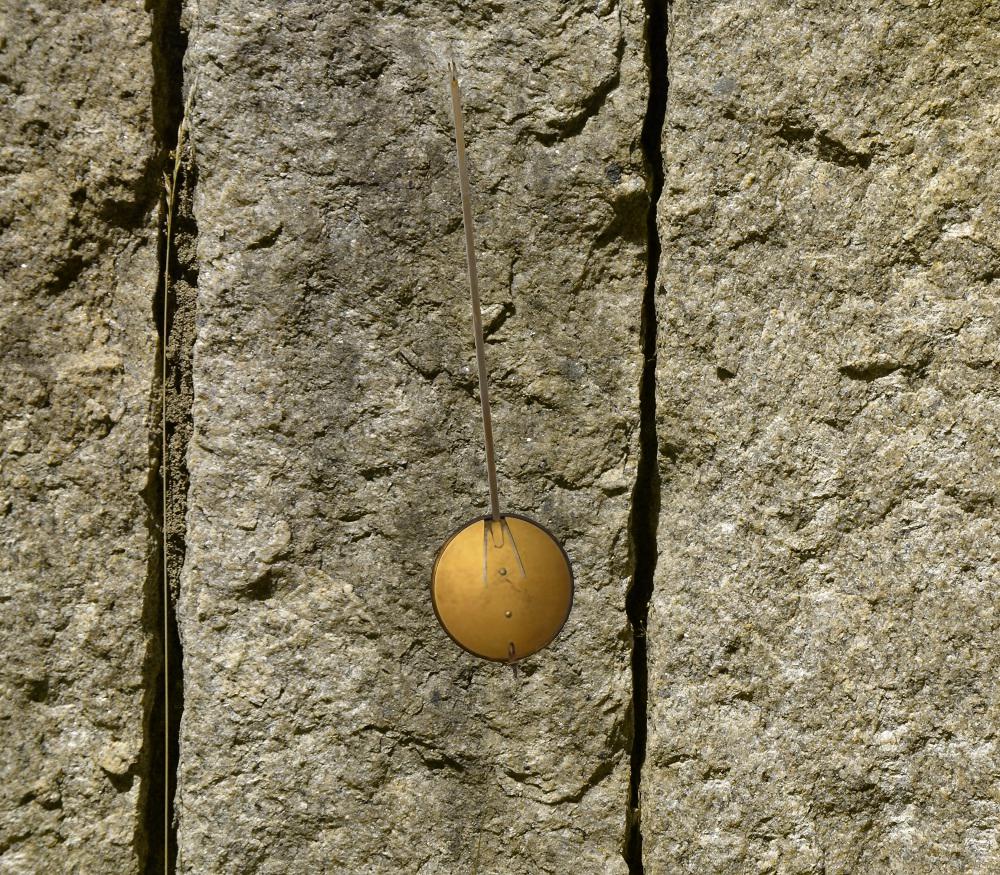}} &
\subfloat{\includegraphics[width=\WidthIms]{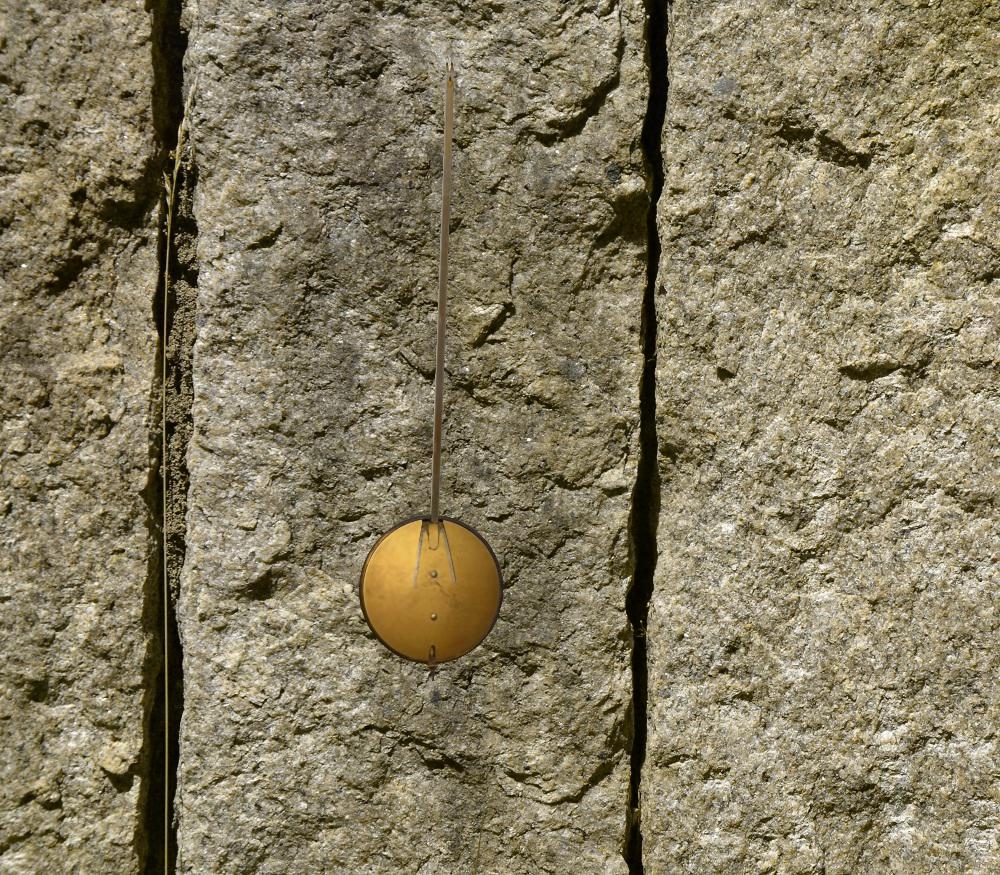}} \\[-1.0ex]
\rotatebox{90}{\spacing GT}~
\subfloat{\includegraphics[width=\WidthIms]{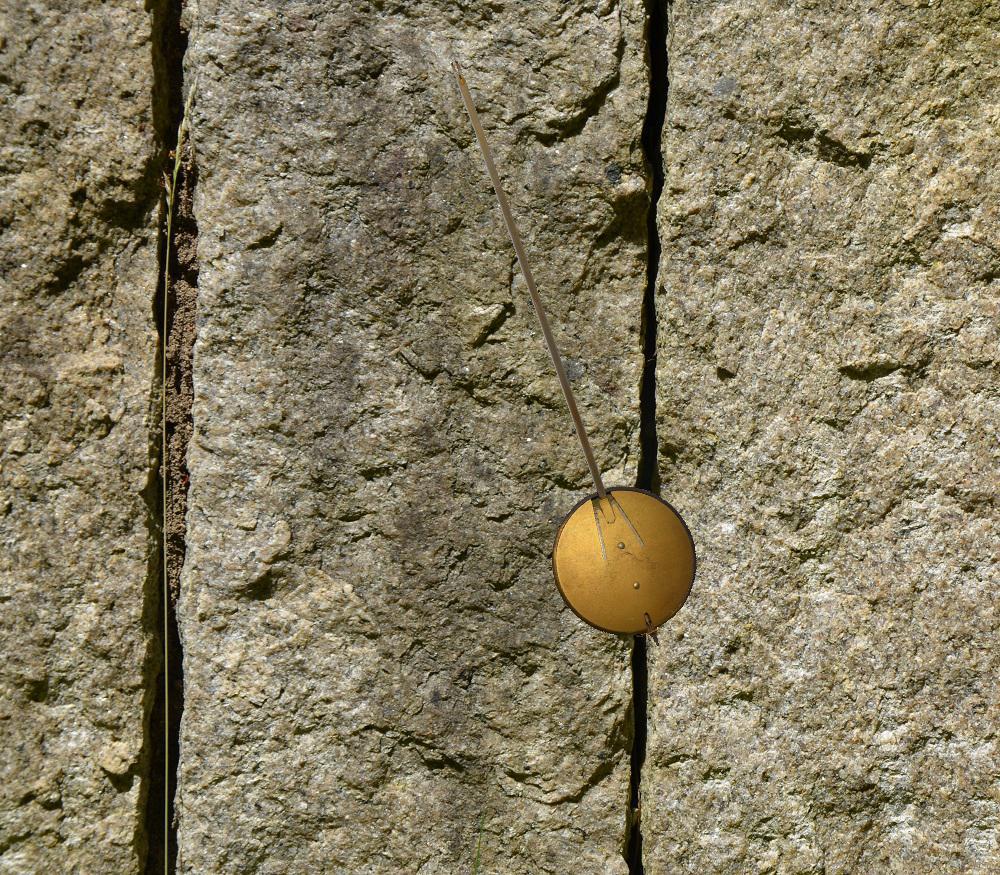}} &
\subfloat{\includegraphics[width=\WidthIms]{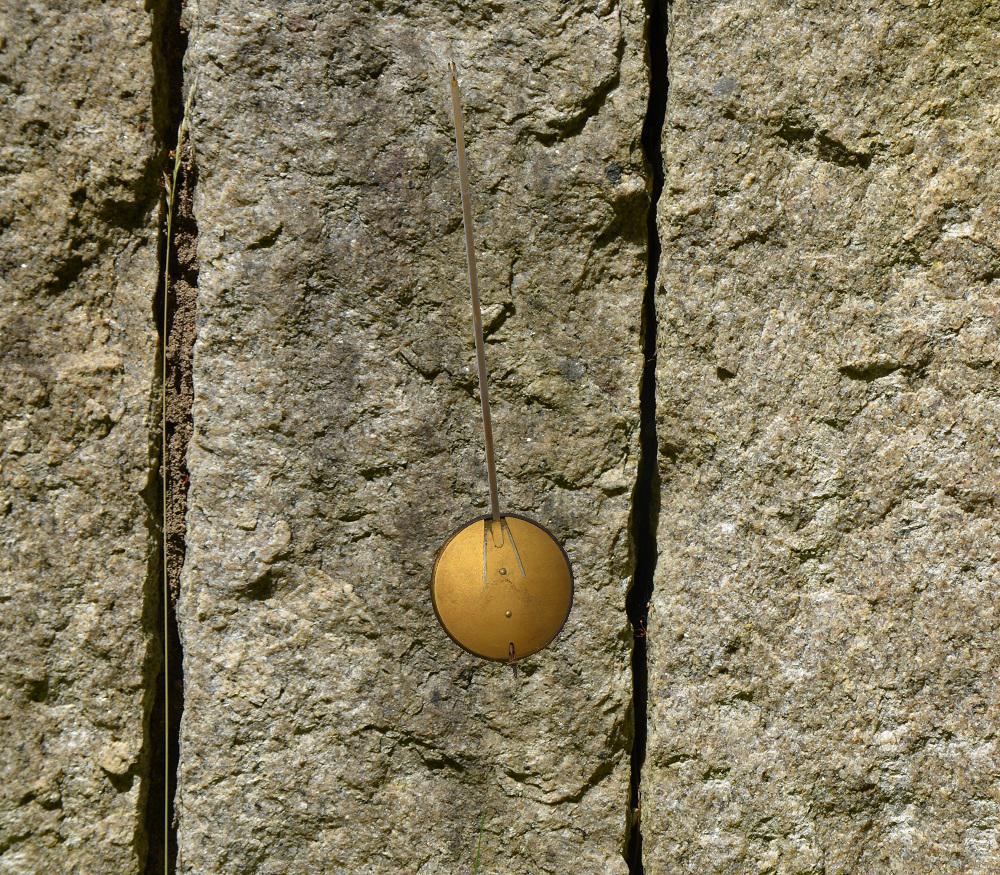}} &
\subfloat{\includegraphics[width=\WidthIms]{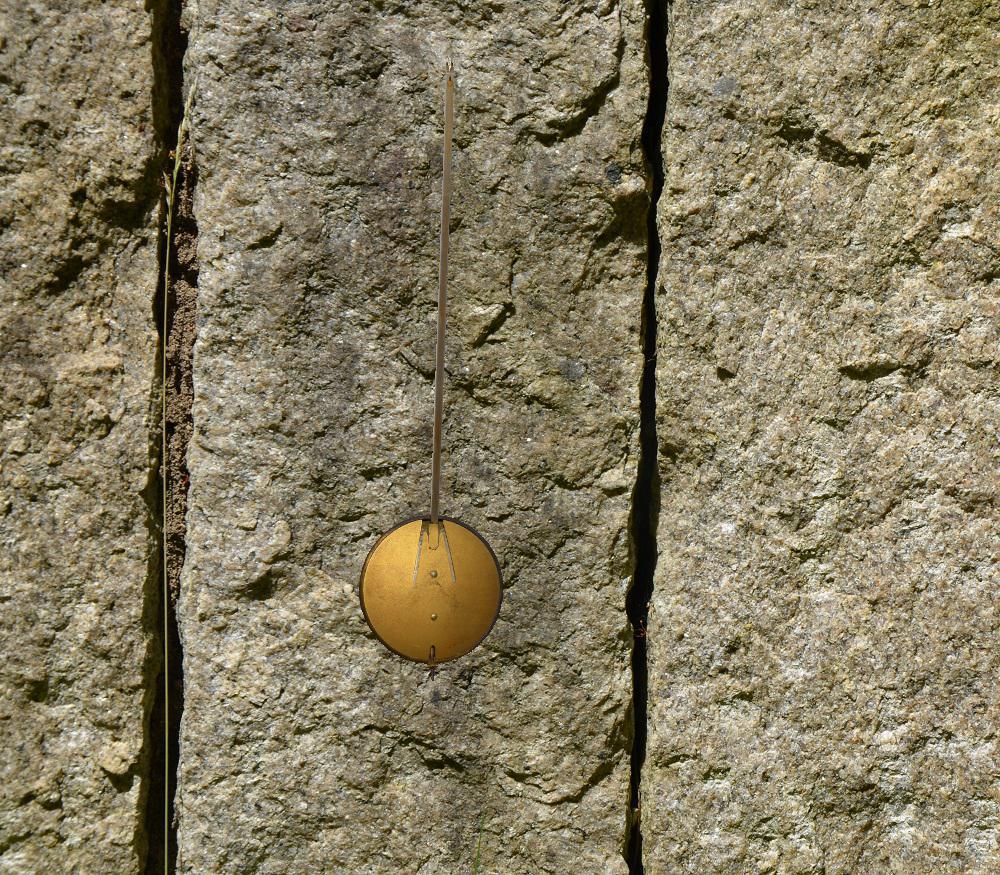}} \\
\hline \\[-3.9ex]
\rotatebox{90}{\spacing Ours}~
\subfloat{\includegraphics[width=\WidthIms]{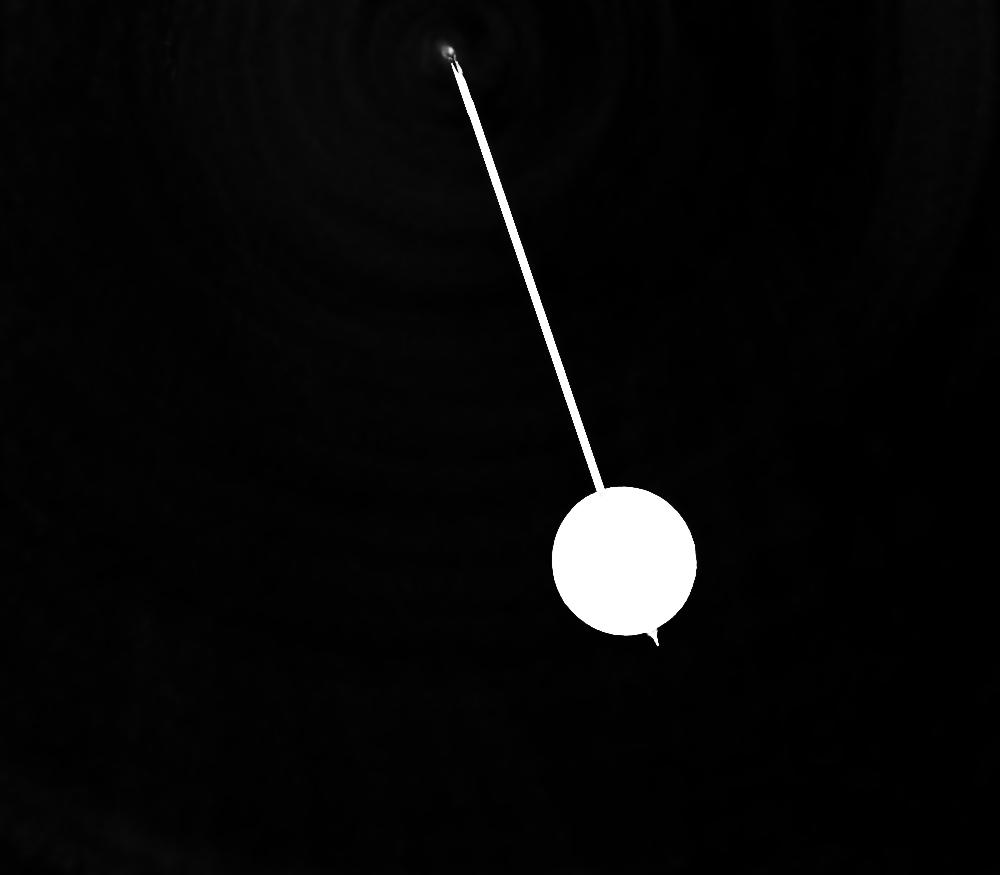}} &
\subfloat{\includegraphics[width=\WidthIms]{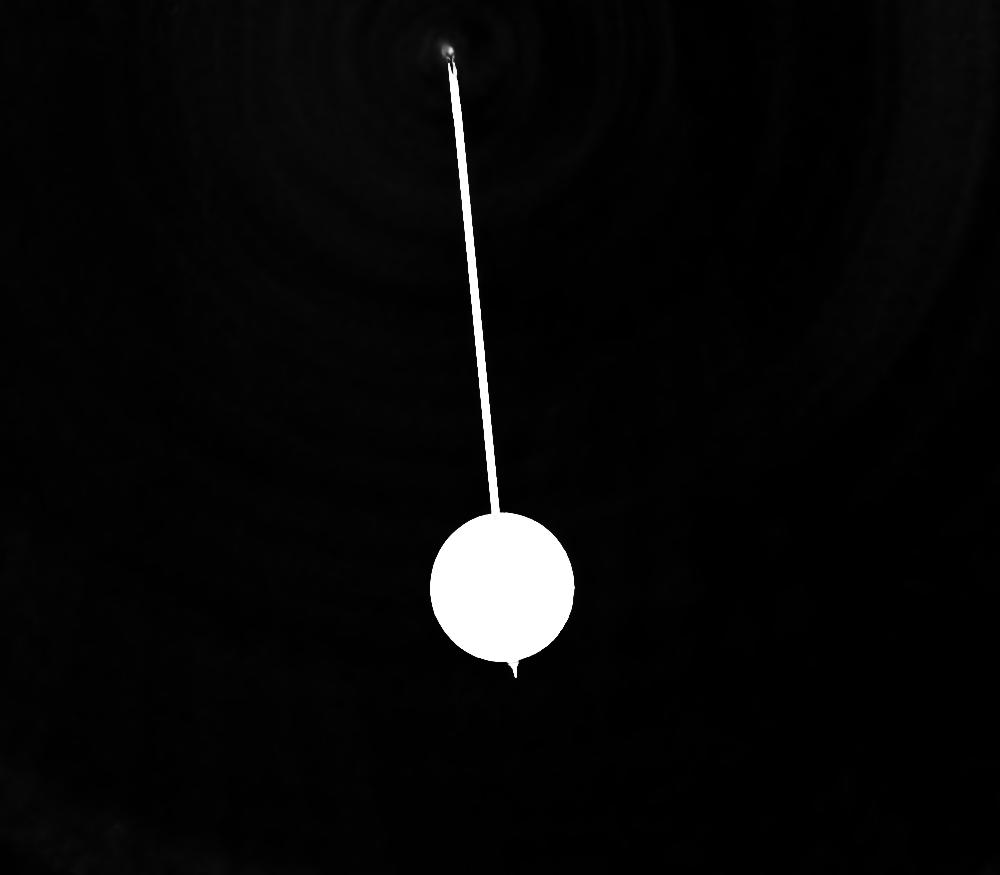}} &
\subfloat{\includegraphics[width=\WidthIms]{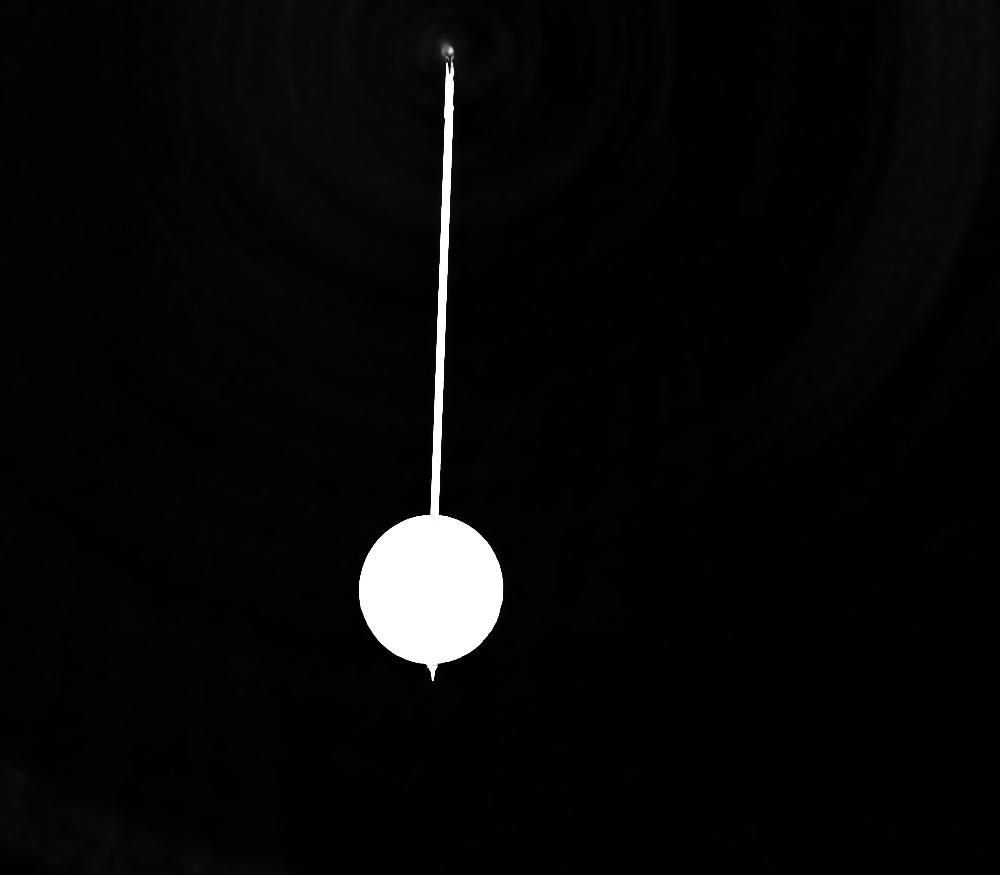}} \\[-1.0ex]
\rotatebox{90}{\spacing GT}~
\subfloat{\includegraphics[width=\WidthIms]{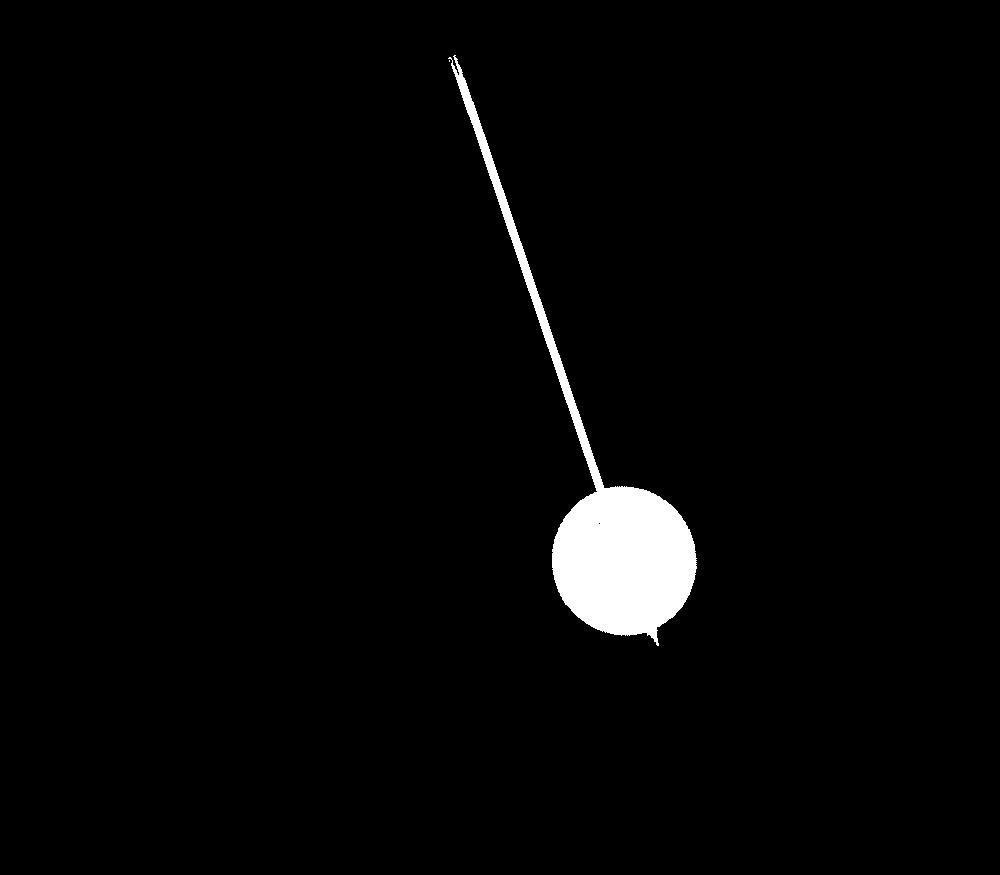}} &
\subfloat{\includegraphics[width=\WidthIms]{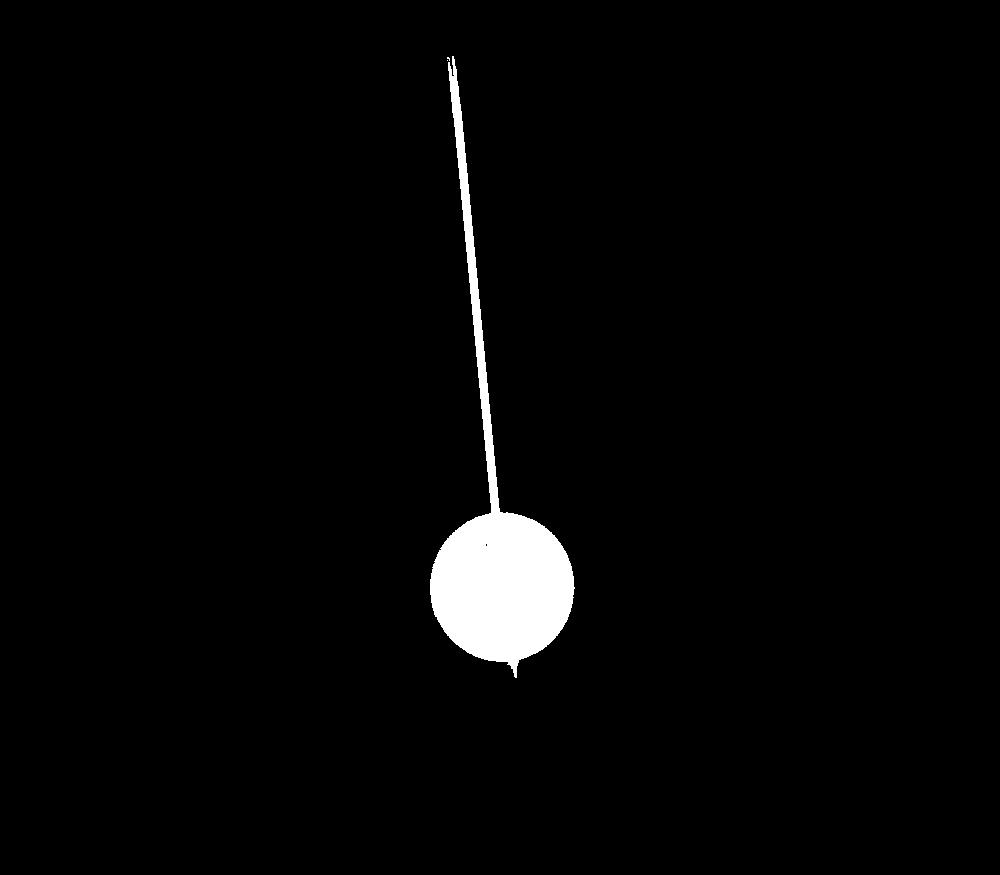}} &
\subfloat{\includegraphics[width=\WidthIms]{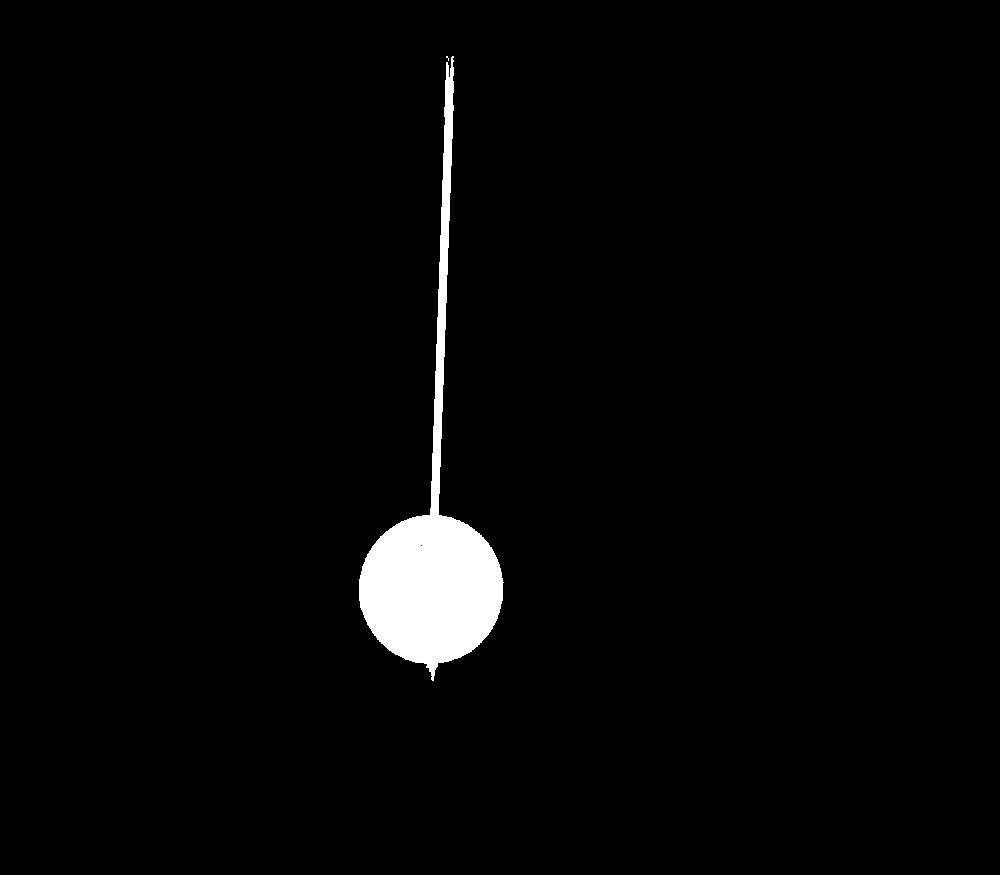}} \\
 {~training frame} & {unseen frame 1} & {unseen frame 2} \\
\end{tabular}
    \caption{Rendered frames for sequence 5 of the stonewall background. The left image is part of the training set, ``unseen frame 1'' is between two training frames, ``unseen frame 2'' is a future frame after the interval seen during training. Our method produces photorealistic predictions for the unseen time instances. Also, it predicts accurate segmentation masks for the object.}
    \label{fig:resultsStonewall}
\end{figure}

\begin{figure}
    \centering
    \newcommand\WidthIms{0.2\columnwidth}
\newcommand\Raiseheight{0.07\columnwidth}
\newcommand\spacing{\qquad\qquad\quad}
\begin{tabular}{c|cc}%
\rotatebox{90}{\spacing Ours}~
\subfloat{\includegraphics[width=\WidthIms]{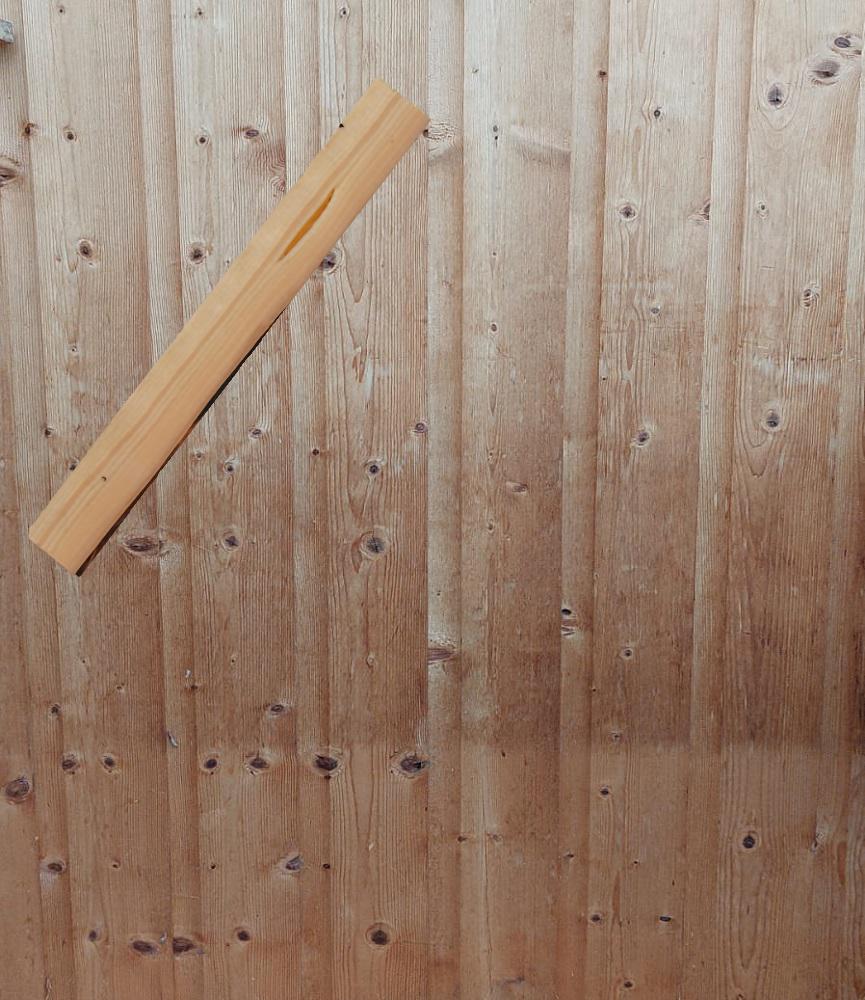}} &
\subfloat{\includegraphics[width=\WidthIms]{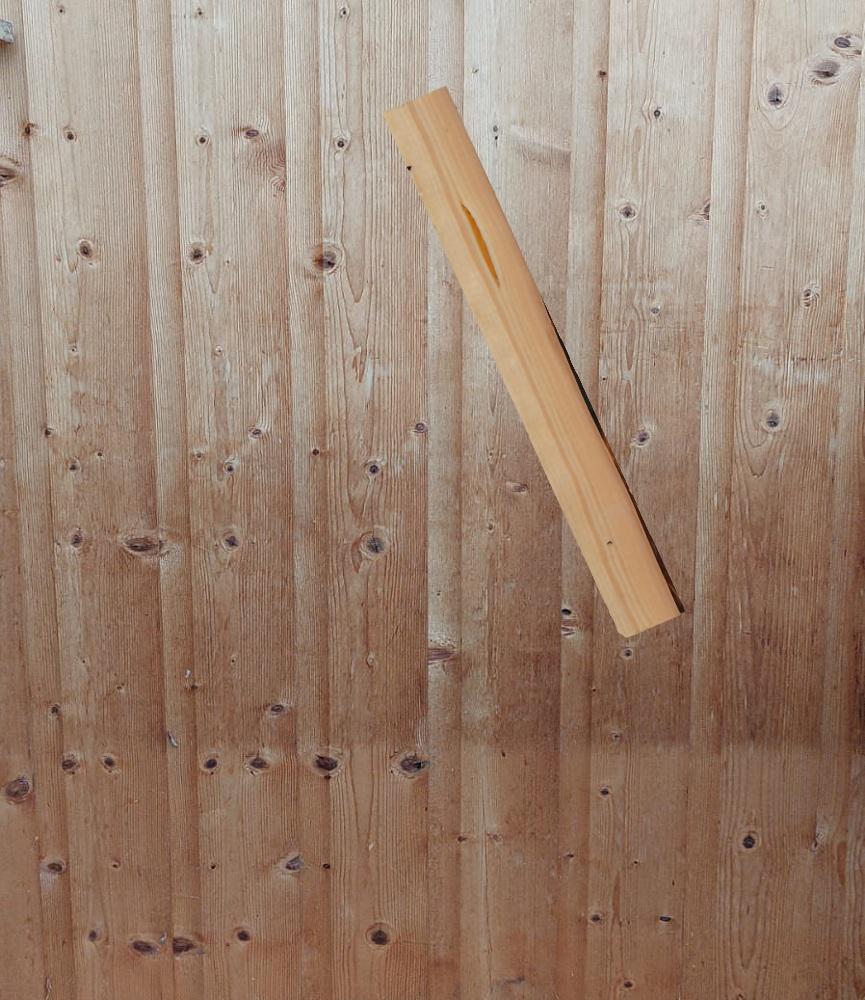}} &
\subfloat{\includegraphics[width=\WidthIms]{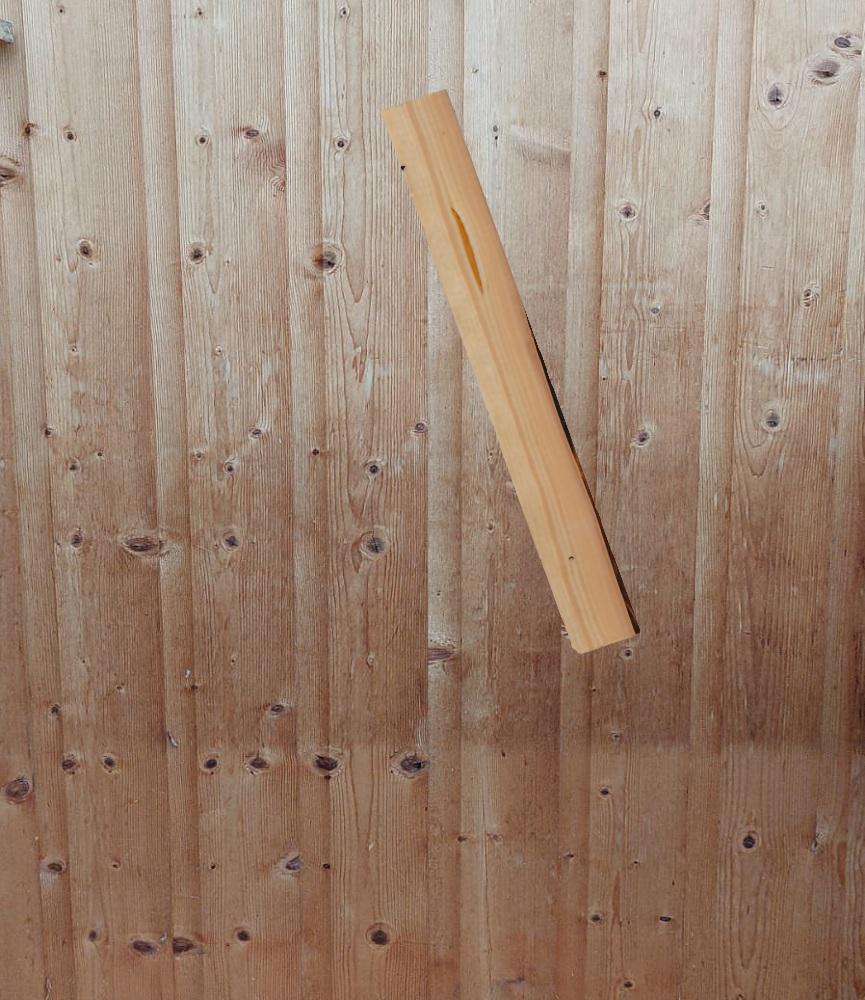}} \\[-1.0ex]
\rotatebox{90}{\spacing GT}~
\subfloat{\includegraphics[width=\WidthIms]{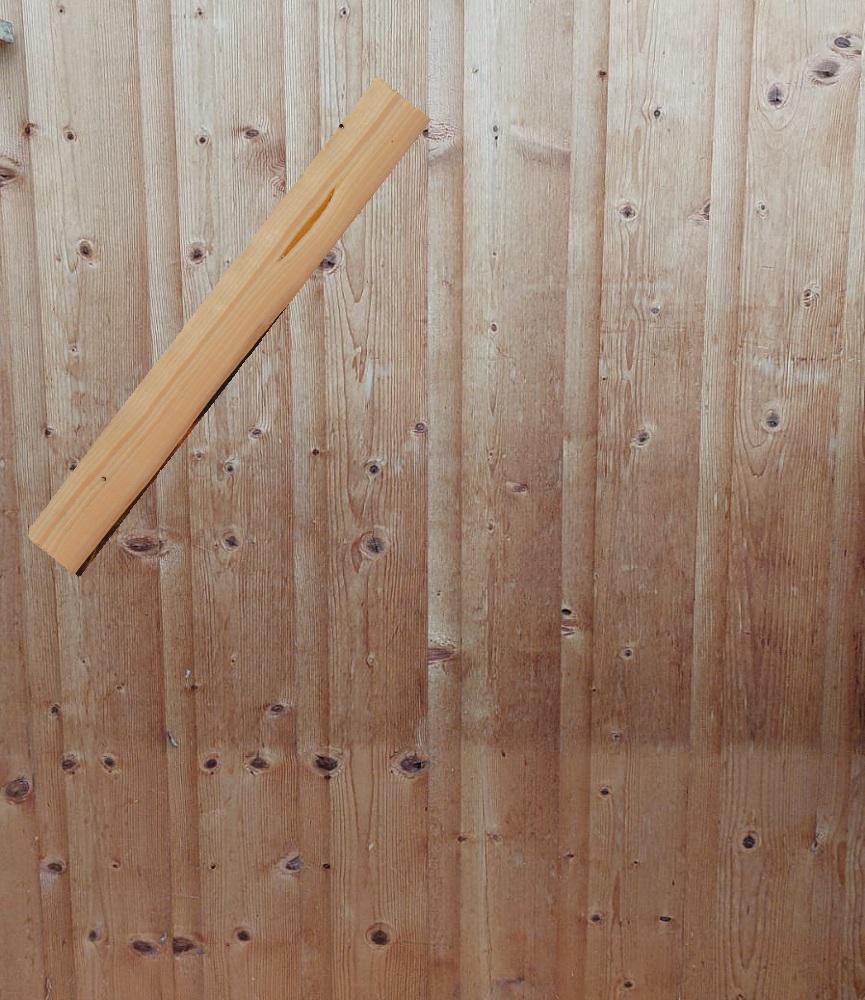}} &
\subfloat{\includegraphics[width=\WidthIms]{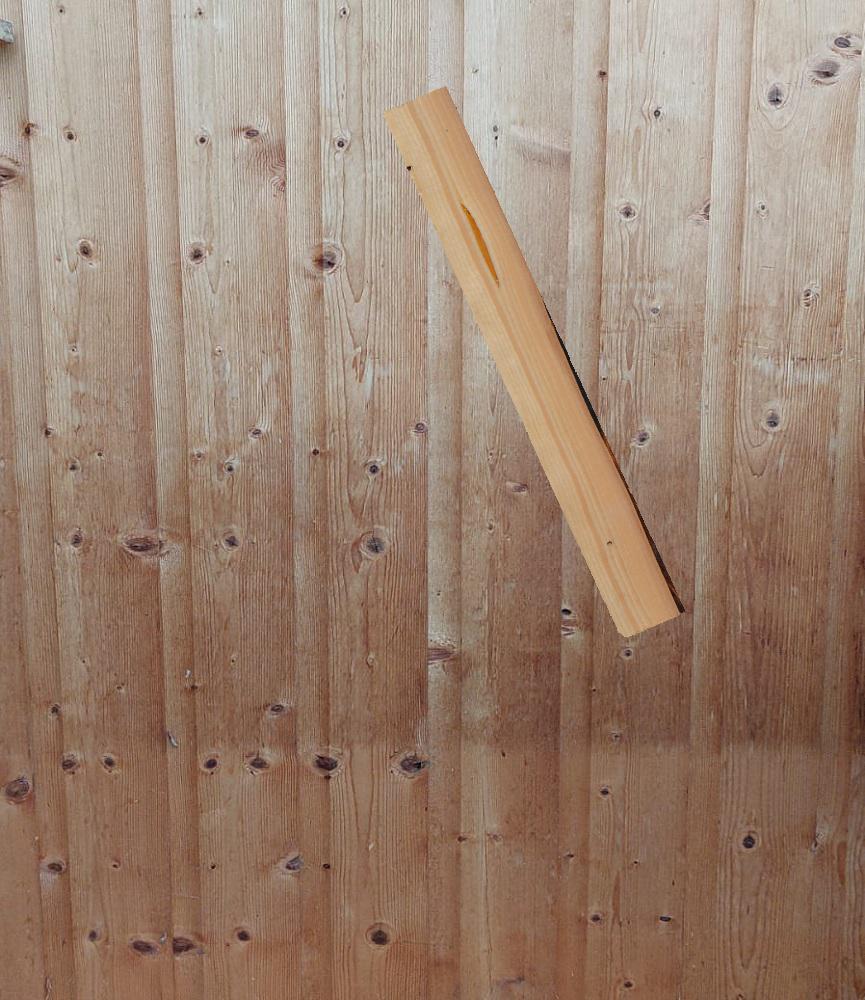}} &
\subfloat{\includegraphics[width=\WidthIms]{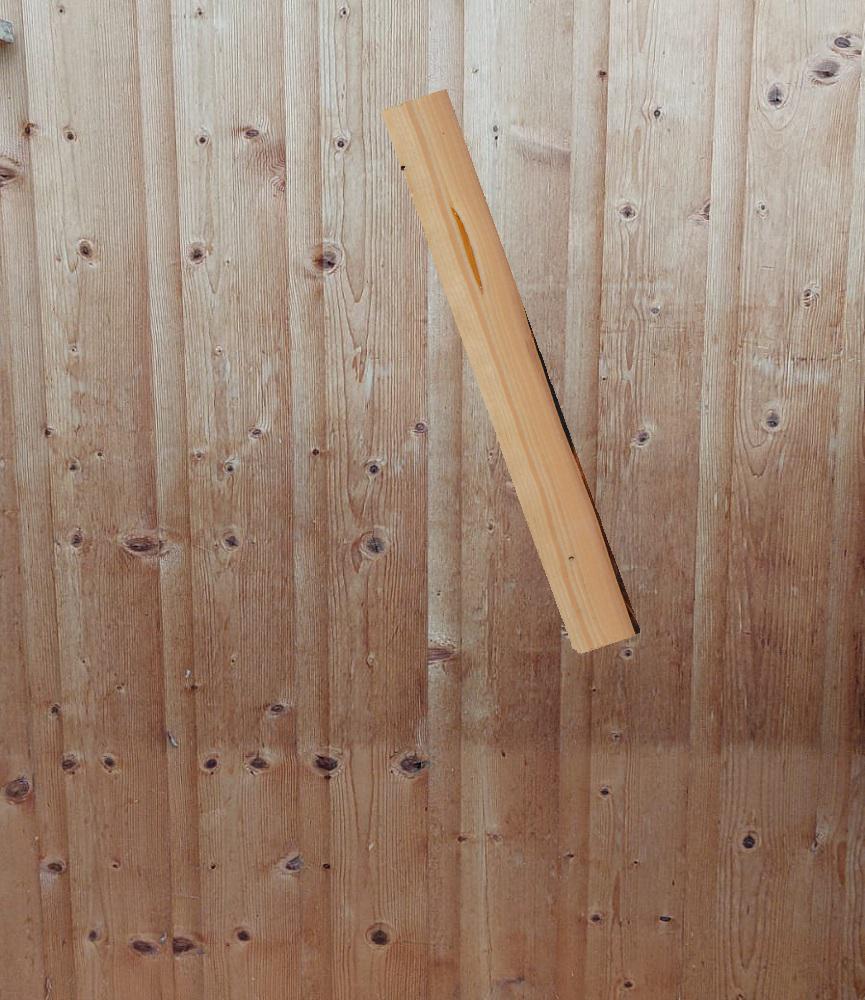}} \\
\hline \\[-3.9ex]
\rotatebox{90}{\spacing Ours}~
\subfloat{\includegraphics[width=\WidthIms]{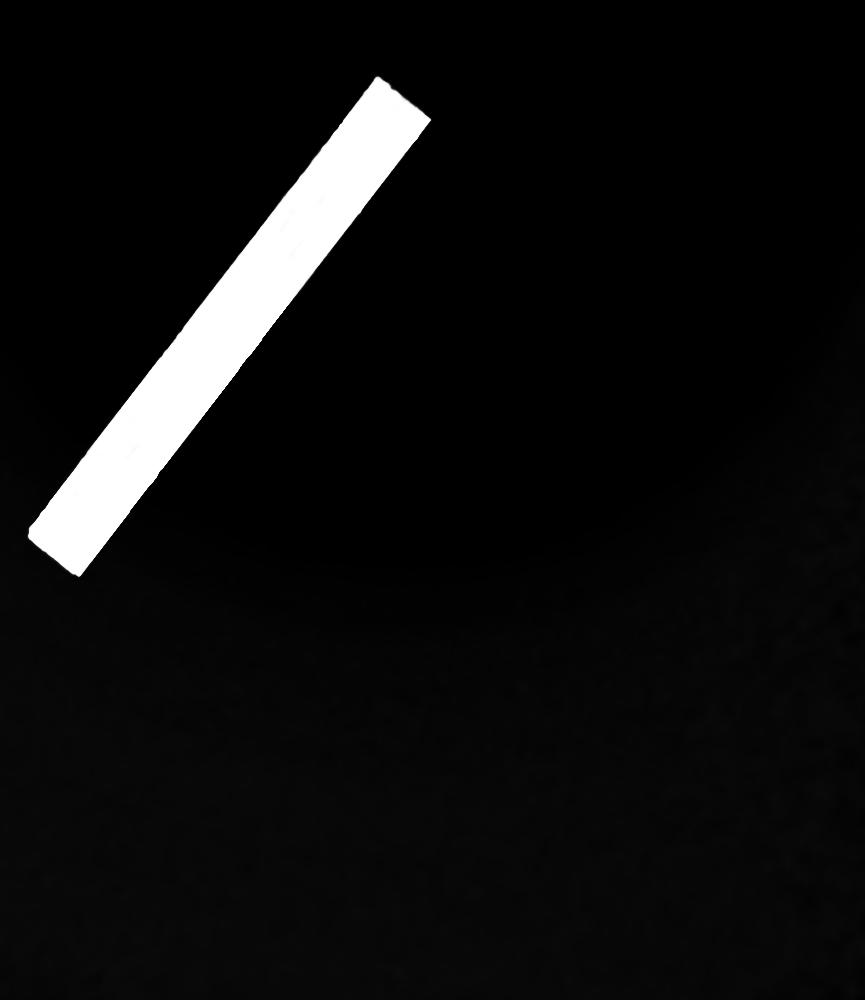}} &
\subfloat{\includegraphics[width=\WidthIms]{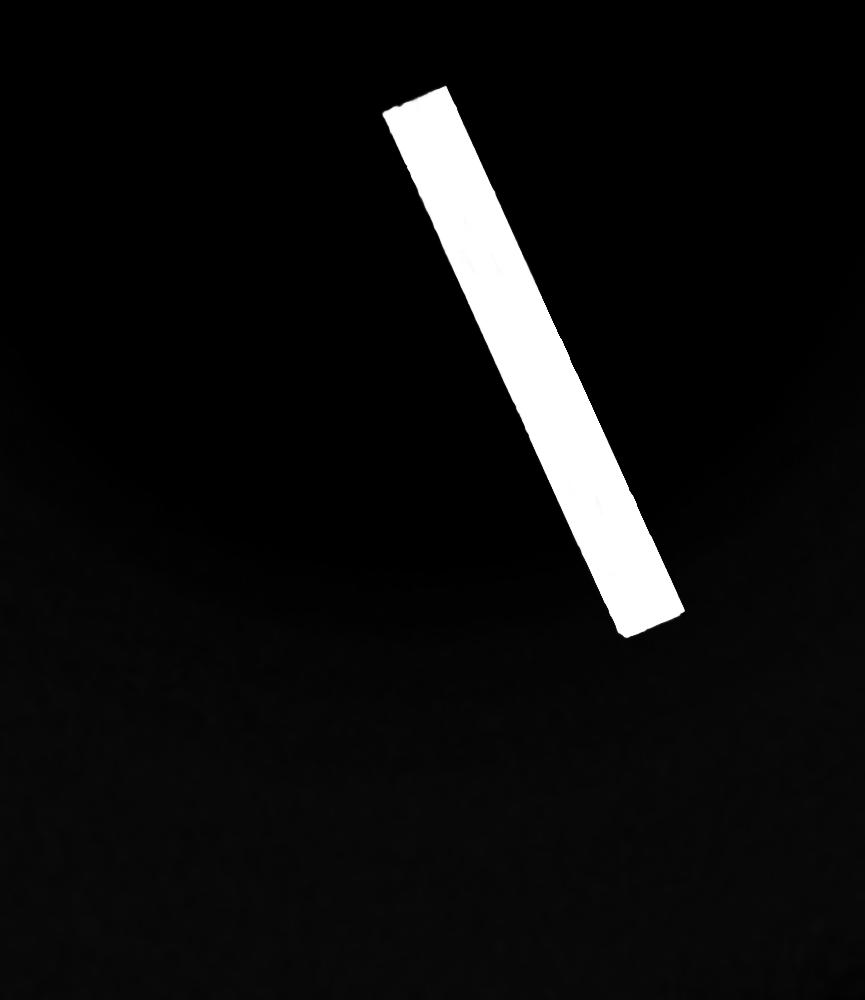}} &
\subfloat{\includegraphics[width=\WidthIms]{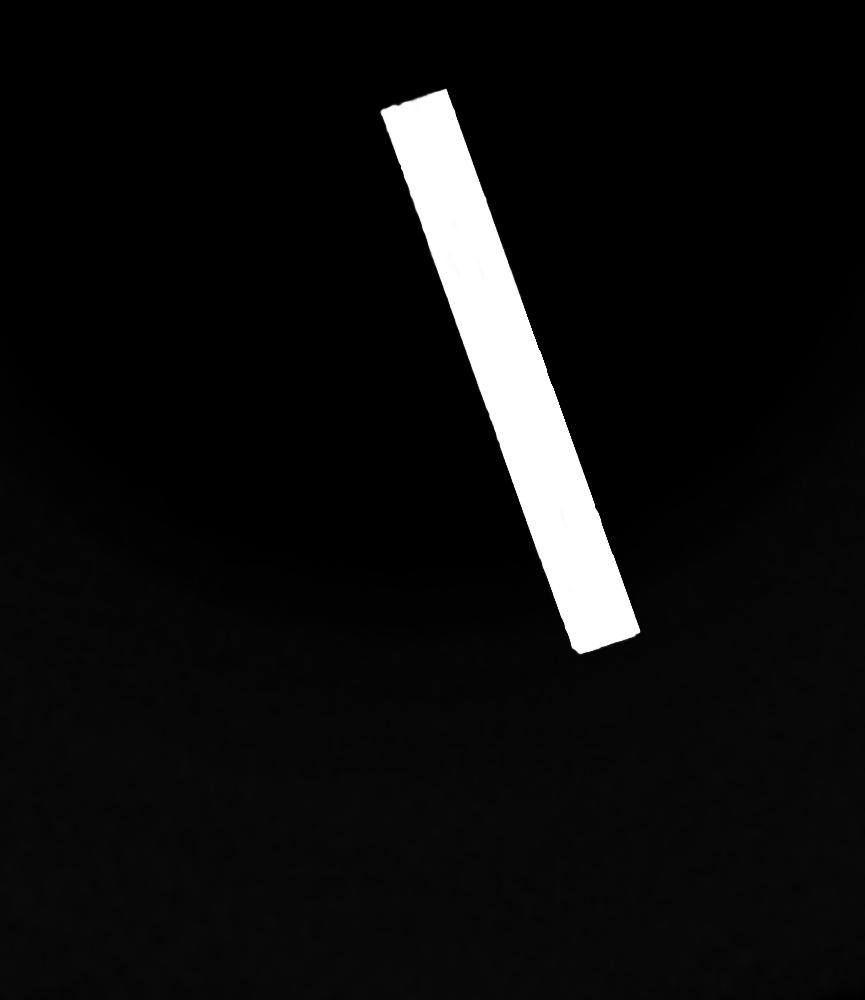}} \\[-1.0ex]
\rotatebox{90}{\spacing GT}~
\subfloat{\includegraphics[width=\WidthIms]{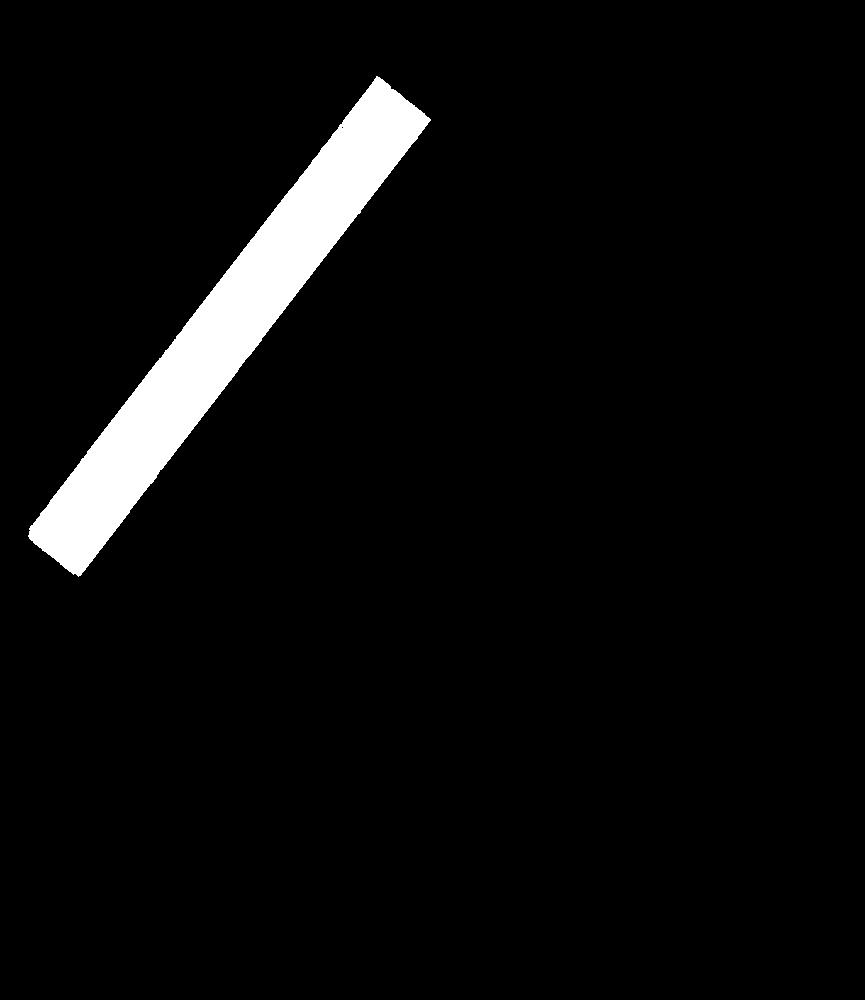}} &
\subfloat{\includegraphics[width=\WidthIms]{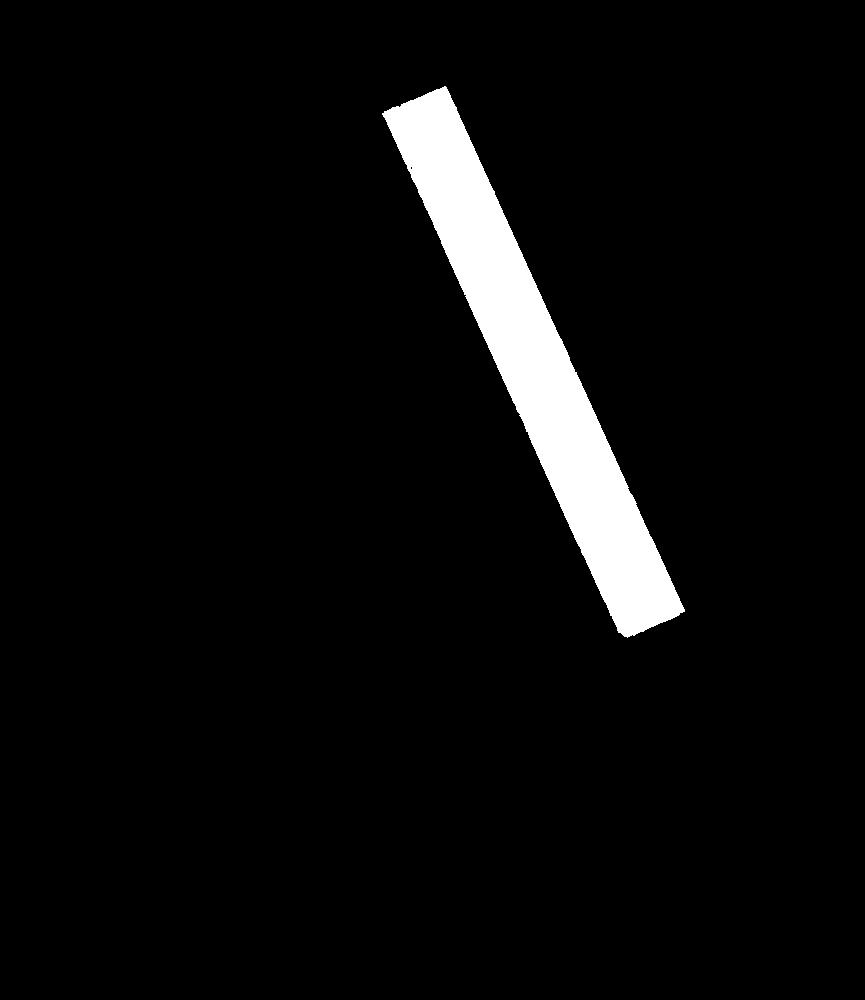}} &
\subfloat{\includegraphics[width=\WidthIms]{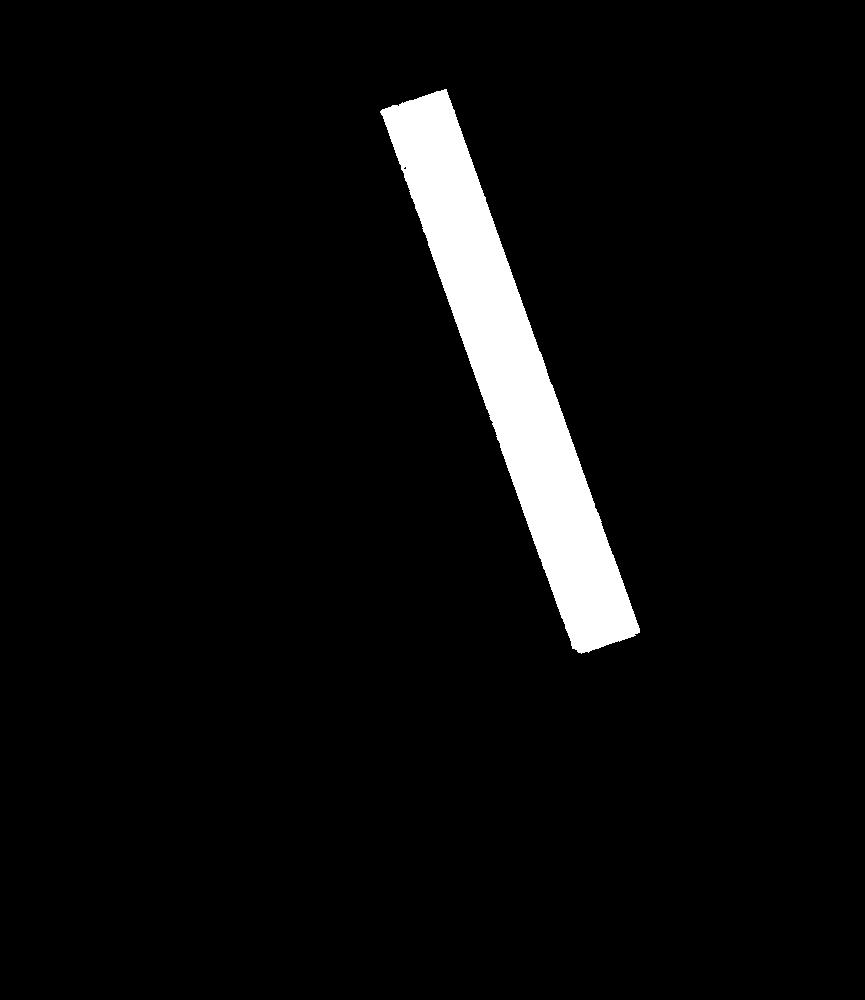}} \\
 {~training frame} & {unseen frame 1} & {unseen frame 2} \\
\end{tabular}
    \caption{Rendered frames for sequence 7 of the woodwall background. The left image is part of the training set, ``unseen frame 1'' is between two training frames, ``unseen frame 2'' is a future frame after the interval seen during training. Our method produces photorealistic predictions for the unseen time instances. Also, it predicts accurate segmentation masks for the object.}
    \label{fig:resultsWoodwall}
\end{figure}

\paragraph{Additional Results}
\Cref{fig:resultsStonewall,fig:resultsWoodwall} show additional results on the stonewall and woodwall background, also showing the predicted and the groundtruth masks. \Cref{fig:resultsWallclock} shows the masks for the sequence considered in the main text, the rendered images are repeated for convenience. The results show, that our method is able to produce excellent reconstruction for unseen time instances, both in terms of visual quality as well as in terms of predicting accurate object masks.

\subsection{Real World Examples}
\paragraph{Experimental details}
The pendulum video is recorded at a rate of $30$ fps. We extract every third frame into the dataset. For the training we select every second frame of this set and train on $10$ frames, covering $1.8$ seconds. This leaves frames between the training frames as well as frames to evaluate the extrapolation qualities. We use $31$ frames for evaluation, covering $3.9$ seconds.

For the sliding block and the ball, the relevant dynamics happen in a significantly shorter amount of time. We record the block at $30$ fps and the ball at $120$ fps. In both cases the frames cover a time interval of $0.4$ seconds. We use again every second frame for training, and use 6 training frames each. This leaves 7 frames for evaluation of the block and 8 for the ball.

\paragraph{Model parameters}
For the background we use an MLP with $N_\text{FC}=8$ fully connected layers of width $W_\text{FC}=512$ and a Fourier mapping with $N_\text{Fourier}=256$ Fourier features and variance $\sigma=30.0$ for the ball and the sliding block, and $\sigma=50.0$ for the pendulum. For the representation of the local objects we use $N_\text{FC}=8$ fully connected layers of width $W_\text{FC}=128$ and a Fourier mapping with $N_\text{Fourier}=128$ Fourier features and variance $\sigma=5.0$ for the ball and $\sigma=15.0$ for the sliding block and the pendulum.

We use an initial learning rate of $r_\text{MLP, 0}=9e-4$ for the parameters of the implicit representations and $r_\text{param, 0}=1e-3$ for the physical parameters. We set $\beta_\text{MLP}=0.9, n_\text{decay,MLP}=25$. We do not decay the learning rate for the physical parameters. We activate the mask regularization after $N_{reg}=100$ epochs and use $\lambda_{reg}=1e-3$ to balance the regularization with the photometric loss.

For the online training scheme, we start with $n_{fr,0}=5$ frames for the block and the pendulum and $n_{fr,0}=8$ for the ball. For the ball and the sliding block we increase the number of frames by one every $n_{incrT}=10$ steps, for the pendulum every $n_{incrT}=20$ steps. We train for $1200$ epochs, where one epoch is completed, when all the pixels have been considered.

We initialize the friction coefficient for the sliding block as $\mu=0$, the damping for the pendulum as $c=0.5$ and the pendulum length as $0.4$.

\begin{figure}
    \centering
    \newcommand\WidthIms{2.1cm}%
\definecolor{plotRed}{rgb}{0.85000,0.32500,0.09800}
\newcommand\Raiseheight{0.07\columnwidth}
\begin{tabular}{cccccccc}%
\rotatebox{90}{\qquad Reconstruction}~
\subfloat{\includegraphics[width=\WidthIms]{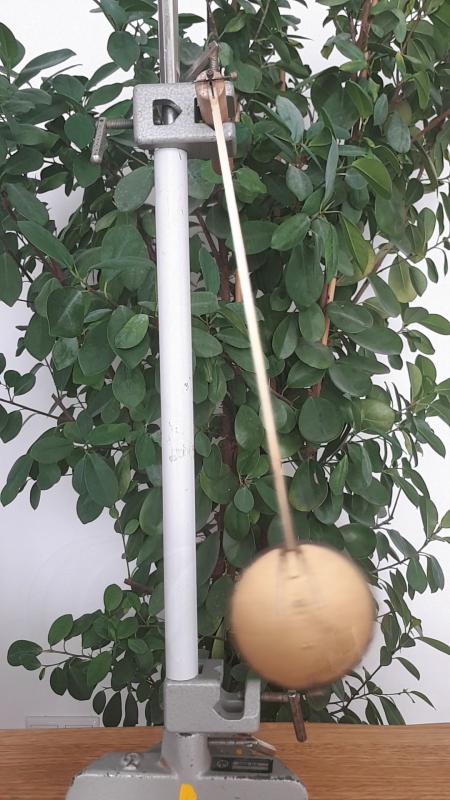}} &
\subfloat{\includegraphics[width=\WidthIms]{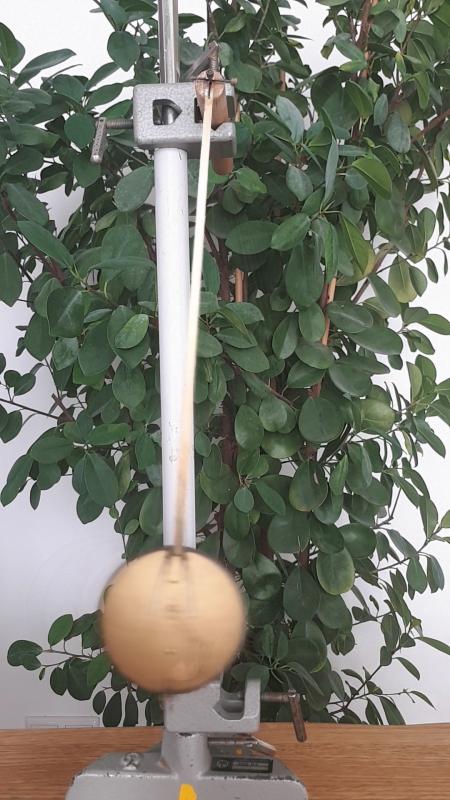}} &
\subfloat{\includegraphics[width=\WidthIms]{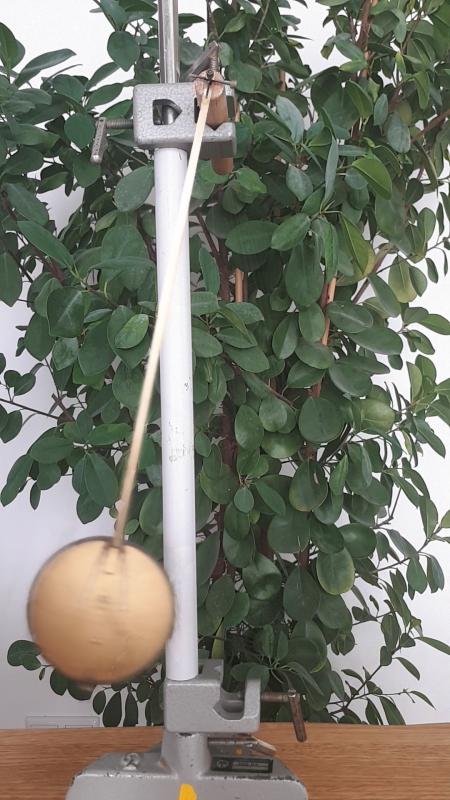}} &
\subfloat{\includegraphics[width=\WidthIms]{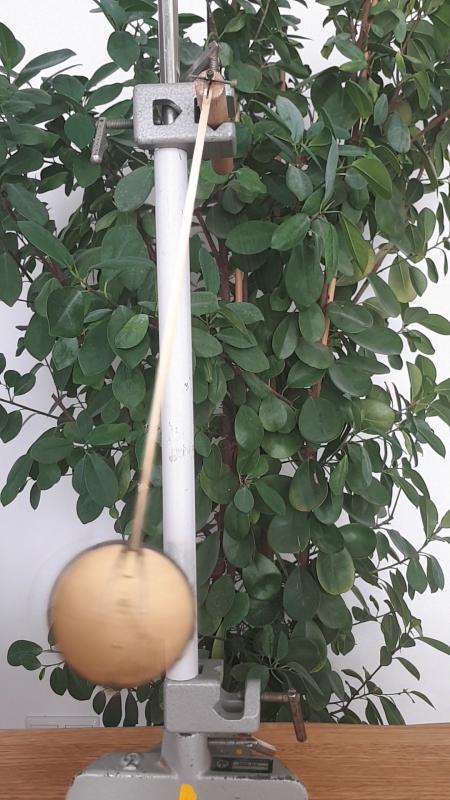}} & 
\subfloat{\includegraphics[width=\WidthIms]{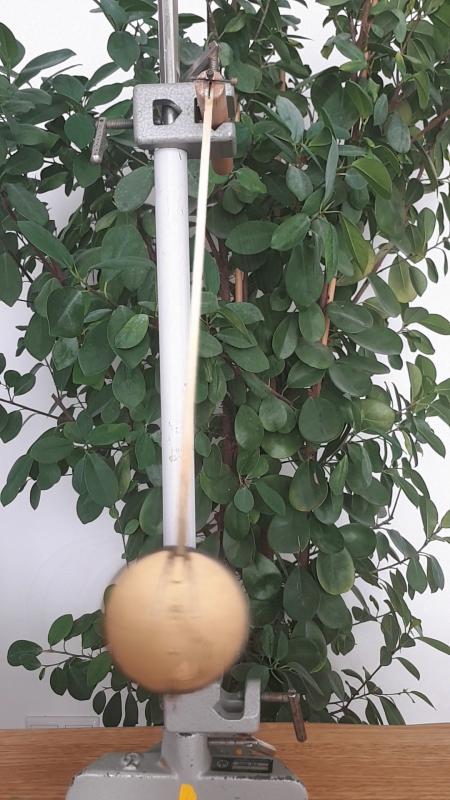}} & 
\subfloat{\includegraphics[width=\WidthIms]{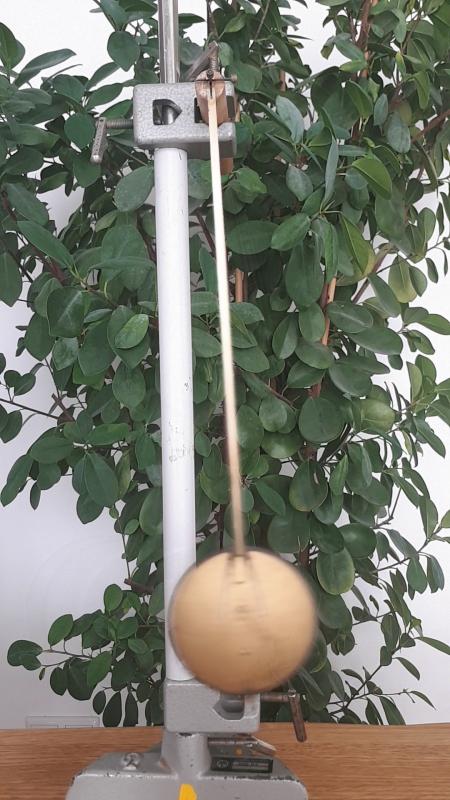}} &
\subfloat{\includegraphics[width=\WidthIms]{realWorld/Pendulum/13_eval.jpg}} \\[-1.0ex]
\rotatebox{90}{\qquad\qquad GT}~
\subfloat{\includegraphics[width=\WidthIms]{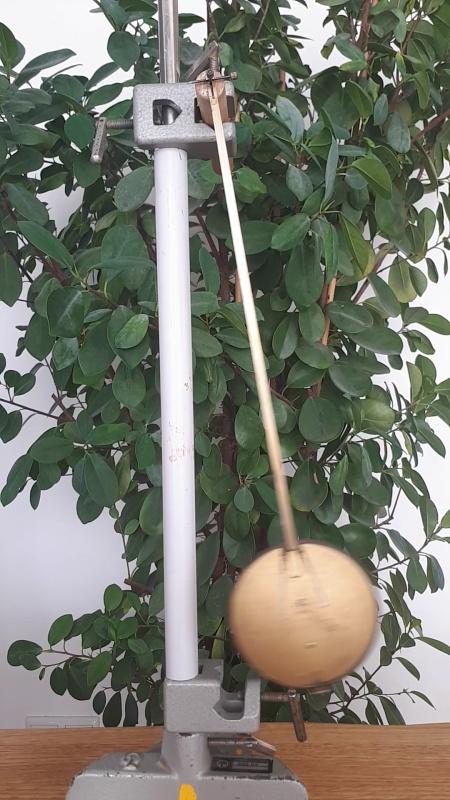}} &
\subfloat{\includegraphics[width=\WidthIms]{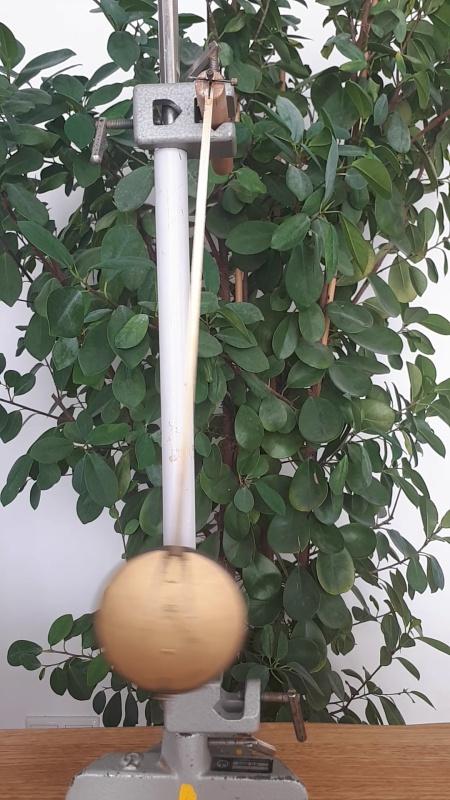}} &
\subfloat{\includegraphics[width=\WidthIms]{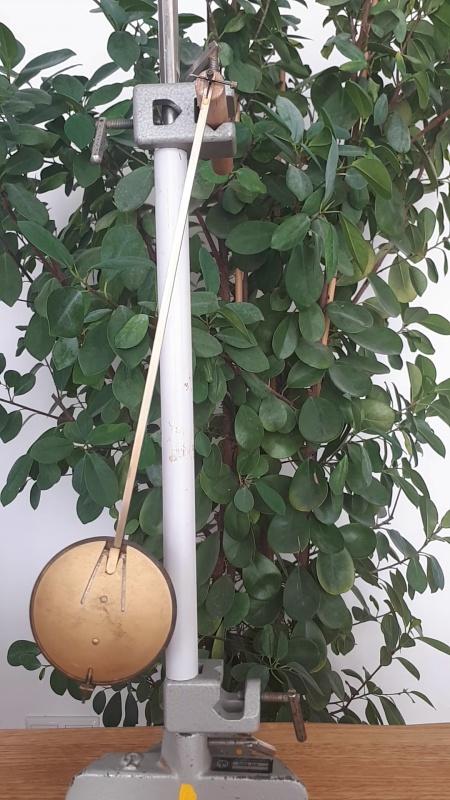}} &
\subfloat{\includegraphics[width=\WidthIms]{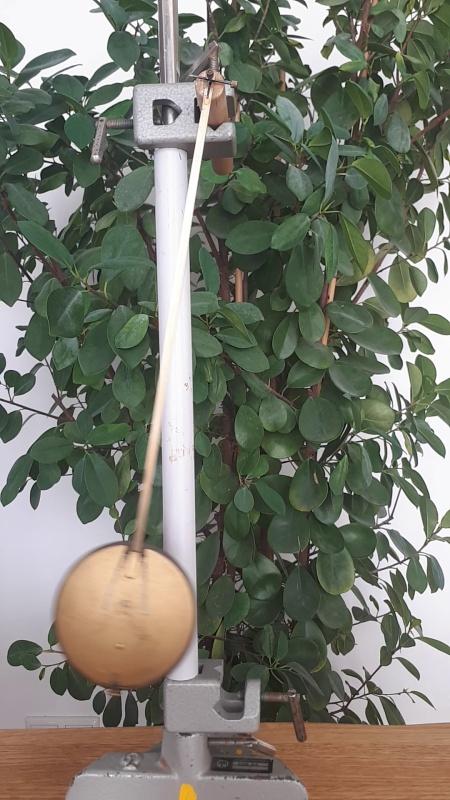}} & 
\subfloat{\includegraphics[width=\WidthIms]{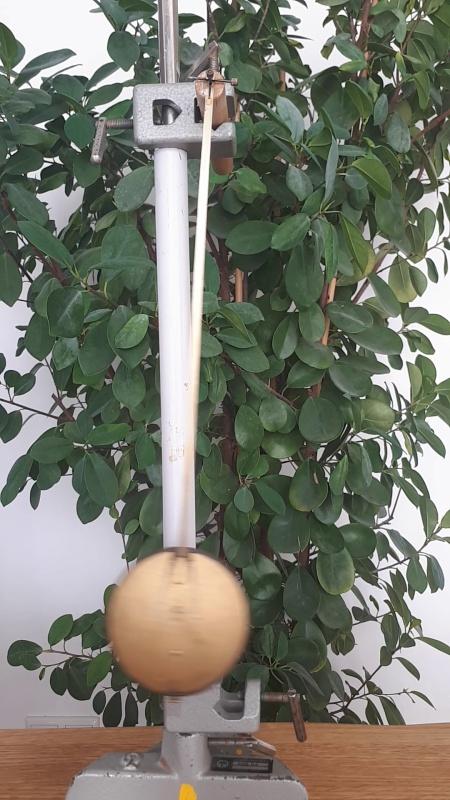}} & 
\subfloat{\includegraphics[width=\WidthIms]{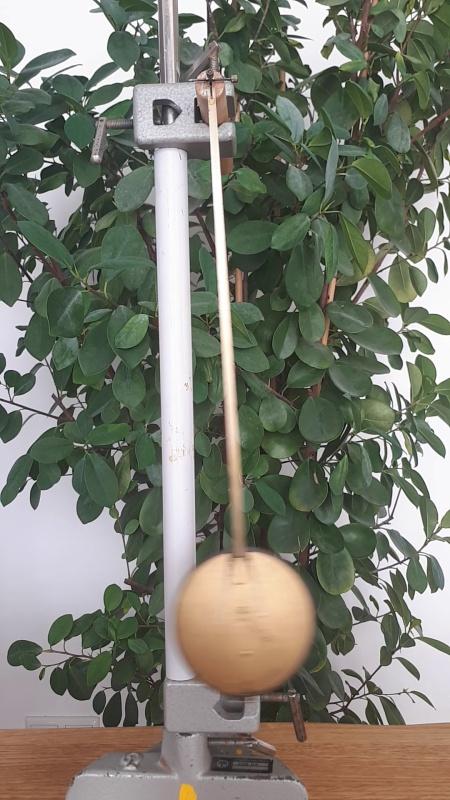}} &
\subfloat{\includegraphics[width=\WidthIms]{realWorld/Pendulum/13_gt.jpg}} \\
\hline \\[-3.9ex]
\rotatebox{90}{\qquad\quad Masks}~
\subfloat{\includegraphics[width=\WidthIms]{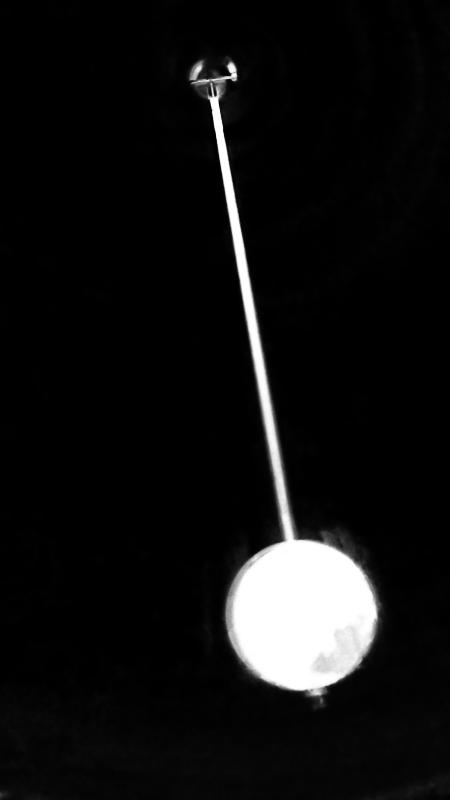}} &
\subfloat{\includegraphics[width=\WidthIms]{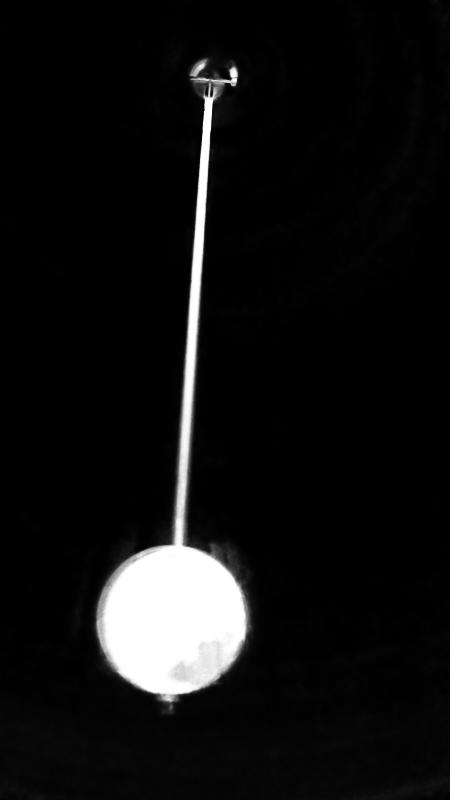}} &
\subfloat{\includegraphics[width=\WidthIms]{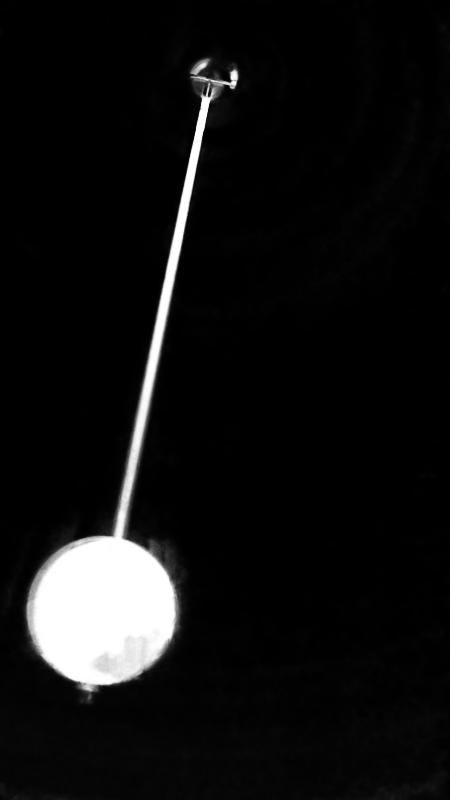}} &
\subfloat{\includegraphics[width=\WidthIms]{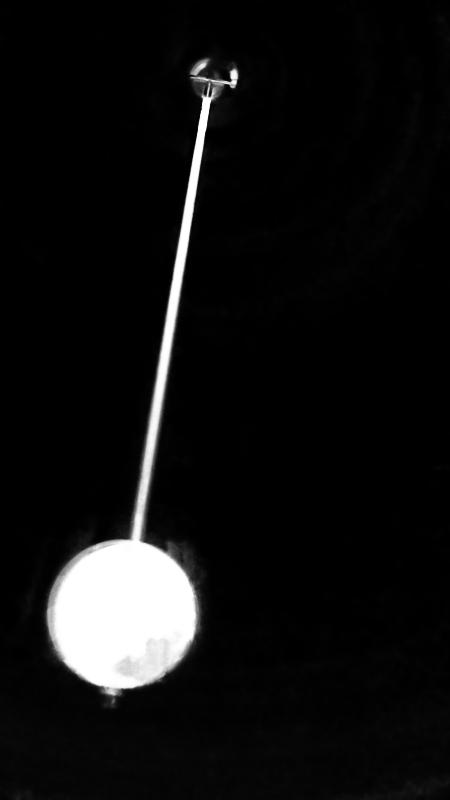}} & 
\subfloat{\includegraphics[width=\WidthIms]{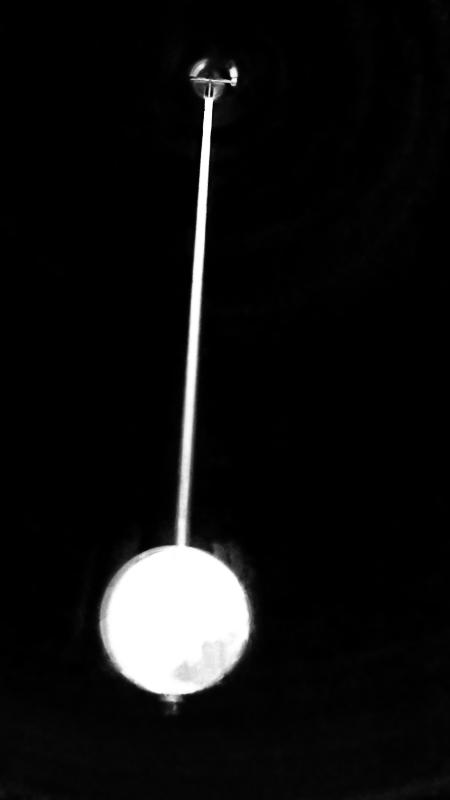}} & 
\subfloat{\includegraphics[width=\WidthIms]{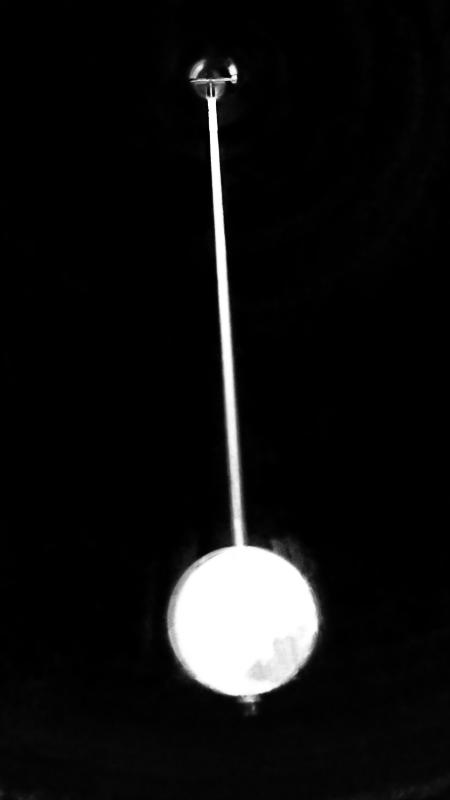}} & 
\subfloat{\includegraphics[width=\WidthIms]{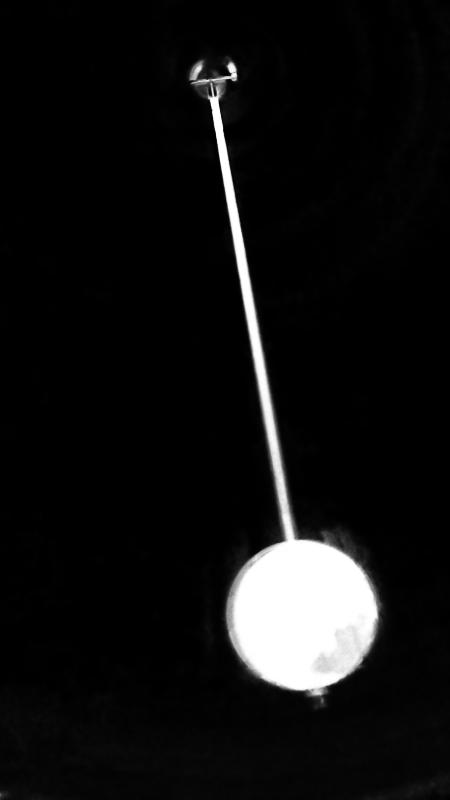}}\\
&
&
\raisebox{0.1cm}{\begin{tikzpicture} \draw [-stealth, plotRed](0,0) -- (0.7,0);\end{tikzpicture}}&
&
&
&
\end{tabular}
    \caption{Rendered frames 8-13 of the test set for the real world pendulum sequence. The two left frames are between training frames, the remaining frames are extrapolated (indicated by the red arrow). Our method produces photorealistic predictions for the unseen time instances. Also, it predicts accurate segmentation masks for the object. }
    \label{fig:renderingsRealPendulum}
\end{figure}

\begin{figure}
    \centering
    \newcommand\WidthIms{0.23\columnwidth}
\newcommand\Raiseheight{0.07\columnwidth}
\begin{tabular}{cccc}%
\rotatebox{90}{\quad Renderings}~
\subfloat{\includegraphics[width=\WidthIms]{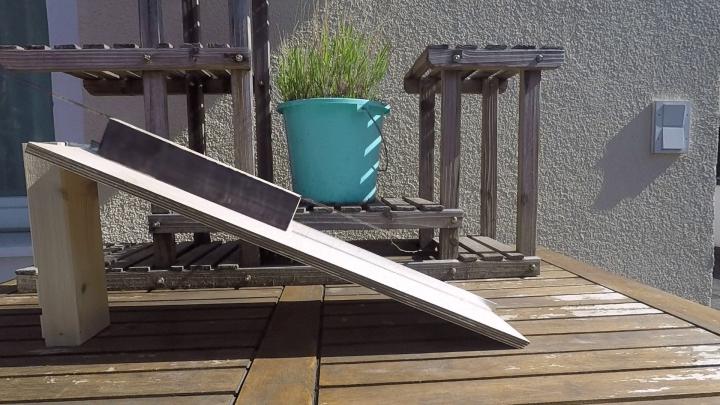}} &
\subfloat{\includegraphics[width=\WidthIms]{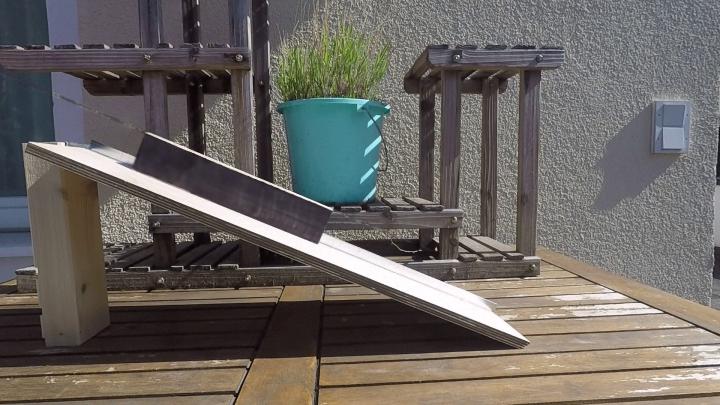}} &
\subfloat{\includegraphics[width=\WidthIms]{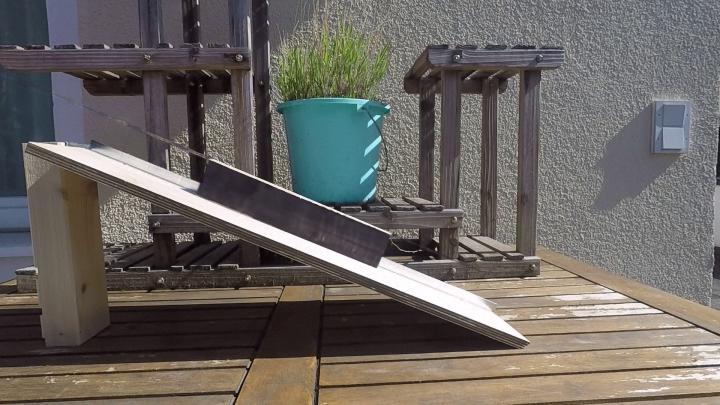}} & 
\subfloat{\includegraphics[width=\WidthIms]{realWorld/Block/7_eval.jpg}}\\[-1.0ex]
\rotatebox{90}{\qquad~ GT}~
\subfloat{\includegraphics[width=\WidthIms]{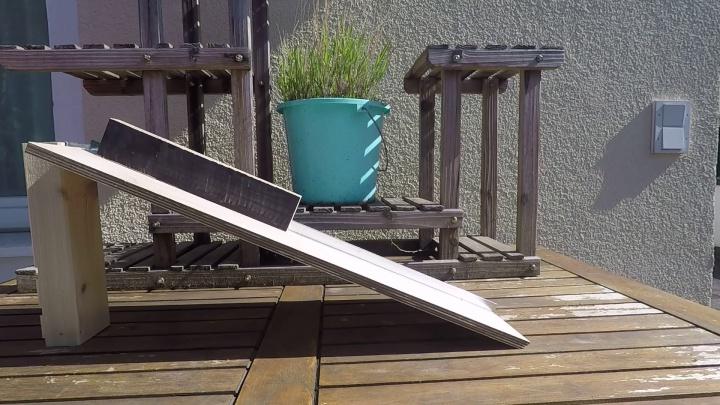}} &
\subfloat{\includegraphics[width=\WidthIms]{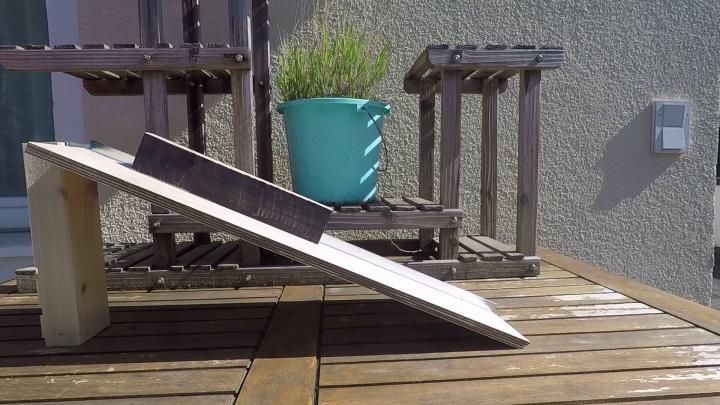}} &
\subfloat{\includegraphics[width=\WidthIms]{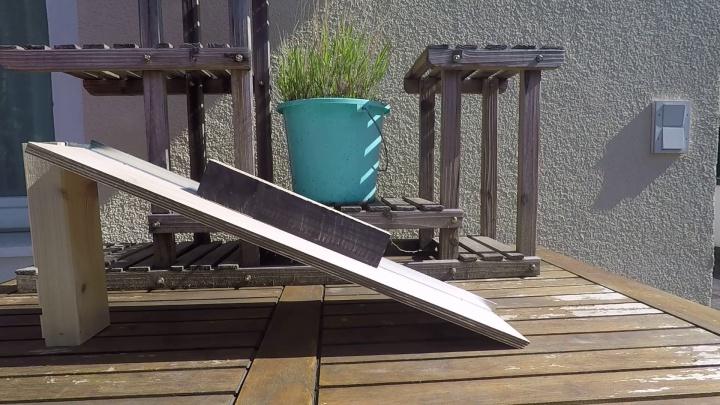}} & 
\subfloat{\includegraphics[width=\WidthIms]{realWorld/Block/7_gt.jpg}}\\
\hline \\[-3.9ex]
\rotatebox{90}{\qquad Masks}~
\subfloat{\includegraphics[width=\WidthIms]{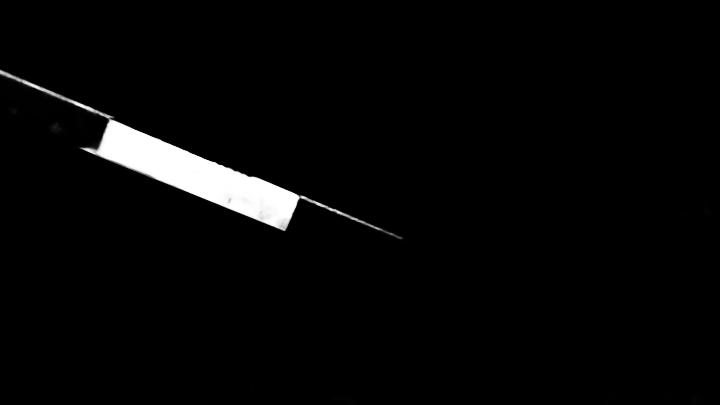}} &
\subfloat{\includegraphics[width=\WidthIms]{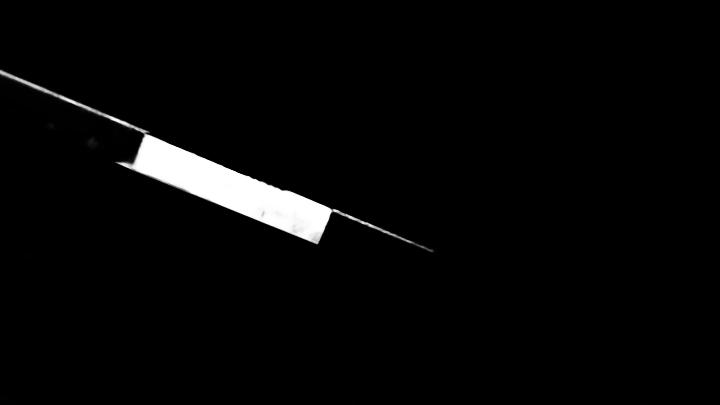}} &
\subfloat{\includegraphics[width=\WidthIms]{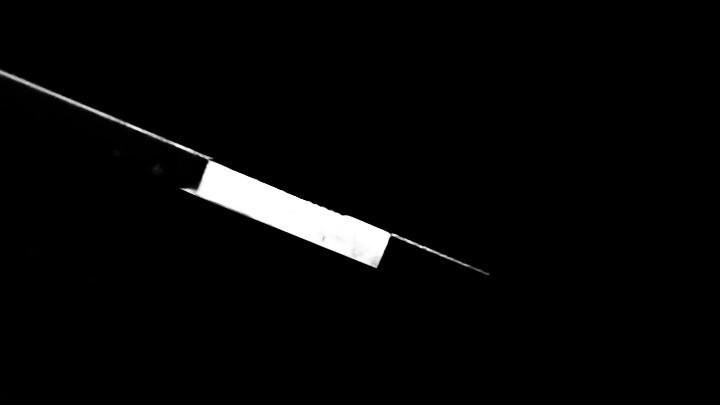}} & 
\subfloat{\includegraphics[width=\WidthIms]{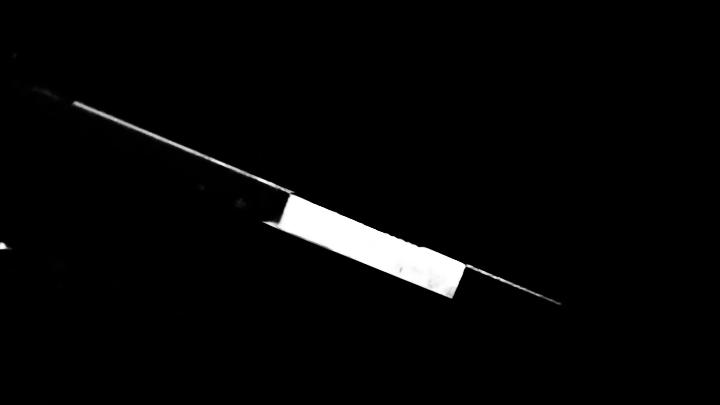}}\\
\end{tabular}
    \caption{Rendered frames 1, 3, 5, and 7 of the test set for the real world sliding block sequence. The frames are between training frames. Our method produces photorealistic predictions for the unseen time instances. Also, it predicts accurate segmentation masks for the object. }
    \label{fig:renderingsBlock}
\end{figure}

\begin{figure}
    \centering
    \newcommand\WidthIms{0.23\columnwidth}
\newcommand\Raiseheight{0.07\columnwidth}
\begin{tabular}{cccc}%
\rotatebox{90}{\quad Renderings}~
\subfloat{\includegraphics[width=\WidthIms]{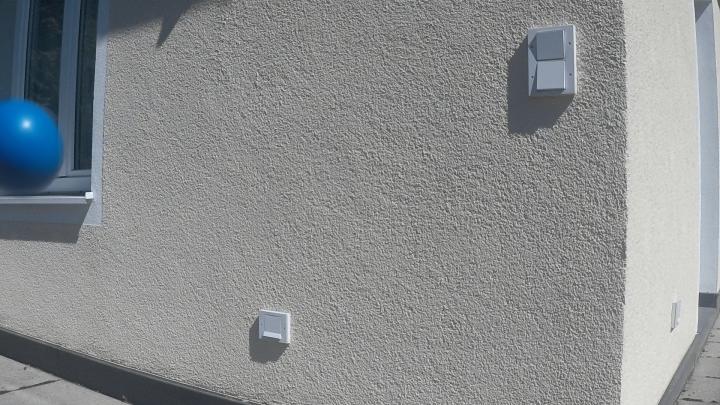}} &
\subfloat{\includegraphics[width=\WidthIms]{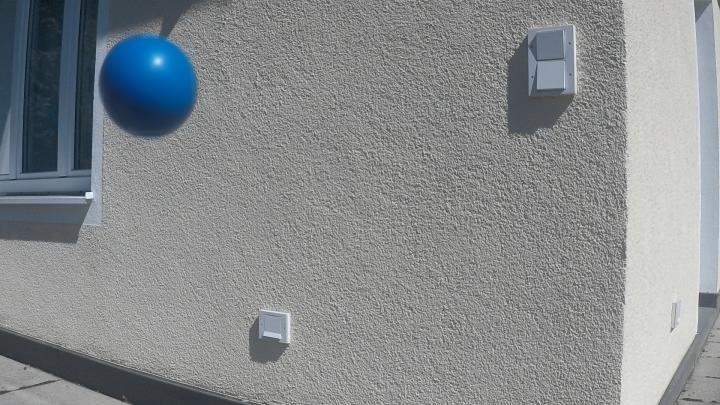}} &
\subfloat{\includegraphics[width=\WidthIms]{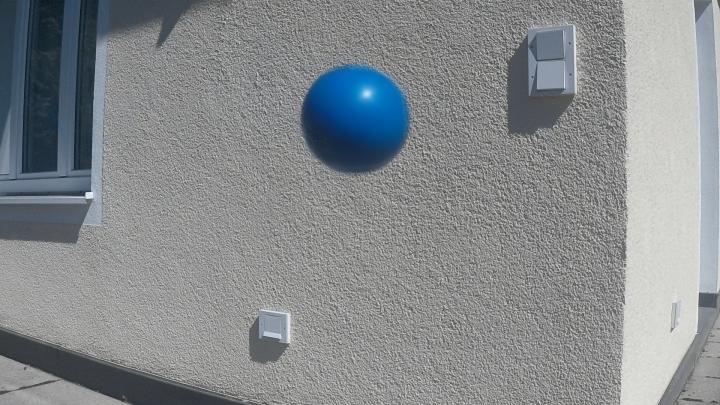}} & 
\subfloat{\includegraphics[width=\WidthIms]{realWorld/Ball/7_eval.jpg}}\\[-1.0ex]
\rotatebox{90}{\qquad~ GT}~
\subfloat{\includegraphics[width=\WidthIms]{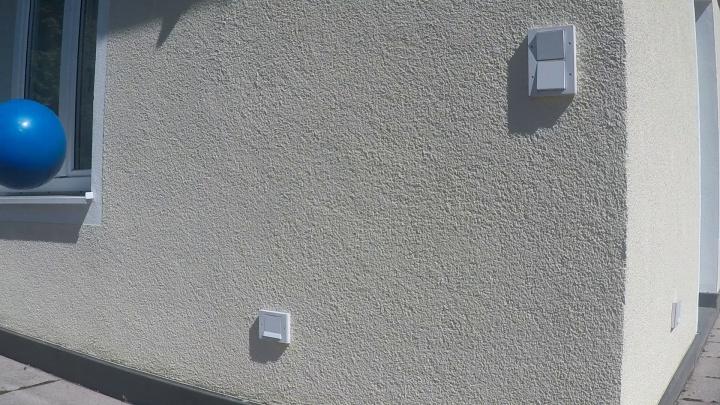}} &
\subfloat{\includegraphics[width=\WidthIms]{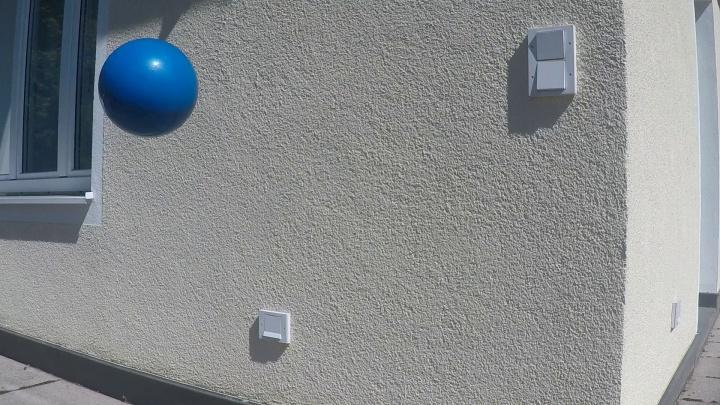}} &
\subfloat{\includegraphics[width=\WidthIms]{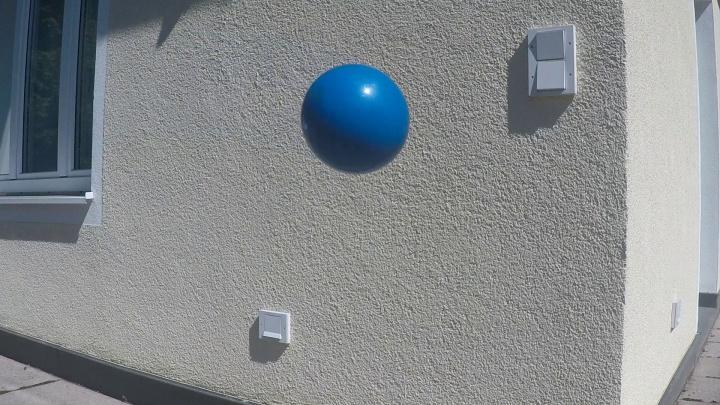}} & 
\subfloat{\includegraphics[width=\WidthIms]{realWorld/Ball/7_gt.jpg}}\\
\hline \\[-3.9ex]
\rotatebox{90}{\qquad Masks}~
\subfloat{\includegraphics[width=\WidthIms]{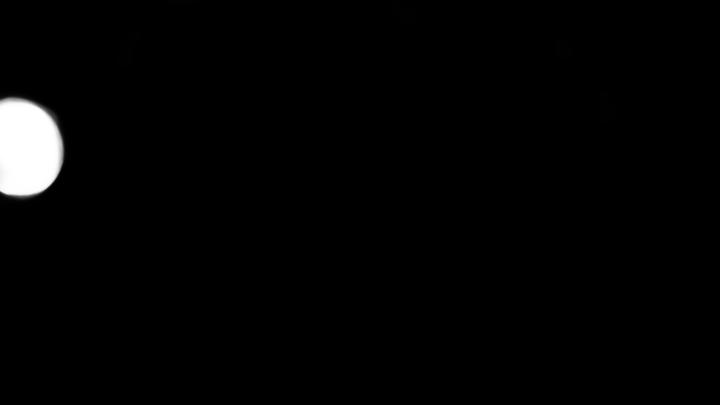}} &
\subfloat{\includegraphics[width=\WidthIms]{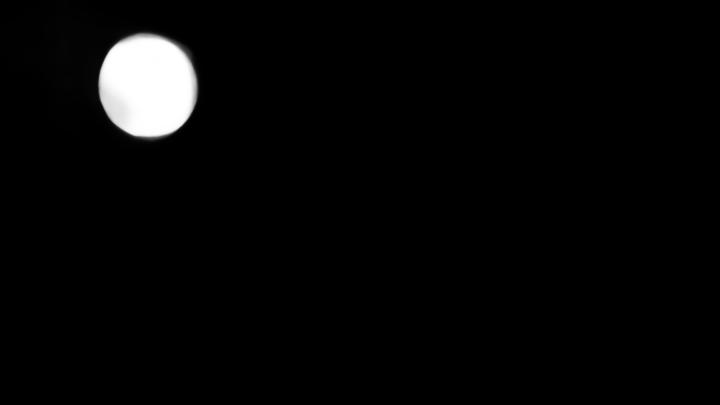}} &
\subfloat{\includegraphics[width=\WidthIms]{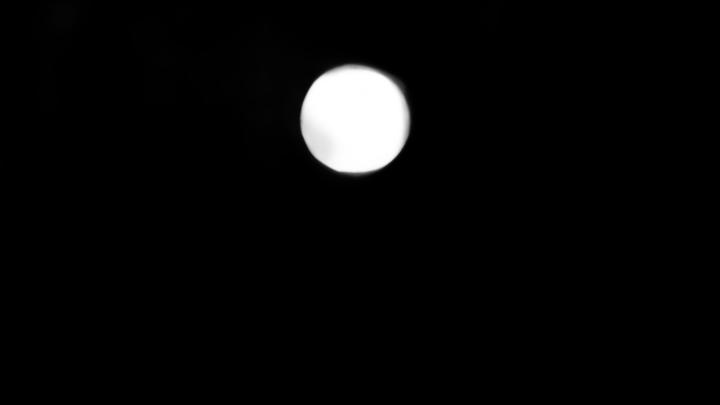}} & 
\subfloat{\includegraphics[width=\WidthIms]{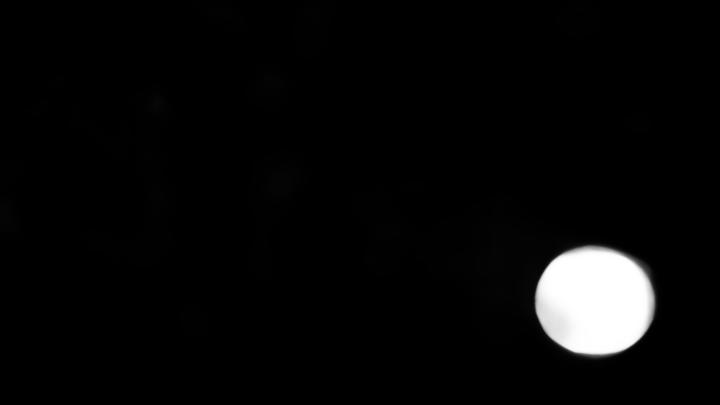}}\\
\end{tabular}
    \caption{Rendered frames 1, 3, 5, and 7 of the test set for the real world ball sequence. The frames are between training frames. Our method produces photorealistic predictions for the unseen time instances. Also, it predicts accurate segmentation masks for the object. }
    \label{fig:renderingsBall}
\end{figure}

\paragraph{Additional Results}
\Cref{fig:renderingsRealPendulum,fig:renderingsBlock,fig:renderingsBall} and show renderings for the test frames of the real world data, as well as visualizations of the object masks obtained by our method. The results show, that our method is able to achieve highly detailed reconstruction for all 3 real world scenarios. Moreover, We obtain accurate masks for the dynamic objects observed in the scene.

\subsection{Potential Negative Societal Impact}
\label{sec:negativeSocietalImpact}
This work attempts to learn interpretable physical models from video clips. While the work is mostly fundamental, it enables a user to edit a scene in a physically plausible manner, at least if the dynamics can be modelled explicitly and the camera and the rest of the scene are static. However, for the physical scenarios that we show, we could not think of possible usages of our method, that could be harmful to individuals or groups of people. In our opinion, the potential for harmful missuse of methods operating on videos is given in particular if the model can alter the actions, expressions or in general the behavior of humans in that scene. In the current state, our method is not able to do such things. 


\end{document}